\PassOptionsToPackage{table}{xcolor}
\documentclass[11pt,letterpaper]{preprint}

\usepackage[utf8]{inputenc}
\usepackage[T1]{fontenc}
\usepackage{url}
\usepackage{booktabs}
\usepackage{multirow}
\usepackage{amsfonts}
\usepackage{amsmath}
\usepackage{amssymb}
\usepackage{nicefrac}
\usepackage{xcolor}
\usepackage{graphicx}
\usepackage{wrapfig}
\usepackage{placeins}
\usepackage[ruled,linesnumbered]{algorithm2e}
\usepackage{siunitx}
\usepackage[capitalize,noabbrev]{cleveref}
\usepackage{bxcoloremoji}
\usepackage{fontawesome5}

\newcommand{\appcontentssection}[3]{%
  \noindent\hyperref[#3]{\textbf{#1}\quad\textbf{#2}}\dotfill\hyperref[#3]{\textbf{\pageref{#3}}}\par
}
\newcommand{\appcontentssubsection}[3]{%
  \noindent\hspace{1.8em}\hyperref[#3]{#1\quad #2}\dotfill\hyperref[#3]{\pageref{#3}}\par
}
\newcommand{\appcontentssubsubsection}[3]{%
  \noindent\hspace{3.8em}\hyperref[#3]{#1\quad #2}\dotfill\hyperref[#3]{\pageref{#3}}\par
}

\graphicspath{{figs/}}

\newcommand{\graycite}[1]{\begingroup\hypersetup{citecolor=gray!70}\cite{#1}\endgroup}

\title{Learning Latent Protein Languages for\\Autoregressive Generation}
\runningtitle{Learning Latent Protein Languages for Autoregressive Generation}

\author[1]{Mahdi Pourmirzaei}
\author[1]{Farzaneh Esmaili}
\author[2]{Amir Ziashahabi}
\author[3]{Mohammadreza Pourmirzaei}
\author[1]{Dong Xu}

\affil[1]{University of Missouri}
\affil[2]{University of Southern California}
\affil[3]{Independent Researcher}

\correspondingauthor{Mahdi Pourmirzaei~(\texttt{mpngf@missouri.edu}), Farzaneh Esmaili~(\texttt{f.esmaili@missouri.edu}), Amir Ziashahabi~(\texttt{ziashaha@usc.edu}), Mohammadreza Pourmirzaei~(\texttt{mo.pourmirzaei@outlook.com}), Dong Xu~(\texttt{xudong@missouri.edu})}

\begin{document}

\begin{abstract}
    Autoregressive transformers are the dominant generative recipe across most tokenized modalities, yet they remain comparatively weak for protein sequence and structure generation. We study the role of target representation in this gap: amino acid tokens encode residue identities without explicit contextual semantics, while backbone coordinates require a discrete representation in our framework.
    We introduce two learned latent protein languages. \emph{Protein Latent Language} (PLL) maps sequences into a 4,096-state contextual alphabet built on a frozen ESM-2 encoder, retaining one token per residue. \emph{Structure Latent Language} (SLL) adapts GCP-VQVAE Lite with auxiliary sequence and confidence supervision while preserving decoding to backbone coordinates. We separately pretrain autoregressive transformer language models on PLL and SLL tokens using a next-token prediction objective, yielding PLLM and SLLM. Under matched downstream sequence training, PLLM has a fitted compute-scaling exponent of 0.038 versus 0.020 for the amino acid autoregressive protein language model (PLM), a roughly 1.9$\times$ steeper slope. SLLM has a fitted exponent of 0.049 on structure-token data. In unconditional sequence generation, PLLM substantially reduces the fraction of samples below a heuristic 1.5-bit residue-composition entropy threshold by 54\% relative to the amino acid model across a sweep of sampling temperatures. At moderate sampling temperatures, PLLM better matches the sequence lengths and residue-composition entropy of the UniRef50 training data than PLM does. SLL improves performance on supervised protein structure tasks. For sequence-to-structure prediction, replacing the original GCP-VQVAE Lite tokenizer with SLL reduces best validation perplexity by 34\% under matched training. In backbone generation, SLLM compares favorably with other generative models on diversity and novelty. We further use SLLM for sequence-to-structure prediction, where latent-token sampling for long proteins is approximately \num{1000} times faster than MSA-based AlphaFold2 in our measurements, and we observe early signs that using the model's own internal token confidence for inference-time scaling can lift prediction quality beyond a single decoded sample.
    Together, these results position learned latent protein languages as a promising direction for language modeling to explore, offering a practical substrate for bringing autoregressive transformer scaling and inference-time sampling to protein generation.
    
\vspace{4mm}

\faGithub~\textbf{Code Repository}: \href{https://github.com/mahdip72/latent_protein_languages}{github.com/mahdip72/latent\_protein\_languages} \\
\faGlobe~\textbf{Project Page}: \href{https://mahdip72.github.io/latent-protein-languages.github.io/}{mahdip72.github.io/latent-protein-languages.github.io/}
\end{abstract}

\maketitle

\section{Introduction}
\label{sec:introduction}

Autoregressive transformer language models have become the central paradigm of frontier generative artificial intelligence (AI).
Originally developed for text, the same decoder-only next-token recipe now operates as a unified multimodal interface across images~\cite{var}, audio~\cite{soundstream}, video, and most recently embodied actions.
These modalities are mapped to discrete tokens and generated by the same architecture~\cite{anygpt,gpt4o}, as in vision-language-action systems for robotics~\cite{rt2}.
The appeal of this paradigm is a single, scalable recipe that absorbs new modalities through tokenization rather than through architectural redesign.

Proteins are a natural target for this recipe, yet autoregressive modeling has not become the dominant approach on either of their two modalities.
On the sequence side, autoregressive protein language models exhibit power-law loss scaling with compute, while data repetition can limit gains from further training~\cite{compute_optimal_plm}. Models such as ProtGPT2~\cite{protgpt2} and ProGen~\cite{progen2,progen3} nevertheless remain less effective for protein design than non-autoregressive alternatives such as the discrete diffusion protein language model (DPLM)~\cite{dplm}.
Their autoregressive sampling is also prone to drifting into pathological low-complexity, repetitive sequences that lack structural plausibility~\cite{plm_repetition}.
On the structure side, the gap is sharper because state-of-the-art de novo backbone generation is dominated by diffusion and flow-matching models~\cite{chroma,rfdiffusion,foldflow}, while autoregressive structure generation is essentially absent from the strongest baselines.
The fact that the recipe powering frontier generative AI in other domains underperforms across both protein modalities is, in our view, the central problem.

We study how the choice of target representation affects autoregressive protein modeling.
Amino acid tokens encode residue identity, while learned latent tokens can incorporate contextual information from a pretrained protein encoder.
Raw three-dimensional coordinates require a discrete representation in this framework.
Learned tokenizers in vision and audio motivate a similar approach for proteins: generate decodable latent tokens and recover the original observations.
We therefore investigate learned target languages while retaining a standard autoregressive modeling recipe.

We address tokenization on both modalities and show that, once each side of the protein domain is mapped into a learned discrete latent language with high semantic density, autoregressive modeling becomes competitive with the non-autoregressive approaches on selected diversity and novelty metrics in our generation benchmarks.
We introduce two such languages.
\emph{Protein Latent Language} (PLL) is a sequence-side discrete language built on top of ESM-2~\cite{esm2} semantic representations using a vector-quantized representation autoencoder (VQ-RAE) tokenizer~\cite{vqrae}, and is decodable back to amino acid sequences.
\emph{Structure Latent Language} (SLL) is a structure-side discrete language designed to be reconstructable to backbone coordinates while remaining rich in structural and functional semantics.
We then study PLL and SLL along the axes of scaling, de novo generation, and sequence-to-structure prediction.
We refer to autoregressive models trained on PLL and SLL as protein latent language models (PLLMs) and structure latent language models (SLLMs), respectively.
PLLM and SLLM are pretrained separately.
For sequence-to-structure prediction, SLL generation is conditioned on E1~\cite{e1} sequence representations.
The contributions of this study are summarized below.

\begin{itemize}
    \setlength{\itemsep}{-0.25em}
    \setlength{\parsep}{0pt}
    \setlength{\topsep}{0.2em}
    \setlength{\partopsep}{0pt}
    \item We introduce PLL, a contextual, residue-aligned target language derived from ESM-2 representations, and adapt GCP-VQVAE Lite into SLL through auxiliary supervision. Both languages remain decodable to amino acids or backbone coordinates.
    \item We show that PLLM has steeper fitted loss-versus-compute scaling and substantially less low-complexity drift than the amino acid model under matched downstream training.
    \item We apply SLLM to structure generation and sequence-to-structure prediction, combining confidence filtering adapted from DeepConf~\cite{deepconf} with token-space consensus to select candidates before coordinate decoding.
\end{itemize}

\vspace{-1.0em}

\section{Related Work}
\label{sec:related_work}
Autoregressive protein generation is most established for sequences.
ProtGPT2~\cite{protgpt2}, RITA~\cite{rita}, ProGen2~\cite{progen2}, and ProGen3~\cite{progen3} train decoder-only models over amino acid strings, while compute-optimal studies show that objective choice and data diversity shape scaling~\cite{compute_optimal_plm}.
Structure-side autoregression is newer.
Learning the Language of Protein Structure~\cite{llps} tokenizes backbones with a vector-quantized autoencoder and trains a decoder-only transformer over structure codes.
Structure Language Models~\cite{slm_conformation} encode conformations with a discrete variational autoencoder and study conditional language models, including decoder-only variants, for sequence-conditioned ensembles.
Prot2Token~\cite{prot2token} maps protein prediction targets, including three-dimensional structure, into next-token outputs from an autoregressive decoder.
HelixProtX~\cite{helixprotx} uses a large multimodal-model interface for any-to-any sequence, structure, and description generation.

Non-autoregressive models set many current protein-generation baselines.
DPLM~\cite{dplm} uses discrete diffusion for sequence generation, and DPLM-2~\cite{dplm2} extends diffusion to joint sequence-structure modeling with lookup-free quantized structure tokens.
For structures, RFdiffusion~\cite{rfdiffusion}, Chroma~\cite{chroma}, FoldFlow~\cite{foldflow}, ProteinGenerator~\cite{proteingenerator}, and La-Proteina~\cite{laproteina} use diffusion, flow matching, or programmable geometric generation to produce backbones or sequence-structure pairs with strong controllability and designability.
These methods avoid a left-to-right sampling order and motivate our empirical comparison of whether autoregressive transformers become competitive when operating in learned protein latent languages.

\section{Method}
\label{sec:method}
We start from the premise that the failure mode of autoregressive protein modeling is partly a representation problem.
Instead of changing the next-token objective, we change the objects being predicted.
We require a target language for autoregressive generative modeling to satisfy three properties.
First, its vocabulary should be large enough to express fine-grained biological states, consistent with scaling-law evidence that larger language models benefit from larger vocabularies~\cite{vocab_scaling}.
Second, its tokens should carry high-level semantics rather than only local symbols, following the role of compact latent tokenizers in visual reconstruction and generation~\cite{titok}.
Third, it should exhibit autoregressive friendliness, so the token sequence should be easy for a causal generator to model rather than optimized only for reconstruction fidelity~\cite{worse_better_tokenization}.
For a protein sequence $x$ and backbone structure $y$, we therefore learn discrete tokenizations that map each modality into such an ordered latent sequence.
The sequence tokenizer defines PLL, with codes that can be decoded to amino acid identities.
The structure tokenizer defines SLL, with codes that can be decoded to backbone coordinates.

With this interface fixed, we train decoder-only transformers on raw amino acid, PLL, and SLL streams using the same next-token factorization.
The matched amino acid--PLL comparison evaluates the complete PLL target representation, including pretrained ESM-2 semantics and discretization.
This comparison does not isolate the effects of ESM-2 semantics, vector quantization, vocabulary expansion, or learned tokenizer transformations.
SLL is studied separately on structure data.
We refer to models that generate PLL or SLL streams as autoregressive latent language models, while the raw amino acid model serves as a non-latent autoregressive reference.
In the following, we describe how these latent languages are constructed and how autoregressive models are used to generate them.

\subsection{PLL}
\label{sec:pll}
\begin{figure}[t]
    \centering
    \includegraphics[width=\linewidth]{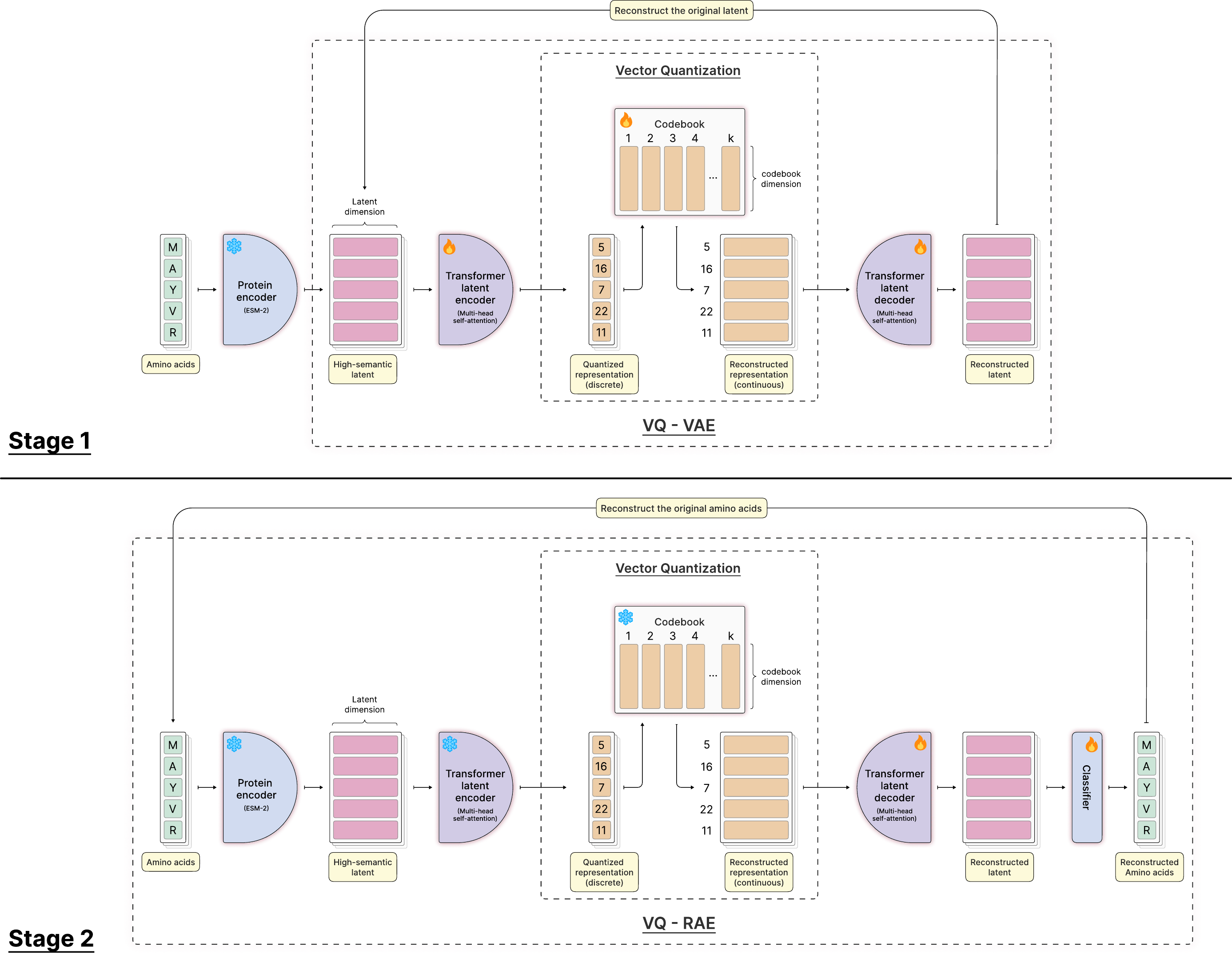}
    \caption{PLL construction. Stage 1 trains a vector-quantized variational autoencoder (VQ-VAE)-style latent reconstructor for frozen ESM-2 representations. Stage 2 finalizes PLL as a VQ-RAE by freezing the latent encoder and codebook and training a reinitialized transformer decoder with a linear residue-classification head.}
    \label{fig:pll_tokenizer}
\end{figure}

PLL provides a contextual re-alphabetization of protein sequences.
Its role is to replace residue symbols with discrete tokens that inherit the semantic organization of a pretrained protein encoder while remaining decodable to amino acid sequences.
The vector-quantization bottleneck follows established VQ-VAE~\cite{vqvae} principles, while the final decodable language uses a VQ-RAE-style~\cite{vqrae} construction.
Given a sequence $x=(x_1,\ldots,x_L)$, a frozen ESM-2 150M encoder~\cite{esm2} produces contextual representations $\mathbf h_{1:L}$.
We use this encoder as a compute-efficient semantic backbone, since the Protein Foundation Model Benchmark~\cite{pfmbench} reports that larger ESM-2 variants give only marginal gains over ESM-2 150M relative to their inference cost.
A transformer latent encoder maps these vectors to $\mathbf u_{1:L}$, and a codebook $\mathbf C=\{\mathbf c_k\}_{k=1}^{K}$ assigns each residue position to a discrete code by negative Euclidean distance,
\begin{equation}
    \label{eq:pll_quantization}
    \ell_{t,k}=-\|\mathbf u_t-\mathbf c_k\|_2,
    \qquad
    z_t^{\mathrm{PLL}}=\arg\max_{k\in\{1,\ldots,K\}}\ell_{t,k},
    \qquad
    \mathbf q_t=\mathbf c_{z_t^{\mathrm{PLL}}}.
\end{equation}
The result is an ordered PLL token sequence with one discrete code per residue.
In our experiments, $K=4096$.
Multiple PLL codes can decode to the same amino acid, allowing the target alphabet to distinguish contextual states while preserving sequence length and residue alignment.
During Stage 1 training, code assignments are sampled from the same distance scores, while evaluation and decoding use the deterministic rule in \cref{eq:pll_quantization}.
Implementation details are given in \cref{app:pll_architecture}.
The two-stage pipeline in \cref{fig:pll_tokenizer} makes the construction explicit.
Stage 1 learns the quantized bottleneck through latent reconstruction, while Stage 2 freezes this bottleneck and trains amino acid recovery from the fixed PLL codes.

Stage 1 trains the tokenizer as a VQ-VAE-style latent reconstruction model.
The ESM-2 encoder is fixed, while the latent encoder and latent decoder $D_\theta$ are optimized around the codebook to reconstruct $\mathbf h_{1:L}$ from the quantized vectors $\mathbf q_{1:L}$, producing $\hat{\mathbf h}_{1:L}=D_\theta(\mathbf q_{1:L})$.
The reconstruction target is therefore the ESM-2 latent representation, not the amino acid sequence.
The Stage 1 training loss combines a latent reconstruction loss and a vector-quantization loss,
\begin{equation}
    \label{eq:pll_stage1_loss}
    \mathcal{L}_{\mathrm{PLL}}^{(1)}
    =
    \mathcal{L}_{\mathrm{rec}}+\mathcal{L}_{\mathrm{vq}}
    =
    10\,\mathcal{L}_{\mathrm{mse}}+1\,\mathcal{L}_{\mathrm{cos}}+0.5\,\mathcal{L}_{\mathrm{commit}} .
\end{equation}
\Cref{app:pll_architecture} gives the detailed definitions of these terms in \cref{eq:app_pll_stage1_terms}.
The reconstruction metrics used for monitoring and diagnostics are described in \cref{app:pll_metrics}.

Stage 2 turns the learned codes into the final VQ-RAE-style PLL.
Starting from the selected Stage 1 checkpoint, we freeze the latent encoder and codebook, discard the Stage 1 decoder weights, and train a reinitialized transformer decoder with a linear 21-class residue-classification head on the fixed quantized vectors.
The classifier covers the 20 standard amino acids plus the unknown residue class $X$.
The Stage 2 training loss, $\mathcal{L}_{\mathrm{PLL}}^{(2)}$, is token-level residue cross-entropy over the fixed PLL code sequence.
Because the latent encoder and codebook are fixed, amino acid recovery depends only on information stored in the discrete code sequence.
At generation time, an autoregressive latent language model produces PLL code IDs, the fixed codebook maps them back to quantized vectors, and the Stage 2 transformer decoder and linear classification head predict amino acid identities as $\arg\max_c p_\phi(c\mid z^{\mathrm{PLL}}_{1:L}, t)$.
ESM-2 is used to construct PLL targets from observed sequences. PLLM sampling and residue decoding run without ESM-2.

\subsection{SLL}
\label{sec:sll}
SLL is the structure-side latent language.
It builds on GCP-VQVAE~\cite{gcp_vqvae}, a geometry-complete vector-quantized autoencoder tokenizer for protein backbones that maps structures to a 4096-token discrete codebook and decodes those codes back to backbone coordinates.
The original GCP-VQVAE work introduces Lite and Large variants. We use GCP-VQVAE Lite as the baseline architecture for SLL, so the comparison isolates the effect of the tokenizer objective rather than increasing model scale.
We keep the Lite backbone-tokenizer interface and use it as the structural counterpart to PLL.
Given a backbone $y$, a GCPNet encoder produces residue-level geometric embeddings, a transformer encoder contextualizes them, and vector quantization assigns one SLL code per residue.
A transformer decoder then maps the quantized representation to backbone coordinates through a six-dimensional rotation head.
For the single-chain monomers studied here, the input's absolute translation and orientation are omitted.
The decoder reconstructs the backbone in its own coordinate frame, and reconstruction is evaluated up to a global rigid transformation.

\begin{figure}[t]
    \centering
    \includegraphics[width=\linewidth]{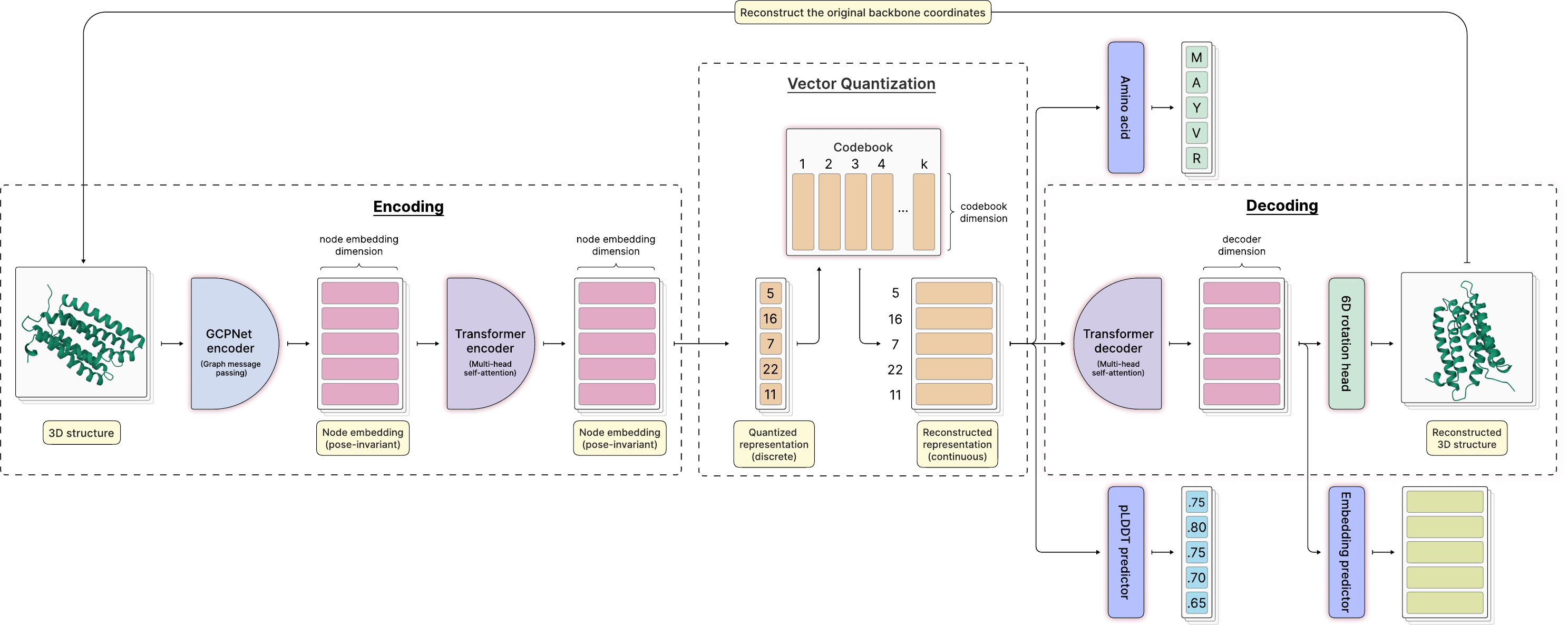}
    \caption{SLL construction. A GCPNet and transformer encoder convert backbone geometry into discrete structure codes. The decoder reconstructs backbone coordinates, while auxiliary heads train the codes to retain residue identity, confidence, and sequence-representation information.}
    \label{fig:sll_tokenizer}
\end{figure}

We adapt the GCP-VQVAE Lite training objective to study the tradeoff between reconstruction fidelity, semantic utility, and autoregressive predictability.
The training objective still includes backbone reconstruction and vector-quantization terms, but adds three auxiliary targets, shown in \cref{fig:sll_tokenizer}.
Two pre-decoder heads predict amino acid identity and per-residue confidence measured by the predicted Local Distance Difference Test (pLDDT) score from the quantized stream, encouraging the codes themselves to retain sequence and confidence information.
A post-decoder head predicts frozen ESM-2 representations, encouraging the reconstructed hidden states to remain aligned with protein-sequence semantics.
These auxiliary heads are used during tokenizer training only.
After training, autoregressive latent language models operate on SLL code IDs, and generated SLL sequences are decoded back to backbones with the fixed structure decoder.

We select the final SLL recipe using the Protein Structure Tokenization benchmark (PST), also known as StructTokenBench~\cite{structtokenbench}, which evaluates whether tokenizer embeddings support functional, physicochemical, and structural prediction tasks beyond coordinate reconstruction.
We treat the auxiliary heads and reconstruction-loss choices as tokenizer-design choices, and select the final active recipe by supervised PST while keeping the GCP-VQVAE Lite architecture and training corpus fixed.
This selection criterion matches the role of SLL in the rest of the paper.
The tokenizer must remain reconstructable, but its primary purpose is to provide an autoregressive-friendly structure language whose tokens carry information useful for generation and sequence-to-structure prediction.
Architecture and metric details are given in \cref{app:sll_architecture,app:sll_metrics}.

\subsection{Autoregressive Protein Modeling}
\label{sec:ar_latent_modeling}
After PLL and SLL are fixed, we model them alongside raw amino acid sequences as three standalone protein languages.
Each language is tokenized residue-wise, and each token stream is pretrained separately with a decoder-only autoregressive transformer using cross-entropy next-token prediction.
This follows the standard Transformer~\cite{transformer} and GPT-2~\cite{gpt2} recipe, while changing only the protein token space.
We use PLLM and SLLM for the autoregressive latent language models trained on PLL and SLL streams, respectively.
For amino acid sequences, the same architecture is a non-latent autoregressive reference.
For unconditional generation, amino acid, PLL, and SLL streams use the same simple format with a beginning token, a content sequence, and an end token.
For controlled de novo generation, we condition the same autoregressive model by prepending special tokens, such as a length token, before the generated content.
The appendix gives the full token inventory and example streams in \cref{app:ar_tokenization_protocol}.

\subsubsection{Sequence-to-Structure Generation}
\label{sec:conditional_s2s}

\begin{figure}[!ht]
    \centering
    \includegraphics[width=0.8\linewidth]{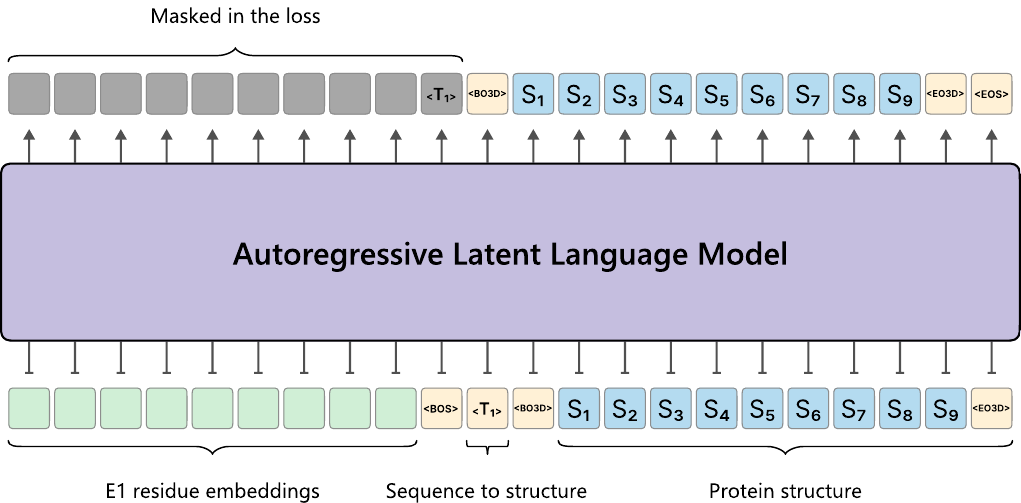}
    \caption{Sequence-to-structure generation via SLLM. E1 encodes the input sequence. The decoder predicts SLL tokens, which the SLL decoder maps to coordinates. Green, yellow, and blue denote protein embeddings, special, boundary, and condition tokens, and SLL tokens.}
    \label{fig:sequence_to_structure}
\end{figure}

We also adapt the pretrained SLLM for sequence-to-structure generation, as shown in \cref{fig:sequence_to_structure}. The input amino acid sequence is encoded outside the causal token stream with Profluent-E1 600M~\cite{e1}.
We use E1 because it provides strong structure-aware sequence representations at moderate encoder scale, including superior unsupervised contact-map prediction relative to comparable publicly available protein encoders.
The resulting residue-level embeddings are projected into the decoder hidden dimension and supplied as continuous context.
We use the sequence-to-structure task token to condition the decoder for structure generation, with example token streams shown in \cref{tab:app_ar_stream_examples}.
The decoder then generates SLL structure tokens autoregressively, and the fixed SLL decoder maps the generated token sequence back to backbone coordinates.

\subsubsection{Confidence-Based Candidate Selection}
\label{sec:confidence_selection}
To test whether sequence-to-structure prediction benefits from inference-time scaling, we sample multiple SLL token sequences for the same input and score each candidate before decoding it back to coordinates.
In an offline selection procedure adapted from Deep Think with Confidence (DeepConf)~\cite{deepconf}, we treat raw full-vocabulary entropy from the SLLM as an internal confidence signal for latent structure generation.
The central question is whether confidence measured in SLL space correlates with final structure quality and can select better candidates without ground truth.
Oracle best-of-$k$ is used only as an evaluation ceiling.
The practical selector uses confidence summaries and token-space consensus over sampled SLL sequences, as described in \cref{app:confidence_selection_details}.

\section{Experiments}
\label{sec:experiments}
We organize the experiments around the two layers of the approach.
First, we evaluate whether the learned tokenizers define useful discrete protein languages.
PLL should preserve sequence information through latent reconstruction and amino acid recovery, while SLL should preserve structure and improve the semantic utility of GCP-VQVAE Lite through auxiliary objectives.
Second, after the tokenizers are fixed, we train an amino acid language model, PLLM, and SLLM and test whether the latent token spaces improve scaling and generation.

All sequence-side experiments use UniRef50 as the source corpus.
SLL tokenizer experiments use AFDB representative structures, while SLLM generation uses the larger AFDB$\cap$UniRef50 structure-token corpus introduced by GCP-VQVAE~\cite{gcp_vqvae}.
Further data-source details are given in \cref{app:data_sources}.
All experiments are implemented in PyTorch 2.8~\cite{pytorch}\footnote{The transformer and vector-quantization layers build on \href{https://github.com/lucidrains/x-transformers}{\texttt{x-transformers}} and \href{https://github.com/lucidrains/vector-quantize-pytorch}{\texttt{vector-quantize-pytorch}}.} and run on four nodes, each with eight NVIDIA A100 80 GB GPUs.
Unless stated otherwise, training uses FlashAttention-2 (FA2)~\cite{flashattention2}, mixed precision~\cite{mixed_precision}, AdamW~\cite{adamw}, and cosine learning-rate decay~\cite{sgdr}.

\subsection{Latent Tokenizers}
\label{sec:latent_tokenizer_results}

\subsubsection{PLL Results}
\label{sec:pll_results}
The PLL tokenizer is evaluated at two points that matter for the rest of the paper.
Stage 1 evaluates reconstruction fidelity in the frozen ESM-2 representation space.
The final Stage 1 checkpoint reaches $\Delta\mathrm{PPL}_{\%}=0.47$, with median normalized mean squared error (NMSE) of 0.044, reconstruction Kullback-Leibler (KL) divergence of 0.003, cosine similarity of 0.970, and full code activation on the validation split.
The $\Delta\mathrm{PPL}_{\%}$ metric is the primary monitoring metric because it measures how much the reconstructed latent distribution changes relative to the original ESM-2 latent distribution.
The latent reconstruction delta perplexity of 0.47\% and cosine similarity of 0.970 indicate close reconstruction of the frozen ESM-2 representations on the validation split. Given the contextual protein information learned by ESM-2, these results are consistent with retention of that information in the discrete PLL representation.

Stage 2 then freezes the Stage 1 encoder and codebook and asks whether the resulting discrete PLL codes are sufficient to recover amino acid identities.
The fixed codes achieve 99.81\% micro token-level residue recovery on the UniRef50 validation split.
All standard amino acids have class-wise recovery above 99.17\%.
Together, these results support using PLL as a decodable sequence-side latent language for the autoregressive experiments that follow.

\subsubsection{SLL Results}
\label{sec:sll_results}
The SLL experiments keep the GCP-VQVAE Lite architecture and training corpus fixed, and change the tokenizer objective.
We refer to the selected SLL tokenizer as GCP-VQVAE Lite 2.
The ablation study trains each candidate recipe on 200,000 samples from the main SLL tokenizer source corpus and uses supervised PST as the primary selection target because the new auxiliary heads are intended to improve the semantic utility of the structure codes.
The combined recipe gives the strongest average supervised PST performance among the comparable ablations, improving over the base tokenizer by 24.2\% in functional-site AUROC and 35.0\% in physicochemical Spearman (\cref{tab:sll_pst_ablation_summary}).
The full set of supervised ablation averages is given in \cref{app:sll_supervised_pst_ablation}.

\begin{wraptable}{r}{0.58\linewidth}
    \vspace{-1.0em}
    \centering
    \scriptsize
    \setlength{\tabcolsep}{2.0pt}
    \caption{SLL supervised PST ablation averages.}
    \label{tab:sll_pst_ablation_summary}
    \begin{tabular}{
        @{}
        >{\raggedright\arraybackslash}p{0.36\linewidth}
        cc
        @{}}
        \toprule
        Configuration & \shortstack{Functional\\AUROC (\%) $\uparrow$} & \shortstack{Physicochemical\\Spearman (\%) $\uparrow$} \\
        \midrule
        Base & 58.26 & 23.93 \\
        MSE + distance & \cellcolor{green!18}61.28 & \cellcolor{green!26}30.32 \\
        pLDDT pre-dec. & \cellcolor{green!18}62.65 & \cellcolor{green!26}28.95 \\
        Inverse folding 0.1 & \cellcolor{green!26}67.15 & \cellcolor{red!10}23.66 \\
        Combined & \cellcolor{green!36}\textbf{72.37} & \cellcolor{green!26}\textbf{32.30} \\
        \bottomrule
    \end{tabular}
    \vspace{-1.0em}
\end{wraptable}

After training the selected recipe on the full SLL tokenizer setup, GCP-VQVAE Lite 2 improves supervised PST averages over the original Lite and Large tokenizers from the original work (\cref{tab:sll_supervised_pst_full_run}).
The detailed task-level table is given in \cref{app:sll_supervised_pst_full_run}.
Unsupervised PST scores remain high but are treated as secondary and reported in \cref{app:sll_unsupervised_pst_full_run}.

\begin{table}[h]
    \centering
    \scriptsize
    \setlength{\tabcolsep}{2.2pt}
    \caption{Average supervised PST comparison with baseline tokenizers. Lite 2 denotes SLL, and parentheses give the four-layer MLP and transformer probe results for GCP-VQVAE variants~\cite{gcp_vqvae}. Gray columns provide external context and are not used for bold and underline ranking.}
    \label{tab:sll_supervised_pst_full_run}
    \resizebox{\linewidth}{!}{%
    \begin{tabular}{lccccccccc}
        \toprule
        Metric &
        \textcolor{gray!70}{\shortstack{FoldSeek\\\graycite{foldseek}}} &
        \textcolor{gray!70}{\shortstack{ProTokens\\\graycite{protokens}}} &
        \textcolor{gray!70}{\shortstack{VanillaVQ\\\graycite{structtokenbench}}} &
        \textcolor{gray!70}{\shortstack{AIDO\\\graycite{aido_structure_tokenizer}}} &
        \textcolor{gray!70}{\shortstack{ESM3\\\graycite{esm3}}} &
        \textcolor{gray!70}{\shortstack{AminoAseed\\\graycite{structtokenbench}}} &
        \shortstack{GCP-VQVAE\\Large~\cite{gcp_vqvae}} &
        \shortstack{GCP-VQVAE\\Lite~\cite{gcp_vqvae}} &
        \shortstack{GCP-VQVAE\\Lite 2} \\
        \midrule
        Functional AUROC (\%) $\uparrow$ &
        \textcolor{gray!70}{51.90} & \textcolor{gray!70}{65.37} & \textcolor{gray!70}{68.30} & \textcolor{gray!70}{69.38} & \textcolor{gray!70}{69.24} & \textcolor{gray!70}{72.43} &
        61.19 (68.10, 73.57) &
        61.84 (68.30, \underline{74.93}) &
        71.32 (71.65, \textbf{78.08}) \\
        Physicochemical Spearman (\%) $\uparrow$ &
        \textcolor{gray!70}{7.80} & \textcolor{gray!70}{9.84} & \textcolor{gray!70}{33.49} & \textcolor{gray!70}{21.81} & \textcolor{gray!70}{37.35} & \textcolor{gray!70}{38.08} &
        16.61 (17.89, 21.37) &
        20.84 (22.73, 23.89) &
        \underline{29.63} (\textbf{29.71}, 27.52) \\
        \bottomrule
    \end{tabular}}
\end{table}

SLL remains competitive with state-of-the-art reconstructive tokenizers.
Its average TM-score is 2.4\% below the original GCP-VQVAE Lite tokenizer and close to AIDO (0.9456 versus 0.9470), while its average RMSD is lower (1.4722 versus 1.8189\,\AA) (\cref{tab:sll_reconstruction_comparison}).
SLL also achieves higher functional-site AUROC and physicochemical Spearman than AIDO in the reported supervised PST comparison.
It retains the GCP-VQVAE Lite backbone, for which the original benchmark reported substantially lower inference latency than AIDO~\cite{gcp_vqvae}.
The full split-level reconstruction table is given in \cref{app:sll_reconstruction_comparison}.

\begin{table}[h]
    \centering
    \scriptsize
    \setlength{\tabcolsep}{2.4pt}
    \caption{Average reconstruction quality over CASP14, CASP15, CASP16, CAMEO-2024, and the zero-shot set from the original benchmark~\cite{gcp_vqvae}, where Lite 2 is our SLL tokenizer. The grayed GCP-VQVAE Large column is not used for bold and underline ranking.}
    \label{tab:sll_reconstruction_comparison}
    \resizebox{0.9\linewidth}{!}{%
    \begin{tabular}{lccccccccc}
        \toprule
        Metric &
        \shortstack{GCP-VQVAE\\Lite 2} &
        \textcolor{gray!70}{\shortstack{GCP-VQVAE\\Large~\graycite{gcp_vqvae}}} &
        \shortstack{GCP-VQVAE\\Lite~\cite{gcp_vqvae}} &
        \shortstack{AIDO\\\cite{aido_structure_tokenizer}} &
        \shortstack{ESM-3\\VQVAE\\\cite{esm3}} &
        \shortstack{DPLM-2\\VQVAE\\\cite{dplm2}} &
        \shortstack{FoldToken 4\\\cite{foldtoken}} &
        \shortstack{Gaujac\\et al.\\\cite{llps}} &
        \shortstack{AminoAseed\\\cite{structtokenbench}} \\
        \midrule
        Average TM-score $\uparrow$ & 0.9456 & \textcolor{gray!70}{0.9844} & \textbf{0.9687} & \underline{0.9470} & 0.9291 & 0.8311 & 0.8788 & 0.8394 & 0.4000 \\
        Average RMSD $\downarrow$ & \underline{1.4722} & \textcolor{gray!70}{0.6172} & \textbf{0.9364} & 1.8189 & 2.8905 & 7.3650 & 3.0106 & 6.0002 & 18.2178 \\
        \bottomrule
    \end{tabular}}
\end{table}

\begin{figure}[!ht]
    \centering
    \includegraphics[width=0.7\linewidth]{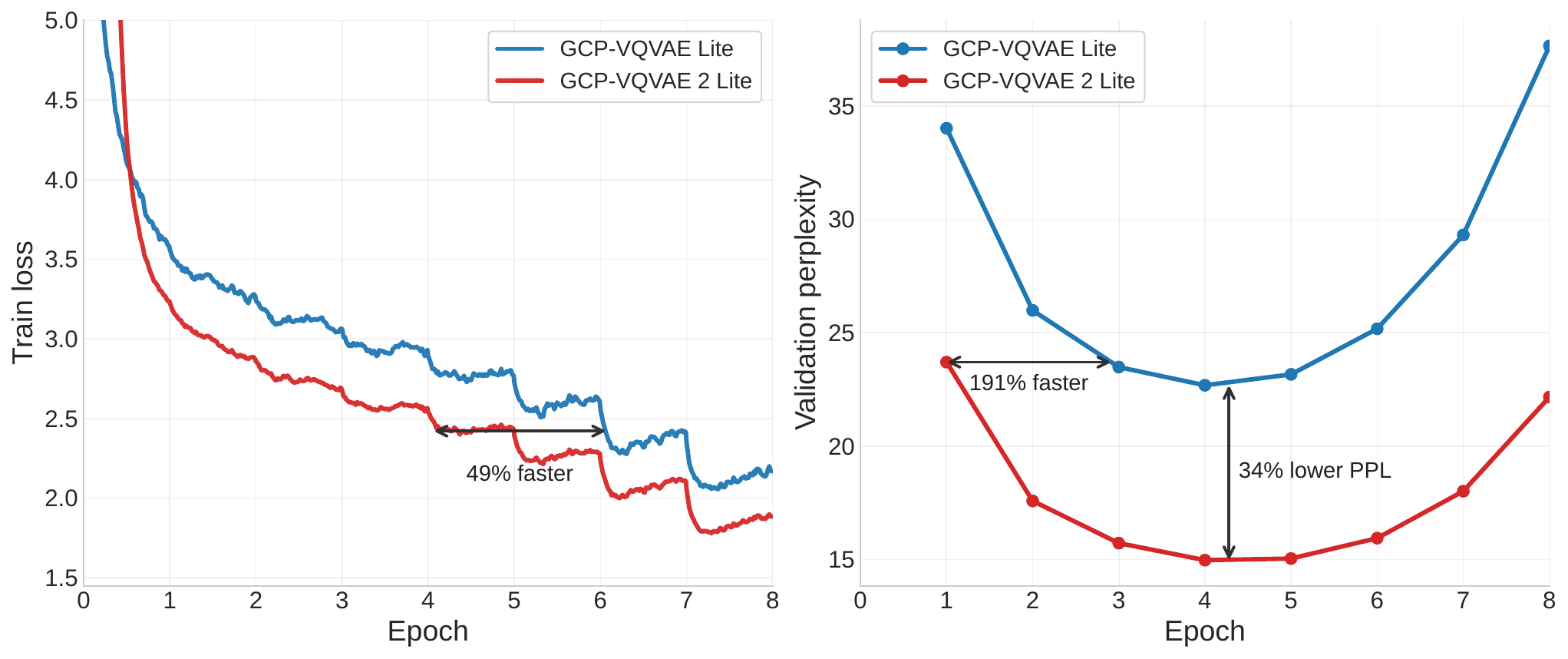}
    \caption{Matched sequence-to-structure tokenizer comparison. Left: train loss. Right: validation perplexity, where Lite 2 lowers the best value by 34\%. The learning-speed annotations compare training progress by epoch.}
    \label{fig:sll_s2s_tokenizer_ablation}
\end{figure}

Finally, we ask whether the improved SLL tokenizer is also a better target language for autoregressive sequence-to-structure prediction.
In a matched Prot2Token~\cite{prot2token}-style setup, changing only the target tokenizer reduces best validation perplexity by 34\% and reaches a comparable training-loss regime 49\% faster when progress is measured by training epoch (\cref{fig:sll_s2s_tokenizer_ablation}).
This suggests that SLL improves not only representation quality, but also autoregressive compatibility.
Full setup details are given in \cref{app:sll_s2s_tokenizer_ablation}.

\begin{figure}[h]
    \centering
    \begin{minipage}{0.32\linewidth}
        \centering
        \includegraphics[width=\linewidth]{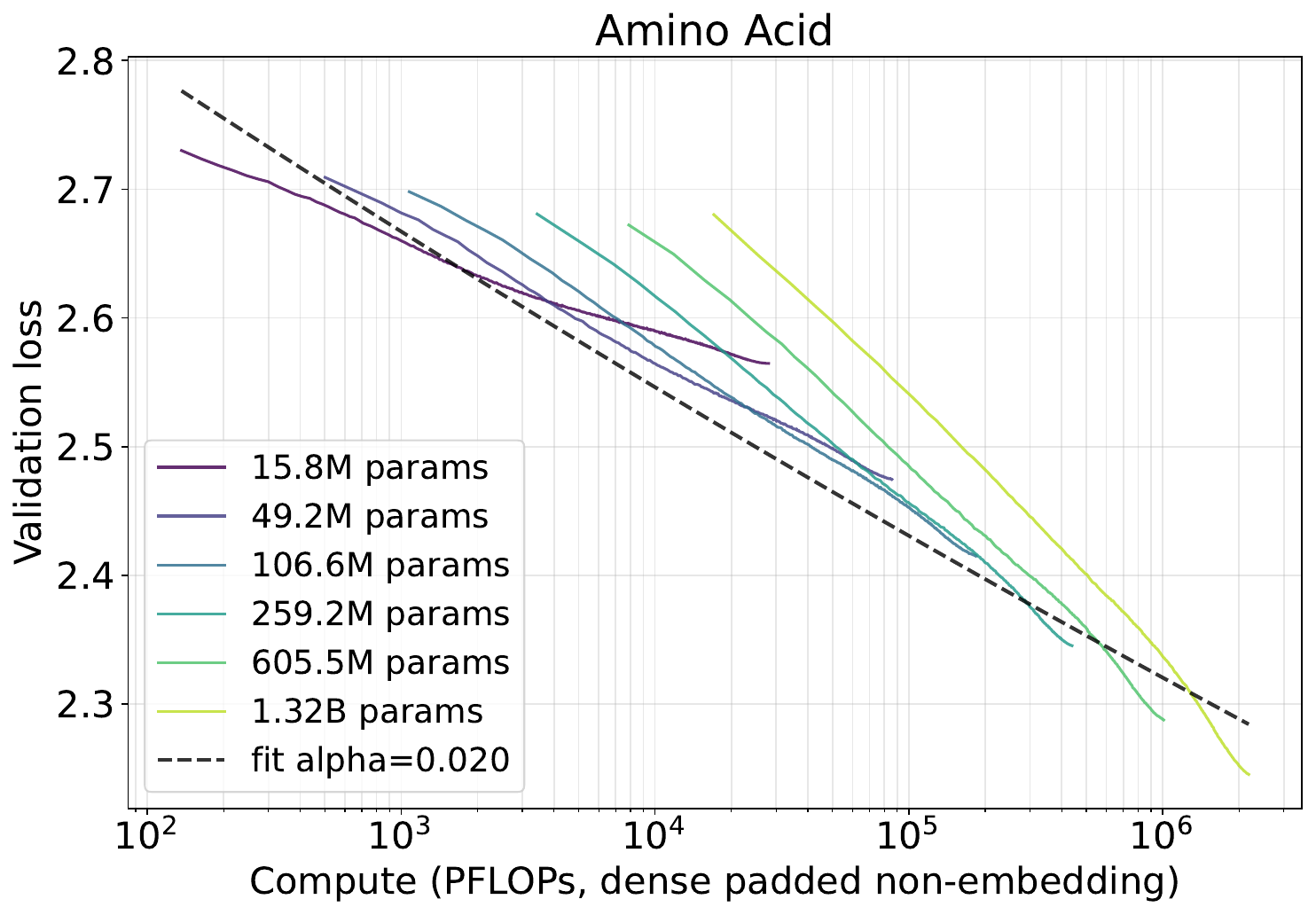}
        \footnotesize (a)
    \end{minipage}
    \hfill
    \begin{minipage}{0.32\linewidth}
        \centering
        \includegraphics[width=\linewidth]{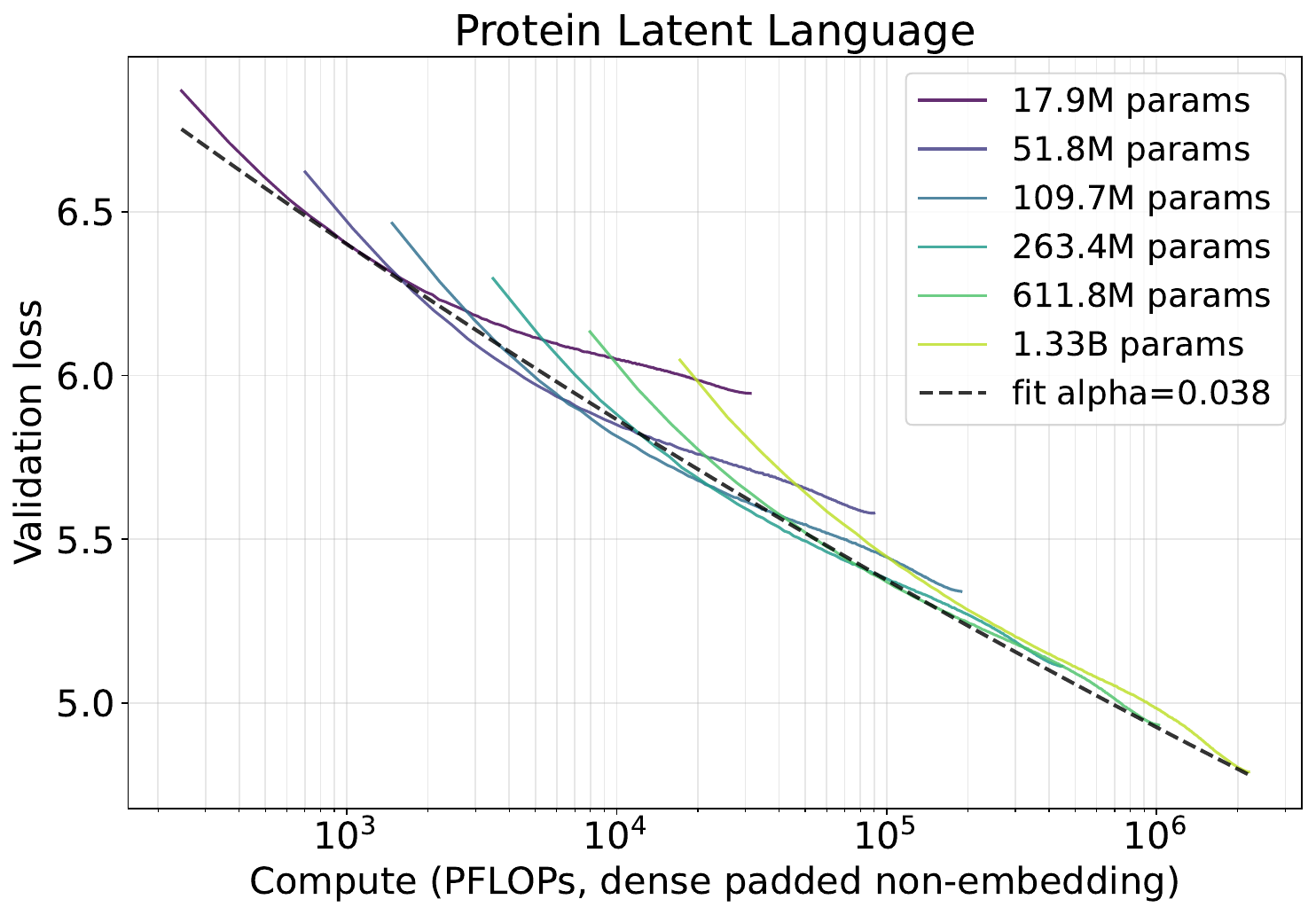}
        \footnotesize (b)
    \end{minipage}
    \hfill
    \begin{minipage}{0.32\linewidth}
        \centering
        \includegraphics[width=\linewidth]{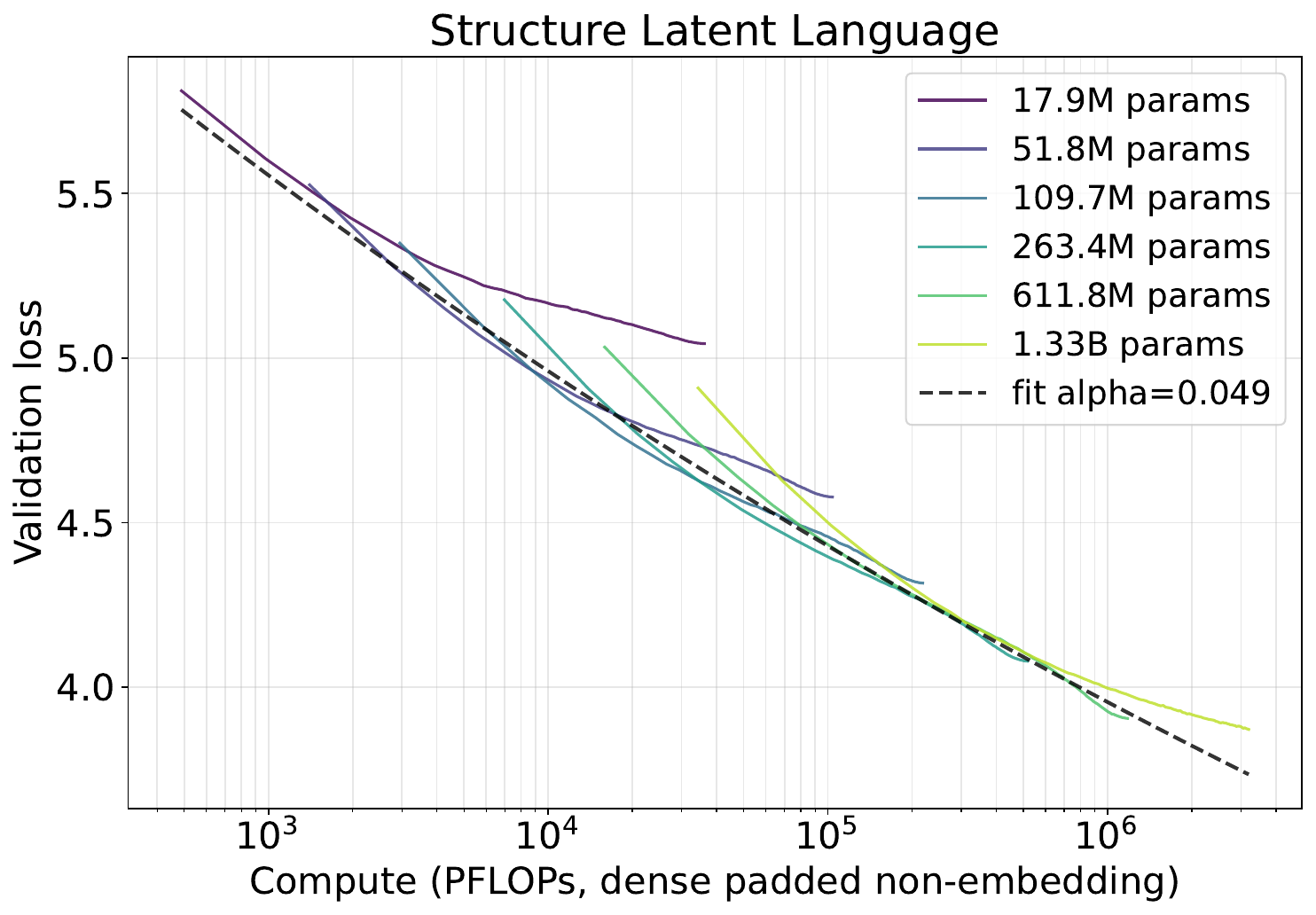}
        \footnotesize (c)
    \end{minipage}
    \caption{Validation loss versus estimated downstream autoregressive training compute. PLLM has a steeper fitted slope than the amino acid model under matched sequence training. SLLM is shown separately with structure-token data and different training exposure.}
    \label{fig:scaling_loss_vs_compute}
\end{figure}
\FloatBarrier

\subsection{Autoregressive Protein Language Models}
\label{sec:ar_plm_results}

\subsubsection{Do Latent Languages Scale Better?}
\label{sec:scaling_results}
We compare validation cross-entropy against estimated downstream autoregressive training compute for decoder-only transformers sharing the same architecture grid and next-token objective.
The amino acid model and PLLM use the same UniRef50 sequences, token budget, and optimization settings.
Their fitted exponents are $\alpha=0.020$ and $\alpha=0.038$, respectively (\cref{fig:scaling_loss_vs_compute}).
PLLM therefore has a roughly \num{1.9}$\times$ steeper fitted slope under matched downstream sequence training.
SLLM has a fitted exponent of $\alpha=0.049$ on structure-token data, with different training exposure.
We compare fitted slopes rather than treating absolute cross-entropy across vocabularies as a common quality scale.
The SLLM exponent is also numerically close to the exponent reported for text language models against minimum training compute adjusted for batch efficiency~\cite{kaplan2020}, although this comparison is qualitative because the corpora, tokenizers, compute accounting, and frontier construction differ.
The fitting protocol, architecture grid, and training settings are given in \cref{app:scaling_law_estimation}.

\subsubsection{PLL De Novo Generation}
\label{sec:pll_de_novo_results}
We first compare unconditional generation from the 1.3B amino acid model and the 1.3B PLLM in the same decoded amino acid space.
Direct amino acid generation shows substantial drift, defined by the entropy heuristic in \cref{app:sequence_de_novo_eval}, visible as low sequence entropy from repeated residue patterns.
This is most severe at low temperatures and often coincides with weak stopping behavior, where samples continue until the maximum context window instead of ending naturally with a \texttt{<EOS>} token.
Across the \num{12}-temperature unconditional sweep, \num{5154}/\num{12000} amino acid samples (43.0\%) and \num{2383}/\num{12000} PLLM samples (19.9\%) have decoded residue-composition entropy below 1.5 bits. PLLM therefore substantially reduces this diagnostic failure mode, although low-complexity samples remain. These counts describe unfiltered unconditional generations, separately from the length-conditioned and filtered folding evaluations below.
As temperature increases, decoded PLLM samples move toward the UniRef50 source distribution, with $T=0.6$ to $T=1.0$ giving the strongest joint match between entropy and length.

\begin{figure}[h]
    \centering
    \begin{minipage}{0.32\linewidth}
        \centering
        \includegraphics[width=\linewidth]{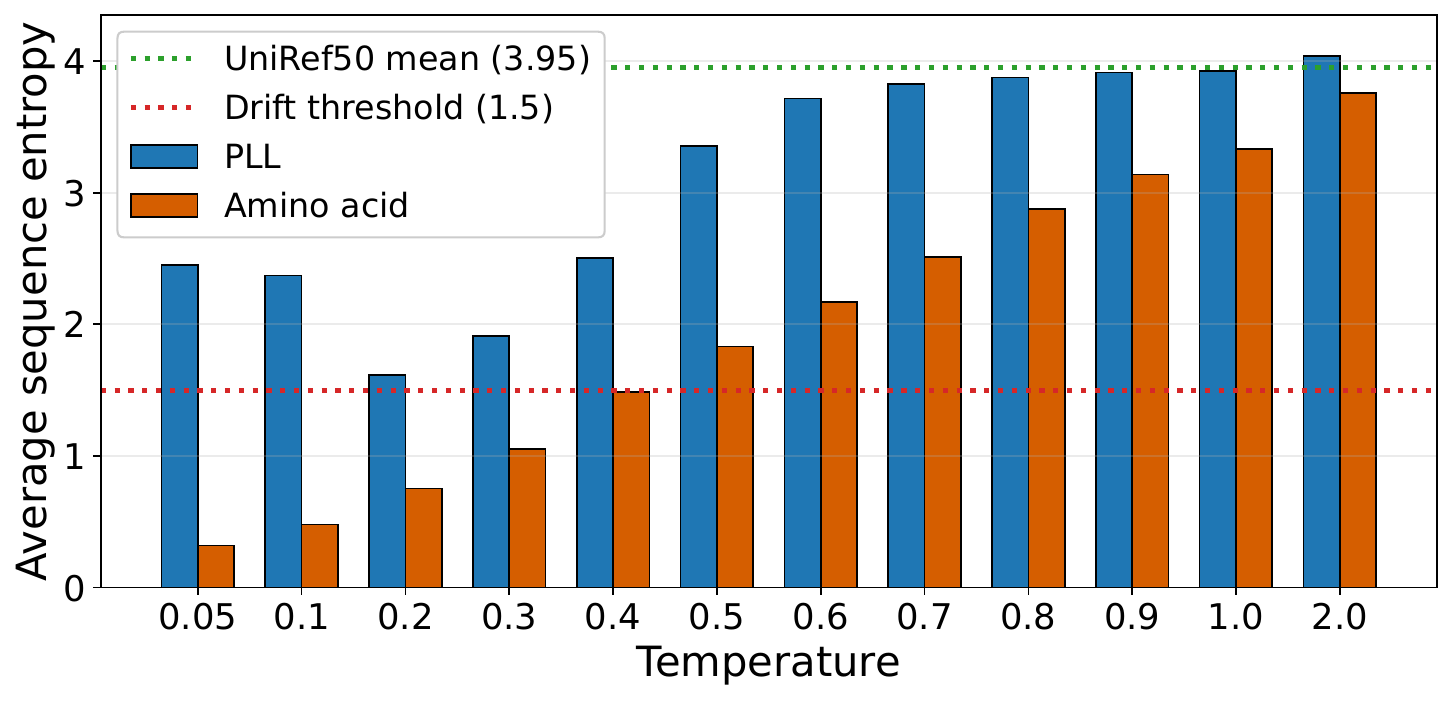}
        \footnotesize (a)
    \end{minipage}
    \hfill
    \begin{minipage}{0.32\linewidth}
        \centering
        \includegraphics[width=\linewidth]{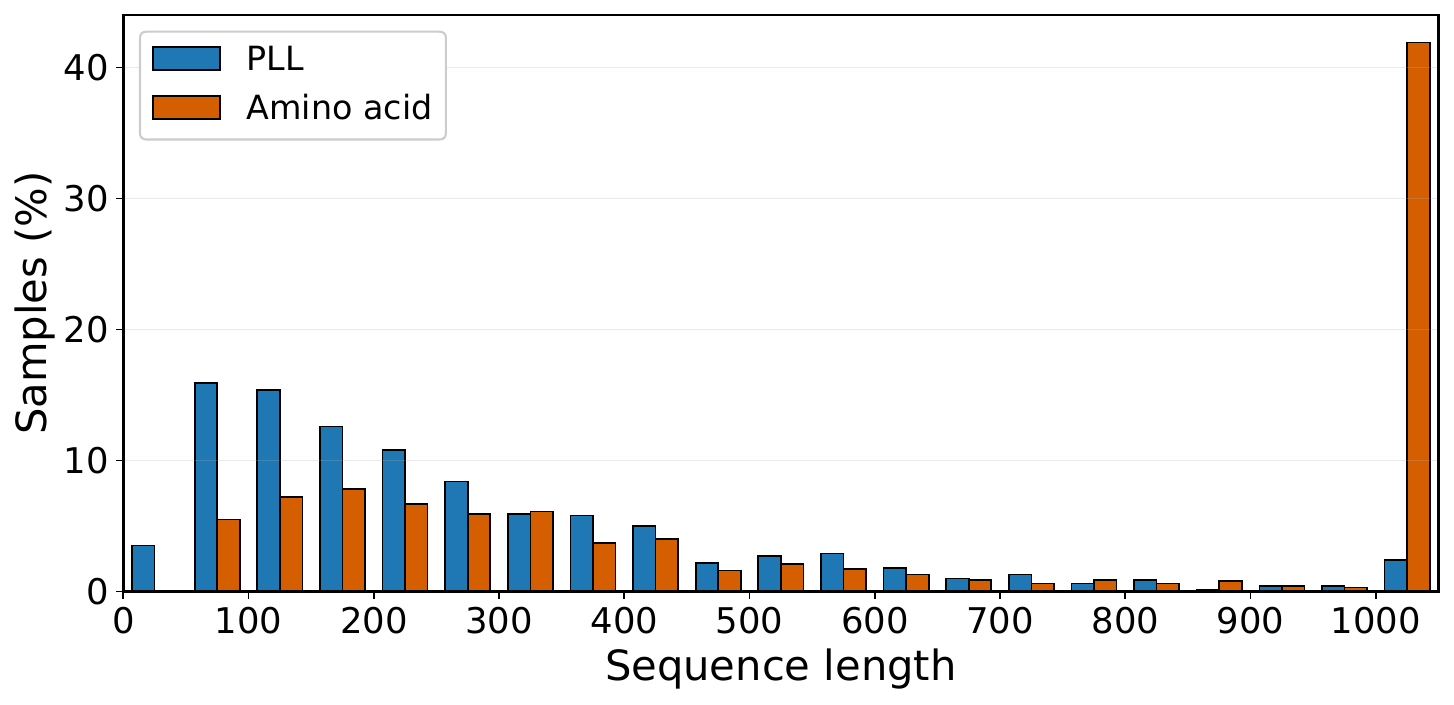}
        \footnotesize (b)
    \end{minipage}
    \hfill
    \begin{minipage}{0.32\linewidth}
        \centering
        \includegraphics[width=\linewidth]{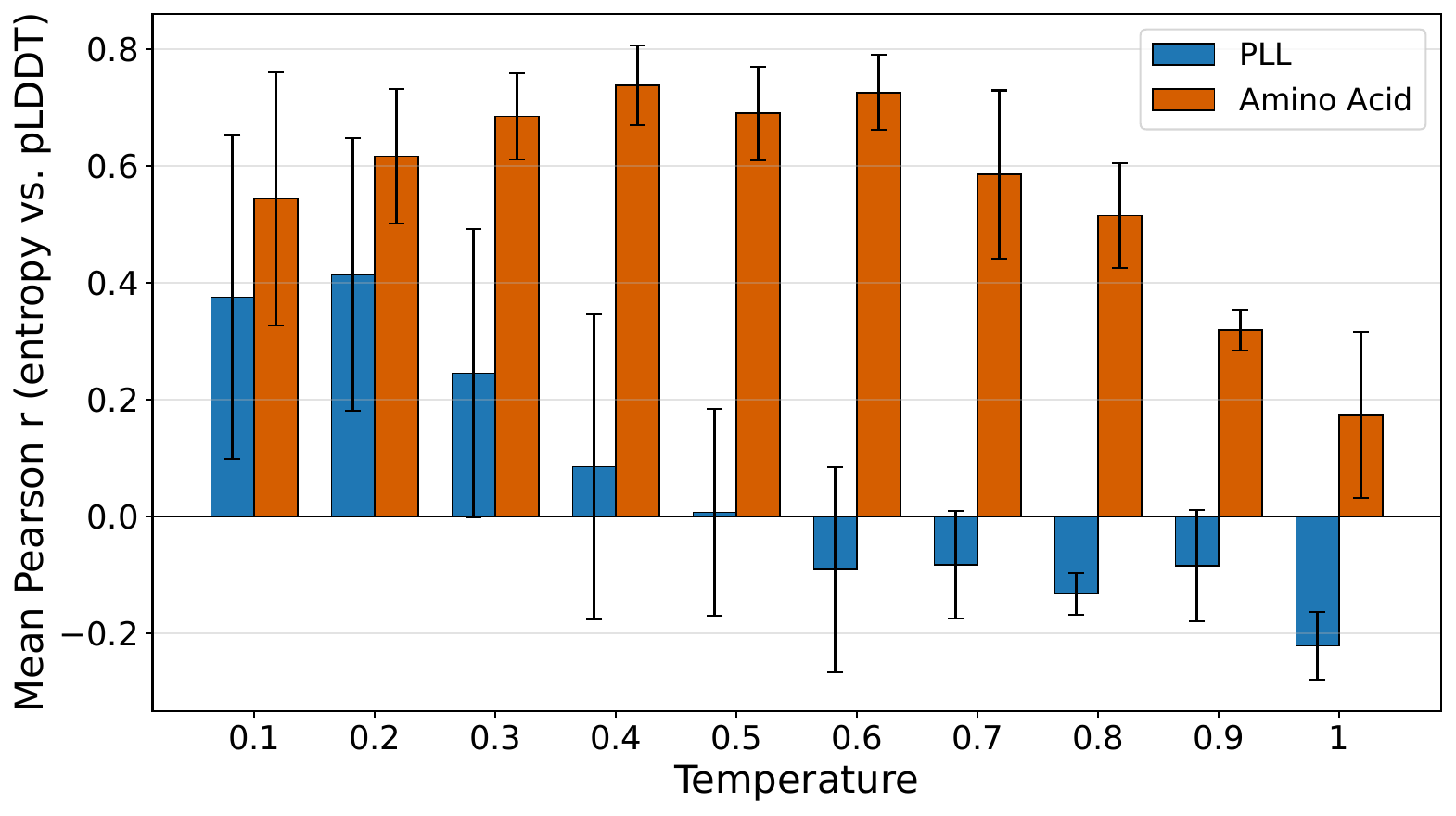}
        \footnotesize (c)
    \end{minipage}
    \caption{Amino acid versus PLLM sequence-generation diagnostics. Left: mean decoded residue-composition entropy across unconditional sampling temperatures. Middle: unconditional length distribution at $T=1.0$. Right: length-conditioned entropy-pLDDT correlation after entropy-filtered resampling.}
    \label{fig:pll_unconditional_de_novo}
\end{figure}
\FloatBarrier

For length-conditioned generation, we continue fine-tuning the largest amino acid model and PLLM from their eight-epoch checkpoints.
The length prefix is applied to 50\% of continuation samples, so each checkpoint retains unconditional sampling while learning to generate at requested lengths; details are in \cref{app:pll_conditional_generation}.
Length conditioning gives both models strong target-length control, but amino acid composition entropy remains lower. This supports the conclusion that stopping and length behavior alone do not explain the complexity difference. Entropy-pLDDT correlations are reported separately for the entropy-filtered length-conditioned samples.

\begin{table}[t]
    \centering
    \begin{minipage}[t]{0.48\linewidth}
        \centering
        \scriptsize
        \setlength{\tabcolsep}{1.8pt}
        \caption{Length-conditioned sequence ProteinBench~\cite{proteinbench}. Native, ProGen2, DPLM, and ESM3 are published reference values. PLM is our amino acid model. Our PLM and PLLM use entropy-filtered samples evaluated with ESMFold.}
        \label{tab:pll_conditional_proteinbench_t05}
        \begin{tabular}{lcccc}
            \toprule
            & Quality & \multicolumn{2}{c}{Diversity} & Novelty \\
            \cmidrule(lr){2-2}\cmidrule(lr){3-4}\cmidrule(lr){5-5}
            Model & pLDDT $\uparrow$ & Pair. TM $\downarrow$ & Clust. $\uparrow$ & Max TM $\downarrow$ \\
            \midrule
            \textcolor{gray!70}{Native} & \textcolor{gray!70}{63.70} & \textcolor{gray!70}{0.5150} & \textcolor{gray!70}{0.7900} & \textcolor{gray!70}{--} \\
            ProGen2\raisebox{0.55ex}[0pt][0pt]{\scriptsize$\ddagger$}~\cite{progen2} & \underline{65.26} & 0.3900 & 0.9325 & 0.6675 \\
            DPLM~\cite{dplm} & \textbf{89.93} & 0.5100 & 0.7450 & 0.8525 \\
            ESM3\raisebox{0.55ex}[0pt][0pt]{\scriptsize$\ddagger$}~\cite{esm3} & 53.24 & 0.3550 & \textbf{0.9750} & 0.6925 \\
            \textbf{PLLM} & 59.86 & \underline{0.2567} & \underline{0.9700} & \underline{0.6368} \\
            \textbf{PLM}\raisebox{0.55ex}[0pt][0pt]{\scriptsize$\dagger$} & \underline{65.26} & \textbf{0.2413} & 0.9250 & \textbf{0.5859} \\
            \bottomrule
        \end{tabular}
        \vspace{0.15em}
        \begin{minipage}{\linewidth}
            \tiny
            Unweighted means over lengths 100, 200, 300, and 500.
            Native, ProGen2, DPLM, and ESM3 values are taken from ProteinBench~\cite{proteinbench}.
            Our PLM and PLLM use $T=0.5$, with \num{50} retained entropy-filtered samples per length evaluated using ESMFold.
            Length-specific results are reported in \cref{tab:sequence_proteinbench_by_length}.
            $^\dagger$ PLM uses additional sampling when needed to fill the entropy-filtered evaluation set. $^\ddagger$ ProGen2 and ESM3 use roughly \num{15}$\times$ and \num{48}$\times$ more protein samples than UniRef50.
        \end{minipage}
    \end{minipage}
    \hfill
    \begin{minipage}[t]{0.48\linewidth}
        \centering
        \scriptsize
        \setlength{\tabcolsep}{1.8pt}
        \caption{Structure-design results on ProteinBench~\cite{proteinbench}, averaged over target lengths 50, 100, 300, and 500.}
        \label{tab:sll_conditional_proteinbench}
        \resizebox{\linewidth}{!}{%
        \begin{tabular}{lccccc}
            \toprule
            & \multicolumn{2}{c}{Quality} & \multicolumn{2}{c}{Diversity} & Novelty \\
            \cmidrule(lr){2-3}\cmidrule(lr){4-5}\cmidrule(lr){6-6}
            Model & scTM $\uparrow$ & scRMSD $\downarrow$ & \shortstack{TM similarity\\$\downarrow$} & Clust. $\uparrow$ & Max TM $\downarrow$ \\
            \midrule
            \textcolor{gray!70}{Native$^\ast$} & \textcolor{gray!70}{0.9525} & \textcolor{gray!70}{0.8250} & \textcolor{gray!70}{0.2900} & \textcolor{gray!70}{0.7500} & \textcolor{gray!70}{--} \\
            Genie$^\ast$~\cite{genie} & 0.4450 & 13.2375 & 0.2725 & 0.9650 & 0.4200 \\
            RFdiffusion$^\ast$~\cite{rfdiffusion} & 0.9200 & 1.8900 & 0.4200 & 0.6325 & 0.6675 \\
            Protpardelle~\cite{protpardelle1c} & \underline{0.9289} & \underline{1.5502} & 0.7477 & 0.2175 & 0.7966 \\
            ProteinGenerator~\cite{proteingenerator} & 0.7524 & 2.5888 & 0.6554 & 0.4175 & \underline{0.5806} \\
            PLAID~\cite{plaid} & 0.7132 & 2.8963 & 0.5903 & 0.6525 & 0.6698 \\
            La-Proteina~\cite{laproteina} & \textbf{0.9677} & \textbf{0.7871} & 0.6231 & 0.6550 & 0.7369 \\
            Pallatom~\cite{pallatom} & 0.7991 & 2.3077 & \underline{0.5277} & \underline{0.7375} & 0.6296 \\
            FoldFlow-2~\cite{foldflow2} & 0.8857 & 1.6703 & 0.5770 & 0.6700 & 0.7684 \\
            DPLM-2$^\dagger$~\cite{dplm2} & 0.8835 & 1.6511 & 0.7258 & 0.5375 & 0.9016 \\
            \textbf{SLLM} & 0.5390 & 3.3974 & \textbf{0.4558} & \textbf{0.9175} & \textbf{0.5536} \\
            \bottomrule
        \end{tabular}}
        \vspace{0.15em}
        \begin{minipage}{\linewidth}
            \tiny
            Bold and underline mark the best and second-best unstarred results.
            Diversity and novelty rankings require model-level mean scTM $>0.5$.
            $^\ast$ Results taken from ProteinBench.
            $^\dagger$ DPLM-2 jointly models amino acid sequences and structure tokens.
            We evaluate its 650M model in unconditional backbone-generation mode.
        \end{minipage}
    \end{minipage}
\end{table}

In the entropy-filtered $T=0.5$ comparison in \cref{tab:pll_conditional_proteinbench_t05}, the amino acid model has higher mean pLDDT than PLLM (65.26 versus 59.86), lower pairwise TM, and lower Max TM. PLLM has the higher cluster ratio (0.9700 versus 0.9250). Its reduced unconditional drift and these filtered folding scores describe different sample populations.

\Cref{tab:sequence_proteinbench_by_length} reports the corresponding values at each target length.
Additional unconditional and length-conditioned generation analyses are provided in \cref{app:pll_unconditional_generation,app:pll_conditional_generation}.

\subsubsection{SLL De Novo Generation}
\label{sec:sll_de_novo_results}

Unlike PLL, SLL does not have a raw-token counterpart that decodes to the same structure space.
We therefore evaluate length-conditioned SLLM generation directly.
Starting from the largest pretrained SLLM, we continue training for two epochs with length prefixes on 50\% of samples, matching the conditioning interface in \cref{app:ar_tokenization_protocol}.
The generated SLL streams are decoded to backbone coordinates and evaluated with the structure-design ProteinBench~\cite{proteinbench} protocol.
La-Proteina~\cite{laproteina} achieves the highest mean scTM and lowest mean scRMSD (\cref{tab:sll_conditional_proteinbench}).
SLLM generates diverse and novel backbones, with the highest cluster ratio and lowest Max TM among generators with mean scTM $>0.5$.
Its self-consistency quality remains below that of specialized backbone generators.
These results compare generation performance across pretrained systems whose training data, model sizes, and training modalities are not matched.
Length-specific results are reported in \cref{tab:structure_proteinbench_by_length,tab:structure_proteinbench_by_length_long}.

\subsubsection{Sequence-to-Structure Prediction}
\label{sec:s2s_results}
We fine-tune the pretrained length-conditioned SLLM for sequence-to-structure prediction by connecting the E1 sequence encoder shown in \cref{fig:sequence_to_structure}.
The evaluated checkpoint sees \num{4.32}B structure-token targets sampled from the same AF2-generated source used for SLL pretraining.
At inference time, we sample multiple SLL structure-token trajectories per input sequence and use internal model confidence before coordinate decoding as an inference-time scaling signal.

\begin{figure}[!ht]
    \centering
    \begin{minipage}{0.48\linewidth}
        \centering
        \includegraphics[width=\linewidth]{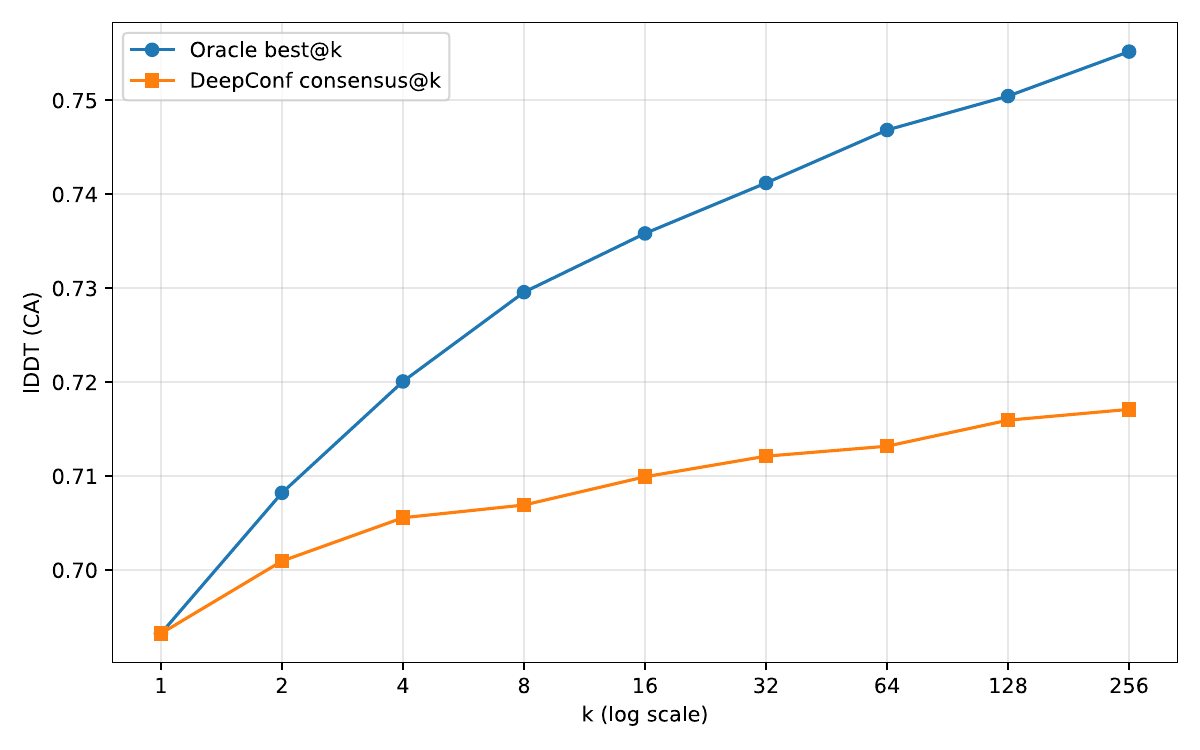}\\
        \footnotesize (a) lDDT-CA
    \end{minipage}\hfill
    \begin{minipage}{0.48\linewidth}
        \centering
        \includegraphics[width=\linewidth]{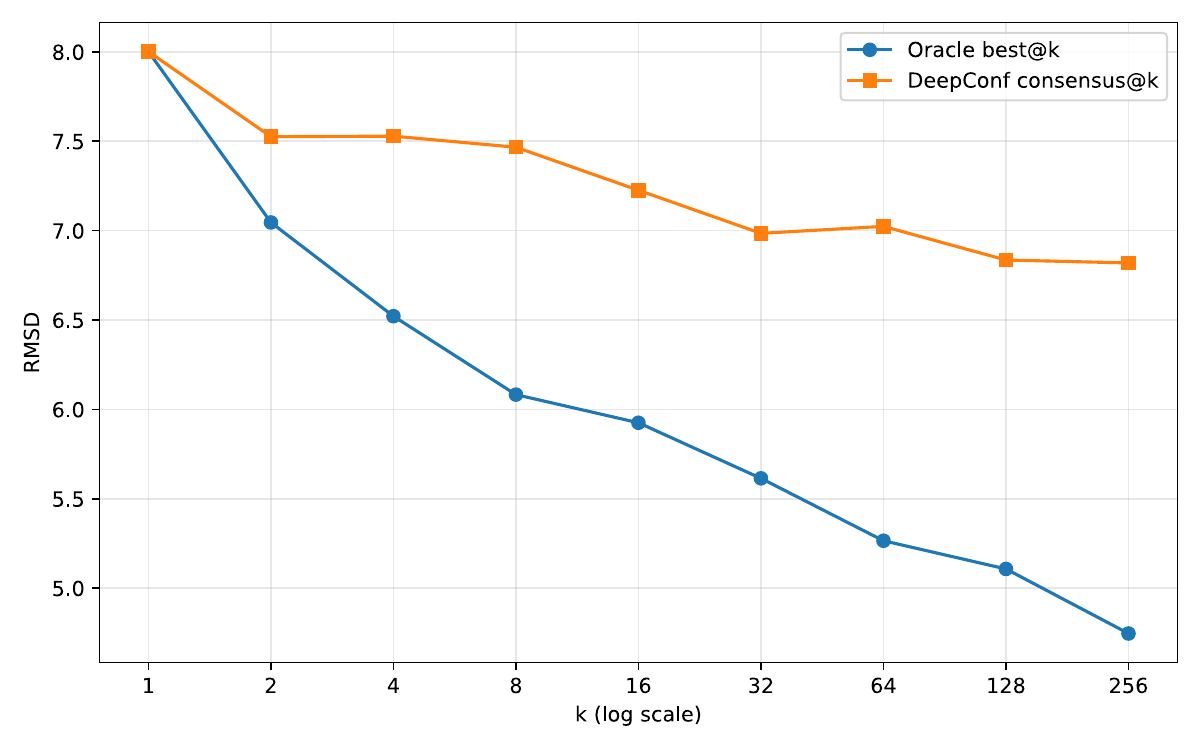}\\
        \footnotesize (b) RMSD
    \end{minipage}
    \caption{Sequence-to-structure inference-time scaling on the held-out CAMEO-2024 set as a function of the number of sampled SLL trajectories per input sequence ($k$). Oracle best@$k$ is a ceiling that picks the best candidate after coordinate decoding, while DeepConf consensus@$k$ uses only internal SLL-token confidence and token-space consensus to pick a single candidate before any coordinate decoding. (a) lDDT-CA $\uparrow$. (b) RMSD $\downarrow$. Both metrics improve overall as the sampling budget increases, and DeepConf recovers a meaningful fraction of the oracle gain. More details are given in \cref{app:s2s_model_selection}.}
    \label{fig:s2s_cameo_deepconf_main}
\end{figure}

We use a tuned DeepConf-style selector to test whether internal SLL-token confidence can identify better candidates before coordinate decoding.
The resulting CAMEO-2024 curves are shown in \cref{fig:s2s_cameo_deepconf_main}.
Full selector settings, oracle comparisons, and additional metrics are provided in \cref{app:s2s_model_selection}.

\section{Discussion}
\label{sec:discussion}
The results support learned target representations as a useful way to improve autoregressive protein modeling.
In the matched sequence comparison, PLLM shows steeper fitted compute scaling and substantially less low-complexity drift than amino acid modeling.
SLL provides a complementary structure-side study, including improved target predictability in the controlled tokenizer comparison and a separate evaluation of backbone generation.

The backbone-generation results show a tradeoff between self-consistency quality, cluster diversity, and novelty relative to PDB references.
La-Proteina leads the self-consistency quality metrics.
SLLM's strengths are diversity and novelty, while self-consistency quality remains a limitation.

PLL and SLL play different but complementary roles in this picture.
PLLM addresses a major drawback of amino acid language models that common benchmarks can obscure.
Direct residue-level generation often drifts into repetitive low-entropy patterns and weak \texttt{<EOS>} behavior, even when post-filtered samples can still score well on downstream folding diagnostics.
Moving generation into PLL substantially reduces low-complexity drift in the unconditional sweep and better matches UniRef50 residue-composition entropy and length statistics at moderate temperatures. In the entropy-filtered length-conditioned comparison, the amino acid model has higher pLDDT, while PLLM has a higher cluster ratio.
SLL shows that a structure tokenizer should not be judged only as a coordinate codec.
By adding sequence, confidence, and representation supervision to the GCP-VQVAE Lite interface, SLL gives a small amount of reconstruction quality back in exchange for substantially stronger supervised PST performance and better autoregressive compatibility.
Together, PLL and SLL frame latent protein languages as generative representations rather than compression artifacts.

The sequence-to-structure experiments highlight another consequence of this framing.
In our measurements, SLLM generates structure-token candidates for long proteins approximately \num{1000} times faster than MSA-based AlphaFold2~\cite{alphafold}, making candidate sampling and selection practical.
The oracle best@k curves show that better structures are present in the sampled SLL candidate pool as the number of samples increases.
The DeepConf-style selector recovers part of this gain using only internal latent-token confidence and token-space consensus, making confidence before coordinate decoding a useful signal for protein structure generation.

Several limitations remain.
The current models are not the strongest methods on quality-centered metrics, especially for de-novo backbone generation.
While we apply SLLM in a cross-modal setting for sequence-to-structure prediction, each of our pretrained autoregressive transformers operates on a single token stream, and we do not evaluate a unified multimodal pretraining setup that jointly models sequence and structure tokens for generation.
Joint PLL--SLL co-generation and compatibility between the independently learned languages on joint tasks remain untested.
Future work should address both directions.

PLL is designed to represent the contextual protein information learned by ESM-2 as a discrete language that can be decoded to amino acid sequences. Our sequence comparison evaluates this complete target-language design. We did not compare it with continuous-target generation or similarly resourced ESM-2 fine-tuning and distillation approaches. Such comparisons would assess the generation performance of the complete PLL approach relative to alternative uses of pretrained protein representations.

Extending SLL to multichain complexes would require modeling relative chain poses, which we have not evaluated here.

\section{Conclusion}
\label{sec:conclusion}
We presented two learned latent protein languages, PLL on the sequence side and SLL on the structure side, and used them as drop-in target vocabularies for standard decoder-only transformers.
Under matched downstream sequence training, PLLM shows steeper fitted compute-scaling trends and substantially less low-complexity drift than amino acid modeling.
SLL improves autoregressive target predictability in the controlled tokenizer comparison and supports structure generation and sequence-to-structure prediction.
We see this as early evidence that learned latent protein languages are a promising direction for autoregressive recipes, and we hope it motivates further work on using this approach for protein sequence and structure generation.

\enlargethispage{\baselineskip}
\section*{Acknowledgments}
We gratefully acknowledge NVIDIA and the National Artificial Intelligence Research Resource (NAIRR) Pilot program for providing the computational resources that made this work possible, under project \texttt{NAIRR240246}.

\bibliography{references}


\appendix
\newpage

\section*{Appendix}
\label{app:appendix}

{\large\scshape Table of Contents\par}
\vspace{1em}
\appcontentssection{A}{Methods}{app:method_details}
\appcontentssubsection{A.1}{PLL Tokenizer Details}{app:pll_architecture}
\appcontentssubsection{A.2}{PLL Metrics}{app:pll_metrics}
\appcontentssubsection{A.3}{SLL Tokenizer Details}{app:sll_architecture}
\appcontentssubsection{A.4}{SLL Metrics}{app:sll_metrics}
\appcontentssubsection{A.5}{Autoregressive Tokenization Protocol}{app:ar_tokenization_protocol}
\appcontentssubsection{A.6}{Confidence-Based Candidate Selection}{app:confidence_selection_details}
\appcontentssubsection{A.7}{Scaling-Law Estimation}{app:scaling_law_estimation}
\appcontentssubsection{A.8}{Sequence-Side De Novo Evaluation}{app:sequence_de_novo_eval}
\appcontentssubsection{A.9}{Structure-Side De Novo Evaluation}{app:structure_de_novo_eval}
\vspace{0.75em}
\appcontentssection{B}{Data Sources}{app:data_sources}
\vspace{0.75em}
\appcontentssection{C}{Additional Experimental Results}{app:additional_results}
\appcontentssubsection{C.1}{PLL Per-Amino-Acid Recovery}{app:pll_per_aa_recovery}
\appcontentssubsection{C.2}{SLL}{app:sll_additional_results}
\appcontentssubsubsection{C.2.1}{SLL Supervised PST Ablation}{app:sll_supervised_pst_ablation}
\appcontentssubsubsection{C.2.2}{SLL Supervised PST Full-Run}{app:sll_supervised_pst_full_run}
\appcontentssubsubsection{C.2.3}{SLL Reconstruction Comparison}{app:sll_reconstruction_comparison}
\appcontentssubsubsection{C.2.4}{SLL Unsupervised PST Full-Run}{app:sll_unsupervised_pst_full_run}
\appcontentssubsubsection{C.2.5}{SLL Sequence-to-Structure Compatibility}{app:sll_s2s_tokenizer_ablation}
\appcontentssubsection{C.3}{Autoregressive Protein Language Models}{app:ar_plm_additional_results}
\appcontentssubsubsection{C.3.1}{PLLM Unconditional Generation}{app:pll_unconditional_generation}
\appcontentssubsubsection{C.3.2}{PLLM Conditional Generation}{app:pll_conditional_generation}
\appcontentssubsubsection{C.3.3}{SLLM Length-Conditioned Continuation}{app:sll_length_conditioned_continuation}
\appcontentssubsubsection{C.3.4}{SLLM Sequence-to-Structure Model Selection}{app:s2s_model_selection}
\newpage

\section{Methods}
\label{app:method_details}

\subsection{PLL Tokenizer Details}
\label{app:pll_architecture}
PLL Stage 1 uses frozen ESM-2 representations as the target semantic space.
For a valid residue position $t$, ESM-2 produces $\mathbf h_t$.
A trainable linear projection, transformer encoder, and projection head produce the vector that enters the vector-quantization layer,
\begin{equation}
    \mathbf r_{1:L}=E_\theta(W_{\mathrm{in}}\mathbf h_{1:L}),
    \qquad
    \mathbf u_t=W_{\mathrm{vq}}\mathbf r_t .
\end{equation}
The vector-quantization layer uses a single Euclidean codebook with one code per residue position.
It computes the distance scores in \cref{eq:pll_quantization} only on valid residue positions.
Padding positions are masked and do not contribute to assignment statistics or losses.

During Stage 1 training, PLL uses hard stochastic code assignment from the same scores,
\begin{equation}
    z_t^{\mathrm{PLL}}
    =
    \arg\max_k
    \left(
    \frac{\ell_{t,k}}{\tau}
    + g_{t,k}
    \right),
    \qquad
    g_{t,k}\sim \mathrm{Gumbel}(0,1),
    \qquad
    \mathbf q_t=\mathbf c_{z_t^{\mathrm{PLL}}}.
\end{equation}
The temperature is fixed to $\tau=0.1$ in Stage 1.
At evaluation, decoding, and Stage 2 tokenization, the Gumbel noise is removed and the assignment reduces to the deterministic nearest-code rule in \cref{eq:pll_quantization}.

The codebook is initialized by k-means and updated by exponential moving-average (EMA) statistics rather than direct gradient descent.
Let $\rho$ denote the EMA decay, and let $m_{t,k}$ be the one-hot assignment for valid position $t$.
For each code $k$, EMA updates assignment counts and embedding sums as
\begin{equation}
    n_k \leftarrow \rho n_k + (1-\rho)\sum_t m_{t,k},
    \qquad
    \mathbf e_k \leftarrow \rho \mathbf e_k + (1-\rho)\sum_t m_{t,k}\mathbf u_t,
    \qquad
    \mathbf c_k \leftarrow \frac{\mathbf e_k}{\tilde n_k},
\end{equation}
where $\tilde n_k$ is the EMA-smoothed assignment count.
In the trained PLL tokenizer, $\rho=0.99$.
Rare codes below the dead-code threshold are replaced from current batch samples, which helps maintain codebook coverage.
For the final tokenizer, the VQ layer uses k-means initialization with 10 iterations, EMA decay 0.99, and a dead-code threshold of two.
Stage 1 uses stochastic code assignment with temperature 0.1.
Stage 2 uses deterministic assignments from the frozen Stage 1 codebook.

The vector-quantization contribution to the Stage 1 objective is the commitment loss under the hard assignment above, while codebook locations are maintained by k-means initialization with 10 iterations, EMA updates with decay 0.99, and dead-code replacement with threshold two.

For valid residue positions $T$, the Stage 1 terms in \cref{eq:pll_stage1_loss} are
\begin{equation}
    \label{eq:app_pll_stage1_terms}
    \begin{aligned}
    \mathcal{L}_{\mathrm{mse}}
    &=
    \frac{1}{|T|}\sum_{t\in T}\|\hat{\mathbf h}_t-\mathbf h_t\|_2^2, \\
    \mathcal{L}_{\mathrm{cos}}
    &=
    \frac{1}{|T|}\sum_{t\in T}
    \left(1-\frac{\langle \hat{\mathbf h}_t,\mathbf h_t\rangle}{\|\hat{\mathbf h}_t\|_2\|\mathbf h_t\|_2}\right), \\
    \mathcal{L}_{\mathrm{commit}}
    &=
    \frac{1}{|T|}\sum_{t\in T}\|\mathbf u_t-\operatorname{sg}(\mathbf q_t)\|_2^2 .
    \end{aligned}
\end{equation}
where $\operatorname{sg}(\cdot)$ stops gradients through the selected code vector.
Thus, $\mathcal{L}_{\mathrm{rec}}=10\,\mathcal{L}_{\mathrm{mse}}+1\,\mathcal{L}_{\mathrm{cos}}$ and $\mathcal{L}_{\mathrm{vq}}=0.5\,\mathcal{L}_{\mathrm{commit}}$ in \cref{eq:pll_stage1_loss}.
The MSE and cosine terms reconstruct the frozen ESM-2 representation, while the commitment term trains the encoder outputs to stay close to their assigned code vectors.

PLL Stage 1 keeps ESM-2 150M fixed and learns a symmetric latent encoder-decoder around a residue-wise VQ codebook.
The latent encoder and decoder each use 12 transformer layers, width 768, 12 attention heads, 64-dimensional attention heads, three key-value heads, and a feedforward multiplier of four.
The VQ codebook contains 4096 codes of dimension 128.
In Stage 2, the Stage 1 decoder weights are discarded and the transformer decoder is reinitialized for residue classification.
The fixed 128-dimensional quantized vectors are projected to 768 dimensions, processed by the 12-layer transformer decoder, and mapped to 21 residue classes by a linear output head.
Stage 1 trains the latent encoder, VQ layer, and latent decoder.
Stage 2 freezes the latent encoder and VQ layer and trains the reinitialized transformer decoder, its input projection, and the residue-classification head.

\begin{table}[h]
    \centering
    \scriptsize
    \caption{PLL training choices for the two tokenizer stages. Stage 1 learns the latent bottleneck, and Stage 2 freezes the latent encoder and codebook and trains the transformer decoder for residue recovery. Maximum sequence length includes the beginning-of-sequence and end-of-sequence tokens from the ESM tokenizer.}
    \label{tab:app_pll_training_choices}
    \begin{tabular}{
        >{\raggedright\arraybackslash}p{0.25\linewidth}
        >{\raggedright\arraybackslash}p{0.32\linewidth}
        >{\raggedright\arraybackslash}p{0.32\linewidth}}
        \toprule
        Setting & Stage 1 & Stage 2 \\
        \midrule
        Objective &
        $\mathcal{L}_{\mathrm{mse}}$, $\mathcal{L}_{\mathrm{cos}}$, $\mathcal{L}_{\mathrm{commit}}$ &
        $\mathcal{L}_{\mathrm{class}}$ \\
        Epochs &
        \num{64} &
        \num{8} \\
        Maximum sequence length &
        \num{1024} &
        \num{1024} \\
        Training size &
        \num{2.0}M &
        \num{20.0}M \\
        Batch size &
        \num{512} (\num{16} per GPU) &
        \num{2048} (\num{64} per GPU) \\
        \addlinespace[0.25em]
        Optimizer &
        AdamW, weight decay 0.01, gradient clipping \num{1.0} &
        Same as Stage 1 \\
        Learning-rate range &
        \num{1e-4} to \num{1e-6} &
        Same as Stage 1 \\
        Schedule &
        \num{4096} warmup steps, cosine decay &
        Same as Stage 1 \\
        \bottomrule
    \end{tabular}
\end{table}

\subsection{PLL Metrics}
\label{app:pll_metrics}
PLL evaluation follows the two-stage training procedure.
In Stage 1, the tokenizer is evaluated as a latent autoencoder.
A sequence is encoded by frozen ESM-2, passed through the quantized bottleneck, and decoded back to ESM-2 space.
Let $\mathbf h_{s,t}$ and $\hat{\mathbf h}_{s,t}$ denote the original and reconstructed ESM-2 vectors for residue $t$ in sequence $s$, and let $T_s$ be the valid residue positions.
The Stage 1 metrics measure how faithfully this process reconstructs the ESM-2 latent representation.

The $\Delta\mathrm{PPL}_{\%}$ metric, or percent delta perplexity, is the primary Stage 1 monitoring target.
It converts the original and reconstructed vectors into distributions and reports the average relative increase in latent perplexity after reconstruction.
Let $\mathrm{CE}^{\mathrm{base}}_s$ be the self-entropy of the original latent distribution and $\mathrm{CE}^{\mathrm{rec}}_s$ be the cross-entropy from the original distribution to the reconstructed one.
We compute
\begin{equation}
    \Delta\mathrm{PPL}_{\%}
    =
    100\left(
    \frac{1}{N}\sum_{s=1}^{N}
    \exp\!\left(\mathrm{CE}^{\mathrm{rec}}_s-\mathrm{CE}^{\mathrm{base}}_s\right)
    -1
    \right).
\end{equation}
Lower values mean that the reconstructed representation behaves more like the original ESM-2 representation under this latent-distribution comparison.

Reconstruction Kullback-Leibler (KL) divergence measures the distributional gap between the same original and reconstructed latent vectors after softmax normalization.
\begin{equation}
    \mathrm{KL}_{\mathrm{rec}}
    =
    \frac{1}{N}\sum_{s=1}^{N}
    \frac{1}{|T_s|}\sum_{t\in T_s}
    D_{\mathrm{KL}}\!\left(
    \operatorname{softmax}(\mathbf h_{s,t})
    \,\middle\|\,
    \operatorname{softmax}(\hat{\mathbf h}_{s,t})
    \right).
\end{equation}
Median normalized mean squared error (NMSE) gives the complementary vector-space view, normalizing each sequence-level squared error by the energy of the original latent representation.
\begin{equation}
    \mathrm{NMSE}_{\mathrm{median}}
    =
    \operatorname{median}_{s}
    \frac{
    \sum_{t\in T_s}\|\mathbf h_{s,t}-\hat{\mathbf h}_{s,t}\|_2^2
    }{
    \sum_{t\in T_s}\|\mathbf h_{s,t}\|_2^2
    }.
\end{equation}
The median reports typical reconstruction quality without letting unusually difficult sequences dominate the summary.

In Stage 2, the learned tokenizer is fixed and the evaluation asks whether PLL codes retain enough information to recover amino acid identities.
The residue vocabulary contains 21 classes, consisting of the 20 standard amino acids and $X$, which represents unknown or non-standard residues.
The primary metric is micro token-level amino acid recovery accuracy.
\begin{equation}
    \mathrm{Acc}_{\mathrm{micro}}
    =
    \frac{
    \sum_s \sum_{t\in T_s}
    \mathbf{1}\!\left[\arg\max_c p_\phi(c\mid z^{\mathrm{PLL}}_{s,1:L_s}, t) = x_{s,t}\right]
    }{
    \sum_s |T_s|
    } .
\end{equation}
This is the overall recovery rate over valid residues.
Padding positions are ignored, special tokens are not included, and unknown or non-standard residues are mapped to the valid class $X$.
Per-amino-acid recovery accuracy repeats the same calculation within each residue class.
\begin{equation}
    \mathrm{Acc}(a)
    =
    \frac{
    \sum_s \sum_{t\in T_s}
    \mathbf{1}[x_{s,t}=a]\,
    \mathbf{1}\!\left[\arg\max_c p_\phi(c\mid z^{\mathrm{PLL}}_{s,1:L_s}, t) = a\right]
    }{
    \sum_s \sum_{t\in T_s}\mathbf{1}[x_{s,t}=a]
    } .
\end{equation}
This class-wise view checks whether high micro accuracy is shared across residue types rather than driven only by frequent amino acids.
Stage 1 reconstruction metrics and Stage 2 residue recovery are tokenizer diagnostics. They are distinct from the downstream next-token validation perplexity reported for the SLL sequence-to-structure comparison in \cref{app:sll_s2s_tokenizer_ablation}.

\subsection{SLL Tokenizer Details}
\label{app:sll_architecture}
SLL uses the GCP-VQVAE Lite~\cite{gcp_vqvae} architecture as its backbone tokenizer.
The input is a protein backbone represented by residue-level geometry.
The GCPNet encoder uses chirality-sensitive message passing with rotation- and translation-invariant scalar features and rotation-equivariant vector features~\cite{gcp_vqvae}.
Its rotation- and translation-invariant scalar residue readouts are contextualized by a bidirectional transformer encoder before quantization.
The quantizer assigns one discrete SLL code per valid residue position from a codebook of size 4096.
Padding and invalid coordinate positions are masked before loss and metric computation.

The decoder receives the quantized residue sequence and reconstructs the three-atom backbone coordinates $(N, C_\alpha, C)$ through a transformer decoder and six-dimensional rotation head.
The final reconstruction objective contains an aligned coordinate term and a pairwise backbone-distance term.
For the aligned coordinate term, target coordinates are Kabsch-aligned to the predicted coordinates over valid residues before computing elementwise mean squared error.
For the distance term, pairwise distances are computed over valid backbone atoms and compared between the predicted and target structures.
The vector-quantization loss keeps encoder outputs close to their assigned code vectors.
The coordinate objective compares backbones up to a global rigid transformation, while the distance objective is invariant to global translation and rotation.
Vector quantization can still introduce internal geometric reconstruction error.
For a structure $s$, let $T_s$ be the valid residue positions and let $\mathcal A_s$ be the corresponding valid backbone atoms.
Let $\hat{\mathbf y}_{s,t,a}$ be the predicted coordinate for atom $a$ at residue $t$, and let $\tilde{\mathbf y}_{s,t,a}$ be the Kabsch-aligned target coordinate.
Let $\hat{\mathcal V}_s=\{\hat{\mathbf v}_{s,m}\}_{m=1}^{6|T_s|}$ and $\tilde{\mathcal V}_s=\{\tilde{\mathbf v}_{s,m}\}_{m=1}^{6|T_s|}$ be the predicted and aligned-target backbone direction and normal vectors used by the GCP-VQVAE direction loss.
The reconstruction terms considered in our tokenizer study are defined per structure and averaged over the batch,
\begin{equation}
    \label{eq:app_sll_reconstruction_terms}
    \begin{aligned}
    \mathcal{L}_{\mathrm{coord}}
    &=
    \frac{1}{3|\mathcal A_s|}\sum_{(t,a)\in\mathcal A_s}
    \|\hat{\mathbf y}_{s,t,a}-\tilde{\mathbf y}_{s,t,a}\|_2^2, \\
    \mathcal{L}_{\mathrm{dist}}
    &=
    \frac{1}{|\mathcal A_s|^2}
    \sum_{i,j\in\mathcal A_s}
    \min\!\left(
    \left(\|\hat{\mathbf y}_{s,i}-\hat{\mathbf y}_{s,j}\|_2
    -\|\tilde{\mathbf y}_{s,i}-\tilde{\mathbf y}_{s,j}\|_2\right)^2,
    25
    \right), \\
    \mathcal{L}_{\mathrm{dir}}
    &=
    \frac{1}{(6|T_s|)^2}
    \sum_{m,n=1}^{6|T_s|}
    \min\!\left(
    \left(
    \langle \hat{\mathbf v}_{s,m},\hat{\mathbf v}_{s,n}\rangle
    -\langle \tilde{\mathbf v}_{s,m},\tilde{\mathbf v}_{s,n}\rangle
    \right)^2,
    20
    \right).
    \end{aligned}
\end{equation}
The distance term clips each squared error at 25.
The direction term compares pairwise dot products among local backbone direction and normal vectors and clips each squared error at 20.
When enabled in reconstruction-loss ablations, it enters with coefficient $0.05$.
It is not used in the final SLL tokenizer.

The SLL-specific change is the addition of semantic auxiliary heads during tokenizer training.
Two heads act before the structure decoder and therefore supervise the quantized stream directly.
The inverse-folding head predicts the 21-class residue identity vocabulary, consisting of the 20 standard amino acids and $X$.
The pLDDT head predicts the per-residue confidence target.
A third head acts after the decoder and predicts frozen ESM-2 representations.
Let $\mathbf q_{s,t}$ be the quantized vector at residue $t$, let $r_{s,t}$ be the pLDDT target, and let $\mathbf e_{s,t}$ be the frozen ESM-2 target representation.
Let $T_s^{\mathrm{pLDDT}}\subseteq T_s$ be the valid positions with observed pLDDT targets, and let $d_{\mathrm{ESM}}$ be the ESM-2 representation dimension.
The auxiliary losses are
\begin{equation}
    \label{eq:app_sll_auxiliary_terms}
    \begin{aligned}
    \mathcal{L}_{\mathrm{if}}
    &=
    \frac{1}{|T_s|}\sum_{t\in T_s}
    \operatorname{CE}\!\left(x_{s,t}, p_{\mathrm{if}}(\cdot\mid \mathbf q_{s,t})\right), \\
    \mathcal{L}_{\mathrm{pLDDT}}
    &=
    \frac{1}{|T_s^{\mathrm{pLDDT}}|}\sum_{t\in T_s^{\mathrm{pLDDT}}}
    |\hat r_{s,t}-r_{s,t}|, \\
    \mathcal{L}_{\mathrm{ESM}}
    &=
    \frac{1}{d_{\mathrm{ESM}}|T_s|}\sum_{t\in T_s}
    \|\hat{\mathbf e}_{s,t}-\mathbf e_{s,t}\|_2^2 .
    \end{aligned}
\end{equation}
These auxiliary terms are also averaged over the batch before weighting.
Using the final SLL tokenizer coefficients, the objective combines the active reconstruction, quantization, and auxiliary terms,
\begin{equation}
    \label{eq:app_sll_total_loss}
    \mathcal{L}_{\mathrm{SLL}}
    =
    0.002\,\mathcal{L}_{\mathrm{coord}}
    +0.01\,\mathcal{L}_{\mathrm{dist}}
    +\mathcal{L}_{\mathrm{vq}}
    +0.02\,\mathcal{L}_{\mathrm{if}}
    +0.1\,\mathcal{L}_{\mathrm{pLDDT}}
    +0.5\,\mathcal{L}_{\mathrm{ESM}} .
\end{equation}
The selected SLL recipe drops $\mathcal{L}_{\mathrm{dir}}$ and keeps the coordinate and pairwise-distance reconstruction losses in \cref{eq:app_sll_total_loss}.
The auxiliary heads are removed from the autoregressive generation interface.
After tokenizer training, an SLL sequence is simply a sequence of structure-code IDs, and generated SLL tokens are decoded with the fixed SLL decoder.

\begin{table}[h]
    \centering
    \scriptsize
    \caption{SLL tokenizer training choices for the final GCP-VQVAE Lite 2 model. The architecture and source dataset follow GCP-VQVAE Lite.}
    \label{tab:app_sll_training_choices}
    \begin{tabular}{
        >{\raggedright\arraybackslash}p{0.34\linewidth}
        >{\raggedright\arraybackslash}p{0.54\linewidth}}
        \toprule
        Setting & Value \\
        \midrule
        Active objective terms &
        $\mathcal{L}_{\mathrm{coord}}$, $\mathcal{L}_{\mathrm{dist}}$, $\mathcal{L}_{\mathrm{vq}}$, $\mathcal{L}_{\mathrm{if}}$, $\mathcal{L}_{\mathrm{pLDDT}}$, $\mathcal{L}_{\mathrm{ESM}}$ \\
        Epochs &
        \num{64} \\
        Maximum sequence length &
        \num{1280} residues \\
        Training size &
        \num{2.28}M \\
        Batch size &
        \num{512} (\num{16} per GPU) \\
        Precision &
        bfloat16 \\
        Optimizer &
        AdamW, weight decay 0.01, gradient clipping \num{1.0} \\
        Learning-rate range &
        \num{1e-4} to \num{1e-5} \\
        Schedule &
        \num{4096} warmup steps, cosine decay \\
        Coordinate augmentation &
        Random rigid rotations, Gaussian jitter with standard deviation \num{0.05}, and coordinate masking \\
        \bottomrule
    \end{tabular}
\end{table}

\subsection{SLL Metrics}
\label{app:sll_metrics}
This subsection defines the reconstruction metrics used to verify that SLL codes remain decodable to backbone coordinates.
Representation-quality evaluation follows the PST protocol from StructTokenBench~\cite{structtokenbench}.
The primary reconstruction metrics are TM-score and RMSD, computed after structural alignment between the reconstructed and target backbones.
TM-score measures global fold similarity with length normalization, where higher is better.
RMSD measures aligned coordinate error in angstroms, where lower is better.

During tokenizer training, we also monitor coordinate-space diagnostics on valid residues.
Mean absolute error and RMSD summarize pointwise coordinate recovery, while GDT-TS and TM-score summarize structure-level agreement.

\subsection{Autoregressive Tokenization Protocol}
\label{app:ar_tokenization_protocol}
All autoregressive experiments use residue-wise tokenization for amino acid, PLL, and SLL streams.
Each valid residue position contributes one content token in the selected language.
The token inventory used in the experiments is summarized in \cref{tab:app_ar_token_inventory}.
During cross-entropy training, targets set to \texttt{<PAD>} are ignored.
Content tokens and boundary tokens are predicted normally, while condition-prefix targets such as \texttt{<LEN\_n>} and \texttt{<sequence\_to\_structure>} are excluded from the loss.

\begin{table}[h]
    \centering
    \scriptsize
    \caption{Token inventory used by the autoregressive tokenizer.}
    \label{tab:app_ar_token_inventory}
    \begin{tabular}{
        >{\raggedright\arraybackslash}p{0.30\linewidth}
        >{\raggedright\arraybackslash}p{0.58\linewidth}}
        \toprule
        Type & Tokens \\
        \midrule
        Special & \texttt{<PAD>}, \texttt{<BOS>}, \texttt{<EOS>}, \texttt{<MASK>}, \texttt{<UNK>} \\
        Amino acid & A C D E F G H I K L M N P Q R S T V W Y X \\
        PLL & learned code IDs $0$ to $\num{4095}$ \\
        SLL & learned structure code IDs $0$ to $\num{4095}$ \\
        Length condition & \texttt{<LEN\_n>} \\
        Protein boundary & \texttt{<BOP>}, \texttt{<EOP>} \\
        Structure boundary & \texttt{<BO3D>}, \texttt{<EO3D>} \\
        Sequence-to-structure task & \texttt{<sequence\_to\_structure>} \\
        \bottomrule
    \end{tabular}
\end{table}

Examples of the resulting streams are shown in \cref{tab:app_ar_stream_examples}.
Unconditional tokenization uses \texttt{<BOS>} content \texttt{<EOS>} for amino acid, PLL, and SLL language modeling.
Conditional tokenization adds prefix or boundary tokens before the generated content.
Length-conditioned continuation uses \texttt{<LEN\_n>} as an instruction, where $n$ is the content length after truncation.
For sequence-to-structure, the amino acid sequence is encoded outside the causal token stream by the E1 protein encoder, and the decoder predicts SLL tokens autoregressively.
All streams are padded to the model context length, and padded targets do not contribute to loss or perplexity.

\begin{table}[h]
    \centering
    \scriptsize
    \caption{Example token streams for the autoregressive experiments. $P_i$ denotes a PLL code token, and $S_i$ denotes an SLL code token.}
    \label{tab:app_ar_stream_examples}
    \begin{tabular}{
        >{\raggedright\arraybackslash}p{0.26\linewidth}
        >{\raggedright\arraybackslash}p{0.62\linewidth}}
        \toprule
        Case & Stream format \\
        \midrule
        Amino acid &
        \texttt{<BOS>} A C D $\ldots$ Y \texttt{<EOS>} \\
        PLL &
        \texttt{<BOS>} $P_1$ $P_2$ $\ldots$ $P_L$ \texttt{<EOS>} \\
        SLL &
        \texttt{<BOS>} $S_1$ $S_2$ $\ldots$ $S_L$ \texttt{<EOS>} \\
        Amino acid length condition &
        \texttt{<BOS>} \texttt{<LEN\_n>} A C D $\ldots$ Y \texttt{<EOS>} \\
        PLL length condition &
        \texttt{<BOS>} \texttt{<LEN\_n>} \texttt{<BOP>} $P_1$ $\ldots$ $P_L$ \texttt{<EOP>} \texttt{<EOS>} \\
        SLL length condition &
        \texttt{<BOS>} \texttt{<LEN\_n>} \texttt{<BO3D>} $S_1$ $\ldots$ $S_L$ \texttt{<EO3D>} \texttt{<EOS>} \\
        Sequence-to-structure &
        \texttt{<BOS>} \texttt{<sequence\_to\_structure>} \texttt{<BO3D>} $S_1$ $\ldots$ $S_L$ \texttt{<EO3D>} \texttt{<EOS>} \\
        \bottomrule
    \end{tabular}
\end{table}

\subsection{Confidence-Based Candidate Selection}
\label{app:confidence_selection_details}
DeepConf~\cite{deepconf} uses model-internal token statistics to score multiple generated traces and filter low-confidence outputs.
We adapt this idea to sequence-to-structure generation in an offline setting, where all candidate SLL streams for an input sequence are generated before selection.
This differs from online DeepConf because we do not stop generation early, and we do not vote over final answers.
The selector chooses one structure-token sequence directly.

For an input sequence $x$, let $\mathbf z^{(j)}=(z^{(j)}_1,\ldots,z^{(j)}_{L_j})$ be candidate $j$ among $k$ sampled SLL streams.
At generated structure-token position $t$, the SLLM defines
$p_\theta(\cdot \mid x,\mathbf z^{(j)}_{<t})$ over the SLL vocabulary.
We measure internal uncertainty before coordinate decoding by the token entropy of the raw model distribution,
\begin{equation}
    H^{(j)}_t
    =
    -\sum_{v\in\mathcal V_{\mathrm{SLL}}}
    p_\theta(v\mid x,\mathbf z^{(j)}_{<t})
    \log p_\theta(v\mid x,\mathbf z^{(j)}_{<t}) .
\end{equation}
Lower entropy indicates higher model confidence in the latent structure-token prediction.
The confidence score is computed from the full vocabulary distribution before top-$p$ filtering and before any temperature rescaling used for sampling.

Following the local-confidence view in DeepConf, we summarize candidate quality with local windows rather than only a full-sequence average.
For a window $G_r$ over generated positions, define
\begin{equation}
    \bar H^{(j)}_r
    =
    \frac{1}{|G_r|}
    \sum_{t\in G_r} H^{(j)}_t .
\end{equation}
The main selector scores a candidate by its least confident local region,
\begin{equation}
    s^{(j)}_{\mathrm{conf}}
    =
    \max_r \bar H^{(j)}_r ,
\end{equation}
and lower scores are preferred.
This is the entropy-space analog of DeepConf's lowest group confidence.
We also use mean, tail, and bottom-percent group entropy as diagnostic variants during selection analysis.

When consensus selection is enabled, we first retain the most confident candidates under $s^{(j)}_{\mathrm{conf}}$.
If only one candidate is retained, selection is trivial.
Among the retained set $\mathcal R$, we select the token-space medoid,
\begin{equation}
    j^\star
    =
    \arg\max_{j\in\mathcal R}
    \frac{1}{|\mathcal R|-1}
    \sum_{\ell\in\mathcal R,\,\ell\ne j}
    \operatorname{sim}(\mathbf z^{(j)},\mathbf z^{(\ell)}) ,
\end{equation}
where $\operatorname{sim}$ is the normalized position-wise overlap between two SLL token streams,
\begin{equation}
    \operatorname{sim}(\mathbf z,\mathbf z')
    =
    \frac{1}{\max(L,L')}
    \sum_{t=1}^{\min(L,L')}
    \mathbf{1}[z_t=z'_t] .
\end{equation}
Here, $L$ and $L'$ are the lengths of $\mathbf z$ and $\mathbf z'$.
The selected stream $\mathbf z^{(j^\star)}$ is then decoded into coordinates by the fixed SLL decoder.

Oracle best-of-$k$ is reported only as a ceiling on the sampled model's performance.
It selects
\begin{equation}
    j^\star_{\mathrm{oracle}}
    =
    \arg\max_{1\le j\le k}
    Q\!\left(D_{\mathrm{SLL}}(\mathbf z^{(j)}), y\right),
\end{equation}
where $D_{\mathrm{SLL}}$ is the SLL coordinate decoder, $y$ is the native structure, and $Q$ is a ground-truth structure-quality metric such as TM-score.
Because this rule uses the native structure, it is not available at inference time.
It measures how much quality exists inside the sampled candidate pool and gives a ceiling for confidence-based selection.

\subsection{Scaling-Law Estimation}
\label{app:scaling_law_estimation}
The scaling analysis compares validation cross-entropy loss against compute for the amino acid model, PLLM, and SLLM.
For each token space, six completed training runs are selected for the fit.
For each validation point $i$, cumulative compute is estimated from dense padded non-embedding training FLOPs,
\begin{equation}
    C_i = \frac{s_i F_{\mathrm{step}}}{10^{15}},
\end{equation}
where $s_i$ is the optimizer step and $F_{\mathrm{step}}$ is the per-step training FLOP estimate.
We use the standard forward-plus-backward convention of three times the forward FLOPs~\cite{chinchilla}, exclude embedding lookups, and report $C_i$ in PFLOPs.
These estimates cover downstream autoregressive training and exclude encoder pretraining, tokenizer learning, and preprocessing of the tokenized corpora.
The amino acid--PLL match concerns this downstream stage.
Total representation-building resources are not matched.

For the compute-scaling fit, validation points are sorted by $C_i$ and reduced to the lower-loss frontier.
We then fit a zero-offset power law,
\begin{equation}
    \label{eq:scaling_power_law}
    L(C)=A C^{-\alpha},
\end{equation}
by ordinary least squares in log space,
\begin{equation}
    \log L_i = \log A - \alpha \log C_i + \epsilon_i .
\end{equation}
The reported scaling exponent is $\alpha$.
These exponents summarize the fitted lower-loss frontiers under the zero-offset model.
Fit confidence intervals and sensitivity to an additive irreducible-loss term are not reported.

\Cref{tab:app_scaling_architecture_grid} gives the shared transformer grid used for all three token spaces.
All models use tied input and output embeddings and rotary positional embeddings.
At each grid point, the transformer blocks have the same dimensions and attention configuration.
Vocabulary size changes the size of the tied embedding/output matrix, output-projection computation, and total parameter count.

\begin{table}[h]
    \centering
    \scriptsize
    \setlength{\tabcolsep}{4.0pt}
    \caption{Decoder-only transformer grid used for the scaling-law runs. The head dimension is \num{64} throughout.}
    \label{tab:app_scaling_architecture_grid}
    \begin{tabular}{lcccccc}
        \toprule
        Size label & Width & Layers & Heads & KV heads & Head dim & FFN dim \\
        \midrule
        17.9M & \num{512} & \num{6} & \num{4} & \num{4} & \num{64} & \num{2048} \\
        51.8M & \num{640} & \num{12} & \num{5} & \num{5} & \num{64} & \num{2560} \\
        109.7M & \num{768} & \num{19} & \num{6} & \num{3} & \num{64} & \num{3072} \\
        263.4M & \num{1024} & \num{26} & \num{8} & \num{4} & \num{64} & \num{4096} \\
        611.8M & \num{1536} & \num{27} & \num{12} & \num{6} & \num{64} & \num{6144} \\
        1.33B & \num{2048} & \num{34} & \num{16} & \num{4} & \num{64} & \num{8192} \\
        \bottomrule
    \end{tabular}
\end{table}

\Cref{tab:app_scaling_training_hyperparams} summarizes the training settings.
The amino acid model and PLLM use the same UniRef50 sequences and optimization settings, each seeing approximately \num{70.7}B tokens over \num{4} epochs.
SLLM uses AFDB$\cap$UniRef50 structure-token data and the same transformer architecture grid and next-token objective, but sees approximately \num{81.5}B tokens over \num{8} epochs.
The minimum learning rate is \num{1e-5} for the amino acid model and PLLM and \num{1e-6} for SLLM.

\begin{table}[h]
    \centering
    \scriptsize
    \setlength{\tabcolsep}{3.0pt}
    \caption{Training hyperparameters for the scaling-law runs.}
    \label{tab:app_scaling_training_hyperparams}
    \begin{tabular}{
        >{\raggedright\arraybackslash}p{0.30\linewidth}
        >{\raggedright\arraybackslash}p{0.28\linewidth}
        >{\raggedright\arraybackslash}p{0.28\linewidth}}
        \toprule
        Setting & Amino acid model and PLLM & SLLM \\
        \midrule
        Training data &
        UniRef50 &
        AFDB$\cap$UniRef50~\cite{gcp_vqvae} \\
        Epochs &
        \num{4} &
        \num{8} \\
        Steps per epoch &
        \num{128}k &
        \num{128}k \\
        Total tokens seen per model &
        \num{70.7}B &
        \num{81.5}B \\
        Batch size &
        \num{512} (\num{16} per GPU) &
        \num{512} (\num{16} per GPU) \\
        Precision &
        bfloat16 &
        bfloat16 \\
        Optimizer &
        8-bit AdamW &
        8-bit AdamW \\
        Learning-rate range &
        \num{1e-4} to \num{1e-5} &
        \num{1e-4} to \num{1e-6} \\
        Weight decay &
        \num{1e-1} &
        \num{1e-1} \\
        Adam parameters &
        $\beta_1=0.9$, $\beta_2=0.98$, $\epsilon=\num{1e-7}$ &
        $\beta_1=0.9$, $\beta_2=0.98$, $\epsilon=\num{1e-7}$ \\
        Gradient clipping &
        \num{1.0} &
        \num{1.0} \\
        Schedule &
        \num{2048} warmup steps, cosine decay &
        \num{2048} warmup steps, cosine decay \\
        \bottomrule
    \end{tabular}
\end{table}

\subsection{Sequence-Side De Novo Evaluation}
\label{app:sequence_de_novo_eval}
The amino acid and PLLM de novo experiments are evaluated in the same decoded residue space.
The amino acid model generates residue tokens directly, while PLLM generates PLL code IDs that are decoded back to amino acid sequences with the fixed PLL decoder.
For length-controlled generation, both models use the same length-token protocol described in \cref{app:ar_tokenization_protocol}.

We use amino acid composition entropy as a distributional diagnostic before structural evaluation.
For a decoded sequence $x=(x_1,\ldots,x_L)$, let
\begin{equation}
    p_x(a)=\frac{1}{L}\sum_{t=1}^{L}\mathbf{1}[x_t=a],
    \qquad
    H_{\mathrm{seq}}(x)=-\sum_{a\in\mathcal A}p_x(a)\log_2 p_x(a),
\end{equation}
where $\mathcal A$ is the decoded residue alphabet.
This entropy is computed once per complete sequence and is not further divided by sequence length or normalized by $\log_2|\mathcal A|$.
We define a generated sequence as drifted when $H_{\mathrm{seq}}(x)<1.5$ bits.
The 1.5-bit threshold is a heuristic for low-complexity residue composition. Since $H_{\mathrm{seq}}$ depends only on residue frequencies, it does not measure residue order, structural quality, or protein function. We use it to compare amino acid and PLLM generation in the same decoded residue space. For the sequence-level ProteinBench comparison~\cite{proteinbench}, we retain sequences with $H_{\mathrm{seq}}(x)>1.5$ bits and evaluate \num{50} folded sequences per model, target length, and sampling temperature. Additional generation is used when needed to reach this retained count.

The low-entropy filter is a first-order fairness control, not a biological correctness label.
pLDDT is a predicted local-confidence measure introduced with AlphaFold~\cite{alphafold} and also reported by ESMFold~\cite{esm2}. It does not provide experimental validation of the structure or biological function of a generated sequence.
Low-complexity protein regions are compositionally biased or repetitive and can have distinct structural behavior~\cite{low_complexity_regions}.
In generated samples, repetitive low-entropy outputs can therefore make pLDDT partly reflect escape from residue-composition collapse rather than the quality of the full generator.

For sequence-level ProteinBench, generated sequences are evaluated along the quality, diversity, and novelty axes.

\paragraph{Quality.}
In this work, the sequence-side quality metric is the mean pLDDT of the predicted structure.
If $c_t$ is the per-residue pLDDT for residue $t$, then
\begin{equation}
    \mathrm{pLDDT}(x)=\frac{1}{L}\sum_{t=1}^{L}c_t .
\end{equation}
Higher pLDDT indicates higher folding-model confidence.
In our experiments, pLDDT is parsed from the ESMFold~\cite{esm2} structure predicted for each generated sequence.
Because the generated object is a sequence rather than a backbone, there is no original generated structure to compare against, so sequence-level evaluation does not report scTM or scRMSD.

\paragraph{Diversity.}
Foldseek~\cite{foldseek} is used to measure structural diversity over the folded generated sequences.
For folded structures $\{g_i\}_{i=1}^{n}$, pairwise TM is the mean of the maximum query or target TM-score over unordered non-self pairs,
\begin{equation}
    \mathrm{pairwise\ TM}
    =
    \frac{2}{n(n-1)}
    \sum_{1\le i<j\le n}
    \max\{\mathrm{TM}_{i\rightarrow j},\mathrm{TM}_{j\rightarrow i}\}.
\end{equation}
Lower pairwise TM indicates higher diversity.
Max Clust. is the number of Foldseek clusters divided by $n$, so higher values indicate a less collapsed generated set.

\paragraph{Novelty.}
Novelty is measured by Max TM, the mean over generated structures of the best TM-score to the Protein Data Bank (PDB) reference database,
\begin{equation}
    \mathrm{Max\ TM}
    =
    \frac{1}{n}\sum_{i=1}^{n}
    \max_{r\in\mathcal R_{\mathrm{PDB}}}
    \max\{\mathrm{TM}_{i\rightarrow r},\mathrm{TM}_{r\rightarrow i}\}.
\end{equation}
Lower Max TM indicates greater novelty relative to known PDB structures.

\begin{table}[t]
    \centering
    \scriptsize
    \setlength{\tabcolsep}{4pt}
    \caption{Length-wise sequence ProteinBench. Rows marked $^{\mathrm{ref}}$ are taken from ProteinBench~\cite{proteinbench}. Our PLM and PLLM use $T=0.5$, retain \num{50} sequences with $H_{\mathrm{seq}}(x)>1.5$ bits per length, and use ESMFold.}
    \label{tab:sequence_proteinbench_by_length}
    \begin{tabular}{rlcccc}
        \toprule
        Length & Model & pLDDT $\uparrow$ & Pair. TM $\downarrow$ & Clust. $\uparrow$ & Max TM $\downarrow$ \\
        \midrule
        100 & Native$^{\mathrm{ref}}$         & 68.46 & 0.5500 & 0.7500 & -- \\
        100 & ProGen2 (700M)$^{\mathrm{ref}}$ & 64.00 & 0.4200 & 0.9400 & 0.6400 \\
        100 & DPLM (650M)$^{\mathrm{ref}}$    & 85.38 & 0.5000 & 0.8000 & 0.7400 \\
        100 & ESM3 (1.4B)$^{\mathrm{ref}}$    & 54.26 & 0.4500 & 0.9000 & 0.6800 \\
        100 & PLLM (1.3B)                   & 42.74 & 0.2825 & 0.9800 & 0.4189 \\
        100 & PLM (1.3B)                    & 64.75 & 0.3019 & 0.8400 & 0.4741 \\
        \midrule
        200 & Native$^{\mathrm{ref}}$         & 61.91 & 0.4900 & 0.7800 & -- \\
        200 & ProGen2 (700M)$^{\mathrm{ref}}$ & 69.91 & 0.4000 & 0.9100 & 0.6900 \\
        200 & DPLM (650M)$^{\mathrm{ref}}$    & 93.54 & 0.5400 & 0.7000 & 0.9100 \\
        200 & ESM3 (1.4B)$^{\mathrm{ref}}$    & 58.45 & 0.3500 & 1.0000 & 0.8000 \\
        200 & PLLM (1.3B)                   & 59.37 & 0.2674 & 0.9600 & 0.6126 \\
        200 & PLM (1.3B)                    & 69.02 & 0.2406 & 0.9400 & 0.6048 \\
        \midrule
        300 & Native$^{\mathrm{ref}}$         & 61.49 & 0.5100 & 0.8500 & -- \\
        300 & ProGen2 (700M)$^{\mathrm{ref}}$ & 65.69 & 0.4200 & 0.9300 & 0.6600 \\
        300 & DPLM (650M)$^{\mathrm{ref}}$    & 93.07 & 0.5700 & 0.6300 & 0.9100 \\
        300 & ESM3 (1.4B)$^{\mathrm{ref}}$    & 48.08 & 0.3200 & 1.0000 & 0.7500 \\
        300 & PLLM (1.3B)                   & 69.21 & 0.2472 & 0.9800 & 0.7441 \\
        300 & PLM (1.3B)                    & 68.23 & 0.2222 & 0.9600 & 0.6419 \\
        \midrule
        500 & Native$^{\mathrm{ref}}$         & 62.95 & 0.5100 & 0.7800 & -- \\
        500 & ProGen2 (700M)$^{\mathrm{ref}}$ & 61.45 & 0.3200 & 0.9500 & 0.6800 \\
        500 & DPLM (650M)$^{\mathrm{ref}}$    & 87.73 & 0.4300 & 0.8500 & 0.8500 \\
        500 & ESM3 (1.4B)$^{\mathrm{ref}}$    & 52.17 & 0.3000 & 1.0000 & 0.5400 \\
        500 & PLLM (1.3B)                   & 68.11 & 0.2297 & 0.9600 & 0.7714 \\
        500 & PLM (1.3B)                    & 59.03 & 0.2005 & 0.9600 & 0.6228 \\
        \bottomrule
    \end{tabular}
    \vspace{0.25em}
    \begin{minipage}{\linewidth}
        \footnotesize
        The main-table means in \cref{tab:pll_conditional_proteinbench_t05} give equal weight to the four length rows.
        ProteinBench specifies \num{50} sequences per reference model and target length but does not specify exact repeat counts or sequence-generation temperatures.
        Our reported subsets use selection seed \num{42}.
        Additional generation under the same condition fills retained sets when needed.
        Our rows report these fixed-seed evaluation sets without uncertainty estimates across independent evaluation repeats.
    \end{minipage}
\end{table}

\FloatBarrier

\subsection{Structure-Side De Novo Evaluation}
\label{app:structure_de_novo_eval}
SLLM de novo generation is evaluated as backbone generation.
SLLM samples an SLL token sequence, and the fixed SLL decoder converts the generated code IDs into backbone coordinates.
We then evaluate the decoded backbones with the structure-design ProteinBench protocol~\cite{proteinbench}.
This path differs from the sequence-side protocol because the generated object is already a structure, so self-consistency metrics are defined.
The same quality, diversity, and novelty axes are used.
For length-conditioned SLLM generation, the largest SLLM is continued for two epochs with length prefixes on 50\% of samples, using the tokenization protocol in \cref{app:ar_tokenization_protocol}.

\paragraph{Quality.}
For each target length, we evaluate 100 generated backbones.
For a generated backbone $b_i$, ProteinMPNN~\cite{proteinmpnn} designs candidate sequences and ESMFold~\cite{esm2} refolds those sequences.
We retain the maximum scTM and independently minimum scRMSD across the candidate designs.
If $d_{i,m}$ is the refolded structure for design $m$, then
\begin{equation}
    m_i^\star=\arg\max_m \mathrm{TM}(b_i,d_{i,m}),
    \qquad
    \mathrm{scTM}_i=\mathrm{TM}(b_i,d_{i,m_i^\star}),
    \qquad
    \mathrm{scRMSD}_i=\min_m\mathrm{RMSD}(b_i,d_{i,m}).
\end{equation}
The maximum-scTM and minimum-scRMSD designs can differ.
Per-length quality scores are means over the \num{100} backbones.
For FoldFlow-2 and DPLM-2, ProteinMPNN produces eight sequences per backbone at sampling temperature 0.1.
Higher scTM and lower scRMSD indicate that the generated backbone can support a sequence that refolds back to the same structure.

\paragraph{Diversity.}
Diversity is measured by nearest-neighbor TM and Max Clust. on the maximum-scTM refold set.
For the $N=100$ selected refolds, let $s_{ij}=\max\{\mathrm{TM}_{i\rightarrow j},\mathrm{TM}_{j\rightarrow i}\}$.
Nearest-neighbor TM is
\begin{equation}
    \mathrm{NN\ TM}=\frac{1}{N}\sum_{i=1}^{N}\max_{j\ne i}s_{ij}.
\end{equation}
Lower nearest-neighbor TM and higher Max Clust. indicate more diverse samples.

\paragraph{Novelty.}
Novelty uses the same Max TM definition as in \cref{app:sequence_de_novo_eval}, with lower values indicating less similarity to PDB references.
All evaluated backbones are normalized to backbone heavy atoms and grouped by target length.
PLAID outputs with lengths 56, 112, 320, and 528 are clipped to the target lengths 50, 100, 300, and 500 before evaluation.
The structure-design comparison in \cref{tab:sll_conditional_proteinbench} reports unweighted means over $L\in\{50,100,300,500\}$.
DPLM-2 uses the 650M model in unconditional backbone-generation mode.
Native, RFdiffusion, and Genie results are taken from ProteinBench and marked with an asterisk.
Their TM similarity values use the published pairwise TM measure, while unstarred rows use the nearest-neighbor TM measure defined above.

\paragraph{Length-wise structure results.}
\Cref{tab:structure_proteinbench_by_length,tab:structure_proteinbench_by_length_long} report the individual target lengths used in the main structure-design comparison.
The main table gives equal weight to $L\in\{50,100,300,500\}$.

\begin{table}[p]
    \centering
    \scriptsize
    \setlength{\tabcolsep}{3.3pt}
    \caption{Structure-design results on ProteinBench~\cite{proteinbench} at target lengths 50 and 100. $^\ast$ Results taken from ProteinBench. $^\dagger$ DPLM-2 jointly models amino acid sequences and structure tokens.}
    \label{tab:structure_proteinbench_by_length}
    \begin{tabular}{rlccccc}
        \toprule
        Length & Model & scTM $\uparrow$ & scRMSD $\downarrow$ & TM similarity $\downarrow$ & Clust. $\uparrow$ & Max TM $\downarrow$ \\
        \midrule
        50 & Native$^\ast$ & 0.91 & 0.74 & 0.29 & 0.66 & N/A \\
        50 & Genie$^\ast$~\cite{genie} & 0.57 & 3.12 & 0.32 & 0.90 & 0.57 \\
        50 & RFdiffusion$^\ast$~\cite{rfdiffusion} & 0.95 & 0.45 & 0.58 & 0.67 & 0.65 \\
        50 & Protpardelle~\cite{protpardelle1c} & 0.9655 & 0.5785 & 0.7369 & 0.3200 & 0.7608 \\
        50 & ProteinGenerator~\cite{proteingenerator} & 0.7984 & 1.2202 & 0.8938 & 0.1300 & 0.3468 \\
        50 & PLAID~\cite{plaid} & 0.8028 & 1.3188 & 0.7012 & 0.5600 & 0.6463 \\
        50 & La-Proteina~\cite{laproteina} & 0.9840 & 0.3816 & 0.7989 & 0.3200 & 0.8458 \\
        50 & Pallatom~\cite{pallatom} & 0.9739 & 0.4937 & 0.7209 & 0.3300 & 0.7683 \\
        50 & FoldFlow-2~\cite{foldflow2} & 0.8866 & 0.9081 & 0.7565 & 0.3700 & 0.8385 \\
        50 & DPLM-2 (650M)$^\dagger$~\cite{dplm2} & 0.8679 & 1.0511 & 0.7791 & 0.5800 & 0.8735 \\
        50 & \textbf{SLLM} & 0.6194 & 1.7590 & 0.6088 & 0.7300 & 0.3418 \\
        \midrule
        100 & Native$^\ast$ & 0.96 & 0.67 & 0.30 & 0.77 & N/A \\
        100 & Genie$^\ast$~\cite{genie} & 0.69 & 3.38 & 0.31 & 0.96 & 0.59 \\
        100 & RFdiffusion$^\ast$~\cite{rfdiffusion} & 0.98 & 0.48 & 0.41 & 0.32 & 0.76 \\
        100 & Protpardelle~\cite{protpardelle1c} & 0.9685 & 0.8236 & 0.8277 & 0.1000 & 0.8280 \\
        100 & ProteinGenerator~\cite{proteingenerator} & 0.9602 & 0.8865 & 0.6844 & 0.3900 & 0.7563 \\
        100 & PLAID~\cite{plaid} & 0.8464 & 1.7470 & 0.6608 & 0.5600 & 0.7596 \\
        100 & La-Proteina~\cite{laproteina} & 0.9887 & 0.4611 & 0.6294 & 0.5600 & 0.7439 \\
        100 & Pallatom~\cite{pallatom} & 0.9829 & 0.5576 & 0.5819 & 0.6200 & 0.6956 \\
        100 & FoldFlow-2~\cite{foldflow2} & 0.9392 & 1.0276 & 0.6738 & 0.3400 & 0.7622 \\
        100 & DPLM-2 (650M)$^\dagger$~\cite{dplm2} & 0.8720 & 1.4944 & 0.7370 & 0.4400 & 0.8995 \\
        100 & \textbf{SLLM} & 0.5805 & 2.5859 & 0.4957 & 0.9600 & 0.5593 \\
        \bottomrule
    \end{tabular}
\end{table}

\begin{table}[p]
    \centering
    \scriptsize
    \setlength{\tabcolsep}{3.3pt}
    \caption{Structure-design results on ProteinBench~\cite{proteinbench} at target lengths 300 and 500. $^\ast$ Results taken from ProteinBench. $^\dagger$ DPLM-2 jointly models amino acid sequences and structure tokens.}
    \label{tab:structure_proteinbench_by_length_long}
    \begin{tabular}{rlccccc}
        \toprule
        Length & Model & scTM $\uparrow$ & scRMSD $\downarrow$ & TM similarity $\downarrow$ & Clust. $\uparrow$ & Max TM $\downarrow$ \\
        \midrule
        300 & Native$^\ast$ & 0.97 & 0.82 & 0.28 & 0.77 & N/A \\
        300 & Genie$^\ast$~\cite{genie} & 0.27 & 20.37 & 0.23 & 1.00 & 0.30 \\
        300 & RFdiffusion$^\ast$~\cite{rfdiffusion} & 0.96 & 1.03 & 0.36 & 0.65 & 0.64 \\
        300 & Protpardelle~\cite{protpardelle1c} & 0.9603 & 1.3968 & 0.7395 & 0.1500 & 0.8152 \\
        300 & ProteinGenerator~\cite{proteingenerator} & 0.8926 & 2.0403 & 0.6409 & 0.2600 & 0.7158 \\
        300 & PLAID~\cite{plaid} & 0.7474 & 3.0922 & 0.5950 & 0.5700 & 0.7085 \\
        300 & La-Proteina~\cite{laproteina} & 0.9870 & 0.6999 & 0.5477 & 1.0000 & 0.6765 \\
        300 & Pallatom~\cite{pallatom} & 0.8612 & 1.9246 & 0.4091 & 1.0000 & 0.5665 \\
        300 & FoldFlow-2~\cite{foldflow2} & 0.9483 & 1.4434 & 0.4652 & 0.9700 & 0.7266 \\
        300 & DPLM-2 (650M)$^\dagger$~\cite{dplm2} & 0.9182 & 1.7516 & 0.6787 & 0.5900 & 0.9077 \\
        300 & \textbf{SLLM} & 0.5210 & 4.0381 & 0.3778 & 0.9800 & 0.6427 \\
        \midrule
        500 & Native$^\ast$ & 0.97 & 1.07 & 0.29 & 0.80 & N/A \\
        500 & Genie$^\ast$~\cite{genie} & 0.25 & 26.08 & 0.23 & 1.00 & 0.22 \\
        500 & RFdiffusion$^\ast$~\cite{rfdiffusion} & 0.79 & 5.60 & 0.33 & 0.89 & 0.62 \\
        500 & Protpardelle~\cite{protpardelle1c} & 0.8212 & 3.4017 & 0.6867 & 0.3000 & 0.7823 \\
        500 & ProteinGenerator~\cite{proteingenerator} & 0.3582 & 6.2083 & 0.4025 & 0.8900 & 0.5036 \\
        500 & PLAID~\cite{plaid} & 0.4563 & 5.4272 & 0.4043 & 0.9200 & 0.5647 \\
        500 & La-Proteina~\cite{laproteina} & 0.9113 & 1.6057 & 0.5162 & 0.7400 & 0.6815 \\
        500 & Pallatom~\cite{pallatom} & 0.3786 & 6.2550 & 0.3991 & 1.0000 & 0.4879 \\
        500 & FoldFlow-2~\cite{foldflow2} & 0.7689 & 3.3023 & 0.4126 & 1.0000 & 0.7463 \\
        500 & DPLM-2 (650M)$^\dagger$~\cite{dplm2} & 0.8759 & 2.3074 & 0.7085 & 0.5400 & 0.9256 \\
        500 & \textbf{SLLM} & 0.4354 & 5.2065 & 0.3408 & 1.0000 & 0.6708 \\
        \bottomrule
    \end{tabular}
\end{table}

\FloatBarrier

\section{Data Sources}
\label{app:data_sources}
The project uses three main data sources.
UniRef50~\cite{uniprot2025} provides the sequence corpus for learning PLL and for sequence generation.
AFDB representative structures provide the structure corpus for learning SLL.
For SLLM generation, we use the larger AFDB$\cap$UniRef50 structure corpus introduced by GCP-VQVAE~\cite{gcp_vqvae}.
This corpus is suitable for generation because SLL codes are trained with an inverse-folding auxiliary objective and therefore retain sequence-identity information.
Starting from UniRef50 sequence clusters limits sequence redundancy at the 50\% identity level before the corresponding AFDB structures are used as structure-token training data.

\begin{table}[h]
    \centering
    \scriptsize
    \caption{Main data sources.}
    \label{tab:app_data_sources}
    \begin{tabular}{
        >{\raggedright\arraybackslash}p{0.34\linewidth}
        >{\raggedright\arraybackslash}p{0.34\linewidth}
        >{\raggedright\arraybackslash}p{0.18\linewidth}}
        \toprule
        Data source & Role & Approx. samples \\
        \midrule
        UniRef50~\cite{uniprot2025} &
        PLL tokenizer; PLLM and amino acid generation &
        \num{65.78}M \\
        AFDB representatives &
        SLL tokenizer &
        \num{2.28}M \\
        AFDB$\cap$UniRef50 following GCP-VQVAE~\cite{gcp_vqvae} &
        SLLM generation &
        \num{38.15}M \\
        \bottomrule
    \end{tabular}
\end{table}

The held-out validation splits contain 10,000 UniRef50 sequences, 10,000 AFDB representative structures, and 5,000 AFDB$\cap$UniRef50 structure-token pairs.

\begin{figure}[h]
    \centering
    \includegraphics[width=\linewidth]{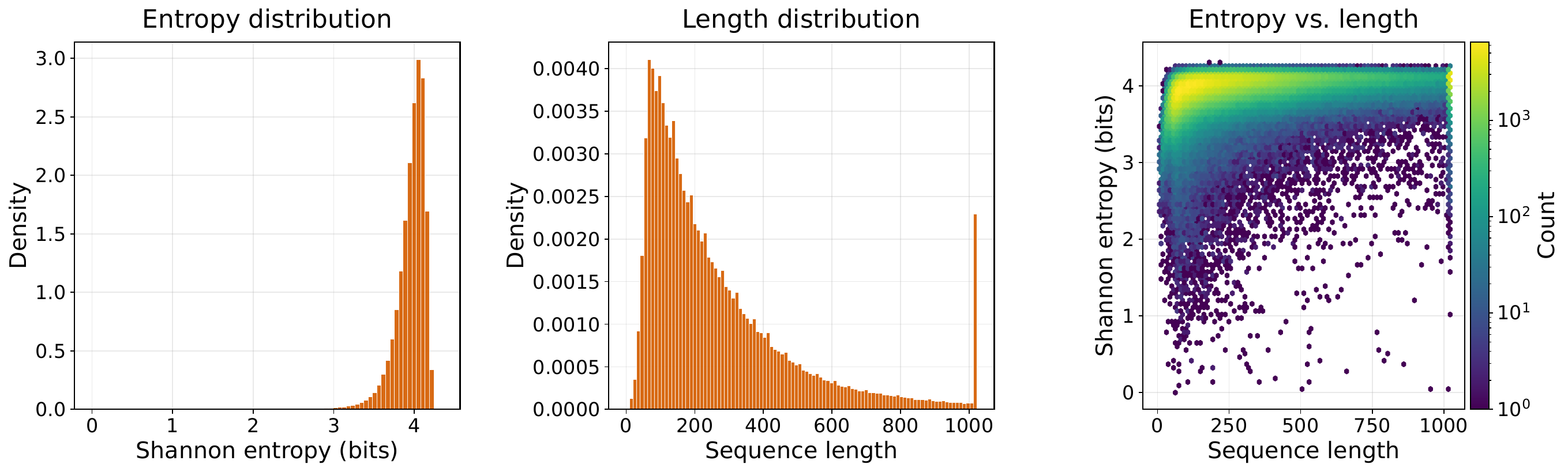}
    \caption{UniRef50 reference statistics after truncating sequences to 1022 residues. These distributions provide the reference length and amino acid composition statistics used when analyzing de novo amino acid and PLLM generation.}
    \label{fig:uniref50_statistics}
\end{figure}

The UniRef50 reference statistics used for sequence-generation analysis are shown in \cref{fig:uniref50_statistics}.

For SLL reconstruction comparisons, we use the benchmark splits and reporting protocol from GCP-VQVAE~\cite{gcp_vqvae}.
For representation selection, we use the PST protocol from StructTokenBench~\cite{structtokenbench}.

\section{Additional Experimental Results}
\label{app:additional_results}

\subsection{PLL Per-Amino-Acid Recovery}
\label{app:pll_per_aa_recovery}
The Stage 2 PLL residue head reaches high recovery across all standard amino acids, not only in the micro average.
\Cref{tab:app_pll_per_aa_recovery} reports the validation accuracy for each residue class.
The $X$ class is lower because it represents unknown or non-standard residues rather than a single conventional amino acid.

\begin{table}[h]
    \centering
    \scriptsize
    \caption{Stage 2 PLL per-amino-acid recovery on the UniRef50 validation split.}
    \label{tab:app_pll_per_aa_recovery}
    \begin{tabular}{
        cS[table-format=2.2]
        cS[table-format=2.2]
        cS[table-format=2.2]}
        \toprule
        Residue & {Accuracy (\%)} & Residue & {Accuracy (\%)} & Residue & {Accuracy (\%)} \\
        \midrule
        A & 99.89 & I & 99.81 & R & 99.81 \\
        C & 99.18 & K & 99.77 & S & 99.87 \\
        D & 99.83 & L & 99.91 & T & 99.86 \\
        E & 99.87 & M & 99.48 & V & 99.87 \\
        F & 99.72 & N & 99.79 & W & 99.20 \\
        G & 99.86 & P & 99.84 & Y & 99.69 \\
        H & 99.56 & Q & 99.80 & X & 90.52 \\
        \bottomrule
    \end{tabular}
\end{table}

\subsection{SLL}
\label{app:sll_additional_results}

\subsubsection{SLL Supervised PST Ablation}
\label{app:sll_supervised_pst_ablation}
\Cref{tab:app_sll_supervised_pst_ablation} reports the supervised PST ablations used to select the final SLL recipe.
Task columns average available splits, and the category average averages all rows within that category.
Cell colors are relative to the Base tokenizer.
The configuration names are shorthand for four groups of tokenizer changes.

\textbf{Auxiliary heads.}
ESM pre-decoder and ESM post-decoder add frozen ESM-2 representation prediction before or after the coordinate decoder.
Inverse folding 0.01, 0.1, and 1.0 add a pre-decoder amino acid recovery head with the indicated loss weight.
pLDDT pre-decoder and pLDDT post-decoder add a per-residue confidence regression head at the indicated location.

\textbf{Reconstruction losses.}
MSE, MSE + distance, MSE + direction, and Distance + direction change the active reconstruction losses.
Distance denotes the pairwise backbone-distance loss, and direction denotes the GCP-VQVAE direction loss.

\textbf{VQ choices.}
Rotation trick changes how gradients pass through the quantized vectors.
The temperature rows test the codebook-temperature setting with or without the rotation trick.

\textbf{Combined recipe.}
Base is the GCP-VQVAE Lite recipe under the same ablation protocol.
The selected combined recipe uses MSE + distance reconstruction, inverse folding with weight 0.1, pLDDT pre-decoder, ESM post-decoder, and the rotation trick.
The two-times-longer row repeats the same combined recipe for a longer ablation run.

\begin{table}[h]
    \centering
    \tiny
    \setlength{\tabcolsep}{1.55pt}
    \caption{Supervised PST ablation results for SLL. Task columns average available splits, and Avg averages all rows within each category. Cells are colored relative to the Base row. Gray rows mark runs that did not converge cleanly.}
    \label{tab:app_sll_supervised_pst_ablation}
    \resizebox{\linewidth}{!}{%
    \begin{tabular}{lccccccccc cccc}
        \toprule
        & \multicolumn{9}{c}{Functional AUROC (\%) $\uparrow$} & \multicolumn{4}{c}{Physicochemical Spearman (\%) $\uparrow$} \\
        \cmidrule(lr){2-10}\cmidrule(lr){11-14}
        Configuration & BindInt & BindBio & BindShake & CatInt & CatBio & Con & Rep & Ept & Avg & RMSF & BFact & NEQ & Avg \\
        \midrule
        Base & 56.48 & 59.36 & 71.84 & 60.02 & 54.90 & 58.21 & 54.97 & 57.10 & 58.26 & 34.66 & 22.89 & 14.26 & 23.93 \\
        ESM pre-decoder & \cellcolor{green!10}57.91 & \cellcolor{green!10}61.15 & \cellcolor{red!10}70.00 & \cellcolor{green!10}61.91 & \cellcolor{green!26}60.18 & \cellcolor{green!26}63.63 & \cellcolor{green!18}59.07 & \cellcolor{red!18}54.70 & \cellcolor{green!18}60.47 & \cellcolor{green!10}35.55 & \cellcolor{red!10}21.90 & \cellcolor{green!18}16.84 & \cellcolor{green!10}24.76 \\
        ESM post-decoder & \cellcolor{green!18}58.94 & \cellcolor{green!18}61.86 & \cellcolor{green!10}73.03 & \cellcolor{green!18}63.11 & \cellcolor{green!18}58.12 & \cellcolor{green!18}61.18 & \cellcolor{green!18}57.07 & \cellcolor{green!10}57.88 & \cellcolor{green!18}60.62 & \cellcolor{green!18}36.70 & \cellcolor{green!10}24.20 & \cellcolor{green!10}14.88 & \cellcolor{green!10}25.26 \\
        Inverse folding 0.01 & \cellcolor{green!10}58.00 & \cellcolor{red!18}55.85 & \cellcolor{green!10}72.42 & \cellcolor{red!10}59.29 & \cellcolor{red!10}54.18 & \cellcolor{green!10}58.63 & \cellcolor{red!10}54.85 & \cellcolor{green!18}59.81 & \cellcolor{red!10}58.24 & \cellcolor{green!10}34.79 & \cellcolor{green!10}23.59 & \cellcolor{green!18}16.63 & \cellcolor{green!10}25.00 \\
        Inverse folding 1.0 & \cellcolor{green!36}67.18 & \cellcolor{green!36}88.90 & \cellcolor{red!26}63.38 & \cellcolor{green!18}63.06 & \cellcolor{green!36}86.18 & \cellcolor{green!18}62.00 & \cellcolor{green!10}55.07 & \cellcolor{green!10}58.48 & \cellcolor{green!36}68.34 & \cellcolor{red!26}25.96 & \cellcolor{red!26}17.15 & \cellcolor{green!26}21.49 & \cellcolor{red!18}21.53 \\
        Inverse folding 0.1 & \cellcolor{green!26}64.32 & \cellcolor{green!36}80.04 & \cellcolor{red!10}71.12 & \cellcolor{green!26}66.27 & \cellcolor{green!36}75.74 & \cellcolor{green!18}62.76 & \cellcolor{green!26}62.04 & \cellcolor{red!10}56.91 & \cellcolor{green!26}67.15 & \cellcolor{red!18}30.52 & \cellcolor{red!18}20.27 & \cellcolor{green!26}20.20 & \cellcolor{red!10}23.66 \\
        \textcolor{gray!70}{Distance + direction (not converged)} & \textcolor{gray!70}{50.00} & \textcolor{gray!70}{58.21} & \textcolor{gray!70}{51.10} & \textcolor{gray!70}{50.56} & \textcolor{gray!70}{50.00} & \textcolor{gray!70}{50.68} & \textcolor{gray!70}{50.68} & \textcolor{gray!70}{51.99} & \textcolor{gray!70}{51.69} & \textcolor{gray!70}{5.36} & \textcolor{gray!70}{3.67} & \textcolor{gray!70}{9.53} & \textcolor{gray!70}{6.18} \\
        MSE & \cellcolor{green!18}60.73 & \cellcolor{red!10}57.63 & \cellcolor{green!18}74.99 & \cellcolor{green!26}65.32 & \cellcolor{green!10}56.02 & \cellcolor{green!26}63.25 & \cellcolor{green!26}62.07 & \cellcolor{red!10}56.78 & \cellcolor{green!18}61.24 & \cellcolor{green!18}38.69 & \cellcolor{green!18}25.47 & \cellcolor{green!10}15.71 & \cellcolor{green!18}26.62 \\
        MSE + distance & \cellcolor{green!18}58.78 & \cellcolor{green!18}61.98 & \cellcolor{green!26}76.92 & \cellcolor{green!18}62.56 & \cellcolor{green!10}56.85 & \cellcolor{green!18}62.08 & \cellcolor{green!18}58.75 & \cellcolor{green!18}60.14 & \cellcolor{green!18}61.28 & \cellcolor{green!26}44.13 & \cellcolor{green!26}29.29 & \cellcolor{green!18}17.55 & \cellcolor{green!26}30.32 \\
        MSE + direction & \cellcolor{green!10}58.37 & \cellcolor{green!10}61.06 & \cellcolor{red!10}69.98 & \cellcolor{green!10}61.12 & \cellcolor{green!18}57.41 & \cellcolor{green!18}61.28 & \cellcolor{green!10}55.87 & \cellcolor{red!18}54.82 & \cellcolor{green!10}59.32 & \cellcolor{red!26}27.09 & \cellcolor{red!18}18.81 & \cellcolor{green!18}16.90 & \cellcolor{red!18}20.93 \\
        pLDDT post-decoder & \cellcolor{green!18}59.75 & \cellcolor{green!26}64.39 & \cellcolor{green!10}73.10 & \cellcolor{green!18}63.40 & \cellcolor{green!26}61.77 & \cellcolor{green!18}60.75 & \cellcolor{green!18}57.98 & \cellcolor{green!18}59.47 & \cellcolor{green!18}61.87 & \cellcolor{green!18}36.98 & \cellcolor{green!18}25.51 & \cellcolor{green!18}16.52 & \cellcolor{green!18}26.33 \\
        pLDDT pre-decoder & \cellcolor{green!18}58.94 & \cellcolor{green!36}71.42 & \cellcolor{red!10}71.82 & \cellcolor{green!10}60.89 & \cellcolor{green!36}70.26 & \cellcolor{green!10}59.65 & \cellcolor{red!10}54.22 & \cellcolor{green!10}58.59 & \cellcolor{green!18}62.65 & \cellcolor{green!26}41.34 & \cellcolor{green!18}25.43 & \cellcolor{green!26}20.08 & \cellcolor{green!26}28.95 \\
        ESM pre + pLDDT pre + IF 0.1 & \cellcolor{green!26}62.11 & \cellcolor{green!36}80.40 & \cellcolor{red!26}65.22 & \cellcolor{green!10}60.84 & \cellcolor{green!36}74.37 & \cellcolor{red!10}58.09 & \cellcolor{green!18}59.50 & \cellcolor{green!18}60.29 & \cellcolor{green!26}65.09 & \cellcolor{red!18}32.20 & \cellcolor{red!18}19.12 & \cellcolor{green!26}23.67 & \cellcolor{green!10}25.00 \\
        Rotation trick & \cellcolor{green!18}60.95 & \cellcolor{green!36}75.85 & \cellcolor{green!10}73.12 & \cellcolor{green!18}63.28 & \cellcolor{green!36}70.42 & \cellcolor{green!18}61.13 & \cellcolor{green!18}59.62 & \cellcolor{red!10}56.45 & \cellcolor{green!26}64.56 & \cellcolor{red!10}34.61 & \cellcolor{red!10}22.81 & \cellcolor{green!10}16.00 & \cellcolor{green!10}24.47 \\
        Rotation trick + temp 0.1 & \cellcolor{green!18}59.77 & \cellcolor{green!18}62.98 & \cellcolor{green!10}72.88 & \cellcolor{green!10}61.70 & \cellcolor{green!26}60.58 & \cellcolor{green!10}59.36 & \cellcolor{green!10}56.81 & \cellcolor{red!10}56.17 & \cellcolor{green!18}60.50 & \cellcolor{green!10}36.04 & \cellcolor{green!10}23.82 & \cellcolor{red!10}14.07 & \cellcolor{green!10}24.64 \\
        \textcolor{gray!70}{Rotation trick + temp 0.5 (not converged)} & \textcolor{gray!70}{59.47} & \textcolor{gray!70}{64.34} & \textcolor{gray!70}{55.82} & \textcolor{gray!70}{56.53} & \textcolor{gray!70}{65.01} & \textcolor{gray!70}{59.39} & \textcolor{gray!70}{59.77} & \textcolor{gray!70}{53.80} & \textcolor{gray!70}{59.49} & \textcolor{gray!70}{17.40} & \textcolor{gray!70}{10.18} & \textcolor{gray!70}{13.95} & \textcolor{gray!70}{13.84} \\
        Sample codebook temp 0.1 & \cellcolor{red!10}56.27 & \cellcolor{green!10}59.37 & \cellcolor{green!10}73.10 & \cellcolor{green!10}61.46 & \cellcolor{green!18}57.14 & \cellcolor{green!10}58.84 & \cellcolor{red!10}54.48 & \cellcolor{red!10}55.59 & \cellcolor{green!10}58.63 & \cellcolor{green!18}38.24 & \cellcolor{green!10}24.64 & \cellcolor{red!10}12.74 & \cellcolor{green!10}25.20 \\
        Sample codebook temp 0.5 & \cellcolor{green!10}57.57 & \cellcolor{green!18}63.71 & \cellcolor{red!18}68.13 & \cellcolor{green!10}60.80 & \cellcolor{green!18}58.36 & \cellcolor{green!10}58.27 & \cellcolor{green!10}55.14 & \cellcolor{green!10}57.61 & \cellcolor{green!10}59.40 & \cellcolor{red!26}29.00 & \cellcolor{red!18}18.72 & \cellcolor{red!10}13.22 & \cellcolor{red!18}20.31 \\
        Combined, selected & \cellcolor{green!36}67.68 & \cellcolor{green!36}93.15 & \cellcolor{red!18}68.39 & \cellcolor{green!26}66.95 & \cellcolor{green!36}90.87 & \cellcolor{green!26}65.92 & \cellcolor{green!26}61.72 & \cellcolor{green!26}62.29 & \cellcolor{green!36}\textbf{72.37} & \cellcolor{green!26}43.01 & \cellcolor{green!18}25.09 & \cellcolor{green!36}28.79 & \cellcolor{green!26}\textbf{32.30} \\
        Combined, trained two times longer & \cellcolor{green!36}67.14 & \cellcolor{green!36}93.09 & \cellcolor{green!10}72.97 & \cellcolor{green!26}68.49 & \cellcolor{green!36}92.16 & \cellcolor{green!26}67.30 & \cellcolor{green!26}63.25 & \cellcolor{green!26}63.50 & \cellcolor{green!36}73.52 & \cellcolor{green!18}39.36 & \cellcolor{green!18}24.91 & \cellcolor{green!36}28.35 & \cellcolor{green!26}30.87 \\
        \bottomrule
    \end{tabular}}
\end{table}

\subsubsection{SLL Supervised PST Full-Run}
\label{app:sll_supervised_pst_full_run}
\Cref{tab:app_sll_supervised_pst_full_run} follows the supervised PST table style used by the original work.
The non-GCP baselines are reported with the PST-standard two-layer MLP probe.
For GCP-VQVAE variants, parentheses report the four-layer MLP and transformer probes.

\begin{table}[h]
    \centering
    \tiny
    \setlength{\tabcolsep}{1.8pt}
    \caption{Supervised PST full-run comparison between SLL and structure-tokenizer baselines. GCP-VQVAE Lite 2 denotes SLL, and parentheses report four-layer MLP and transformer probe results for GCP-VQVAE variants.}
    \label{tab:app_sll_supervised_pst_full_run}
    \resizebox{\linewidth}{!}{%
    \begin{tabular}{llccccccccc}
        \toprule
        Task & Split &
        \shortstack{FoldSeek\\\cite{foldseek}} &
        \shortstack{ProTokens\\\cite{protokens}} &
        \shortstack{VanillaVQ\\\cite{structtokenbench}} &
        \shortstack{AIDO\\\cite{aido_structure_tokenizer}} &
        \shortstack{ESM3\\\cite{esm3}} &
        \shortstack{AminoAseed\\\cite{structtokenbench}} &
        \shortstack{GCP-VQVAE\\Large~\cite{gcp_vqvae}} &
        \shortstack{GCP-VQVAE\\Lite~\cite{gcp_vqvae}} &
        \shortstack{GCP-VQVAE\\Lite 2} \\
        \midrule
        \multicolumn{11}{c}{Functional Site Prediction, AUROC (\%) $\uparrow$} \\
        \midrule
        \multirow{2}{*}{BindInt} & Fold & 53.18 & 44.66 & 47.25 & 44.66 & 44.30 & 47.11 & 52.61 (50.26, 52.85) & 49.43 (49.00, \textbf{55.03}) & 52.60 (\underline{53.73}, 50.85) \\
        & SupFam & 46.20 & 86.05 & 86.71 & 84.21 & \underline{90.77} & 90.53 & 58.04 (70.61, 80.67) & 62.27 (70.27, 80.77) & 79.59 (79.84, \textbf{93.86}) \\
        \addlinespace[0.35em]
        \multirow{2}{*}{BindBio} & Fold & 52.37 & 58.47 & 62.02 & 65.50 & 62.84 & 65.73 & 75.05 (92.74, 84.01) & 80.93 (\textbf{94.61}, 91.74) & 93.97 (\underline{94.28}, 87.99) \\
        & SupFam & 52.41 & 60.47 & 62.92 & 66.70 & 65.22 & 68.30 & 75.64 (93.40, 86.62) & 81.72 (\underline{94.08}, 90.08) & 93.95 (\textbf{94.10}, 88.52) \\
        \addlinespace[0.35em]
        BindShake & Org & 53.40 & 59.82 & 67.04 & 69.28 & 66.10 & 69.61 & 66.97 (67.96, \underline{72.71}) & 66.34 (67.11, 68.08) & 70.81 (70.79, \textbf{75.75}) \\
        \addlinespace[0.35em]
        \multirow{2}{*}{CatInt} & Fold & 53.43 & 58.16 & 58.89 & 57.30 & 61.09 & \underline{62.19} & 55.92 (53.15, 60.67) & 54.36 (55.49, 58.09) & 53.46 (54.62, \textbf{66.03}) \\
        & SupFam & 51.41 & 83.85 & 85.00 & 81.94 & 89.82 & \underline{91.91} & 63.22 (73.43, 84.00) & 62.24 (68.88, 85.18) & 74.56 (71.92, \textbf{93.09}) \\
        \addlinespace[0.35em]
        \multirow{2}{*}{CatBio} & Fold & 56.37 & 56.14 & 67.58 & 73.72 & 65.33 & 65.95 & 67.12 (79.66, 71.64) & 67.25 (\textbf{92.28}, 86.11) & 90.90 (\underline{91.77}, 71.61) \\
        & SupFam & 53.78 & 64.05 & 70.92 & 78.66 & 74.65 & 87.59 & 73.77 (89.64, 86.95) & 75.57 (\textbf{95.72}, 93.38) & 94.19 (\underline{94.88}, 89.89) \\
        \addlinespace[0.35em]
        \multirow{2}{*}{Con} & Fold & 49.26 & 56.23 & 56.98 & 56.64 & 55.22 & 57.23 & 52.64 (53.29, \underline{62.97}) & 51.15 (51.56, 61.52) & 53.73 (53.08, \textbf{65.92}) \\
        & SupFam & 51.39 & 74.33 & 74.60 & 73.79 & 80.53 & \underline{86.60} & 61.31 (68.84, 84.05) & 58.66 (65.23, 85.79) & 71.52 (72.25, \textbf{96.21}) \\
        \addlinespace[0.35em]
        \multirow{2}{*}{Rep} & Fold & 47.70 & \underline{77.25} & 75.99 & \textbf{77.69} & 74.70 & 74.97 & 51.19 (50.95, 66.19) & 52.04 (52.79, 61.22) & 48.81 (50.13, 74.85) \\
        & SupFam & 52.53 & 78.90 & 82.09 & 78.08 & 82.36 & 84.57 & 54.17 (65.26, \underline{85.77}) & 56.50 (54.65, \textbf{85.94}) & 72.89 (75.04, 85.26) \\
        \addlinespace[0.35em]
        \multirow{2}{*}{Ept} & Fold & 54.52 & 54.69 & 59.28 & 60.26 & \textbf{63.69} & \underline{62.16} & 54.24 (54.21, 58.19) & 53.05 (54.63, 56.89) & 57.68 (57.90, 62.11) \\
        & SupFam & 50.56 & 67.52 & 67.24 & \textbf{72.30} & 61.97 & \underline{72.02} & 55.91 (58.03, 66.31) & 56.13 (58.25, 64.09) & 61.08 (60.39, 69.20) \\
        \addlinespace[0.35em]
        \textbf{Average} &  & 51.90 & 65.37 & 68.30 & 69.38 & 69.24 & 72.43 & 61.19 (68.10, 73.57) & 61.84 (68.30, \underline{74.93}) & 71.32 (71.65, \textbf{78.08}) \\
        \midrule
        \multicolumn{11}{c}{Physicochemical Property Prediction, Spearman $\rho$ (\%) $\uparrow$} \\
        \midrule
        \multirow{2}{*}{FlexRMSF} & Fold & 15.35 & 13.81 & 44.22 & 33.15 & \underline{44.53} & \textbf{44.63} & 18.05 (20.40, 24.37) & 24.71 (29.08, 23.79) & 37.12 (36.86, 33.12) \\
        & SupFam & 11.99 & 7.62 & 39.08 & 26.93 & \underline{39.68} & \textbf{40.99} & 20.37 (21.21, 38.89) & 25.67 (28.03, 30.52) & 34.38 (34.84, 39.52) \\
        \addlinespace[0.35em]
        \multirow{2}{*}{FlexBFactor} & Fold & 4.17 & 6.67 & 22.32 & 18.88 & \underline{23.60} & 21.30 & 13.66 (14.88, 18.87) & 16.95 (15.79, \textbf{26.22}) & 21.66 (21.71, 22.44) \\
        & SupFam & 6.97 & 5.47 & 23.73 & 19.31 & \textbf{25.80} & 21.76 & 15.85 (15.40, 11.87) & 17.52 (17.82, 20.38) & 24.11 (\underline{24.50}, 16.78) \\
        \addlinespace[0.35em]
        \multirow{2}{*}{FlexNEQ} & Fold & 5.71 & 12.98 & 35.95 & 16.41 & \underline{45.08} & \textbf{49.64} & 15.41 (17.13, 17.37) & 18.68 (21.43, 18.95) & 29.48 (29.53, 26.18) \\
        & SupFam & 2.60 & 12.50 & 35.61 & 16.17 & \underline{45.43} & \textbf{50.15} & 16.31 (18.33, 16.82) & 21.51 (24.22, 23.48) & 31.03 (30.79, 27.10) \\
        \addlinespace[0.35em]
        \textbf{Average} &  & 7.80 & 9.84 & 33.49 & 21.81 & \underline{37.35} & \textbf{38.08} & 16.61 (17.89, 21.37) & 20.84 (22.73, 23.89) & 29.63 (29.71, 27.52) \\
        \bottomrule
    \end{tabular}}
\end{table}

\subsubsection{SLL Reconstruction Comparison}
\label{app:sll_reconstruction_comparison}
\Cref{tab:app_sll_reconstruction_comparison} reports the full reconstruction comparison behind the average results in \cref{tab:sll_reconstruction_comparison}.
The baseline values follow the reconstruction table from the original work.

\begin{table}[h]
    \centering
    \tiny
    \setlength{\tabcolsep}{2.0pt}
    \caption{Full reconstruction comparison following GCP-VQVAE~\cite{gcp_vqvae}. GCP-VQVAE Lite 2 denotes SLL. The grayed GCP-VQVAE Large column is not used for bold and underline ranking.}
    \label{tab:app_sll_reconstruction_comparison}
    \resizebox{\linewidth}{!}{%
    \begin{tabular}{llccccccccc}
        \toprule
        Dataset & Metric &
        \shortstack{GCP-VQVAE\\Lite 2} &
        \textcolor{gray!70}{\shortstack{GCP-VQVAE\\Large~\graycite{gcp_vqvae}}} &
        \shortstack{GCP-VQVAE\\Lite~\cite{gcp_vqvae}} &
        \shortstack{AIDO\\\cite{aido_structure_tokenizer}} &
        \shortstack{ESM-3\\VQVAE\\\cite{esm3}} &
        \shortstack{DPLM-2\\VQVAE\\\cite{dplm2}} &
        \shortstack{FoldToken 4\\\cite{foldtoken}} &
        \shortstack{Gaujac\\et al.\\\cite{llps}} &
        \shortstack{AminoAseed\\\cite{structtokenbench}} \\
        \midrule
        \multirow{2}{*}{CASP14} & TM-score $\uparrow$ & 0.9569 & \textcolor{gray!70}{0.9890} & \underline{0.9751} & \textbf{0.9831} & 0.9734 & 0.9056 & 0.8989 & 0.8940 & 0.3951 \\
        & RMSD $\downarrow$ & 1.1343 & \textcolor{gray!70}{0.5431} & \underline{0.8435} & \textbf{0.6587} & 1.0968 & 2.0896 & 2.0027 & 1.8913 & 14.6739 \\
        \addlinespace[0.45em]
        \multirow{2}{*}{CASP15} & TM-score $\uparrow$ & 0.9390 & \textcolor{gray!70}{0.9884} & \textbf{0.9665} & \underline{0.9660} & 0.9418 & 0.8268 & 0.9136 & 0.7911 & 0.3831 \\
        & RMSD $\downarrow$ & 2.1261 & \textcolor{gray!70}{0.5293} & \underline{0.9219} & \textbf{0.8611} & 1.3407 & 6.3826 & 1.6856 & 8.5592 & 18.5827 \\
        \addlinespace[0.45em]
        \multirow{2}{*}{CASP16} & TM-score $\uparrow$ & \underline{0.9542} & \textcolor{gray!70}{0.9857} & \textbf{0.9757} & 0.8805 & 0.9103 & 0.7852 & 0.8211 & 0.8345 & 0.3433 \\
        & RMSD $\downarrow$ & \underline{1.5030} & \textcolor{gray!70}{0.7567} & \textbf{1.0614} & 4.9773 & 4.4918 & 14.1934 & 5.4115 & 6.9375 & 25.9508 \\
        \addlinespace[0.45em]
        \multirow{2}{*}{CAMEO-2024} & TM-score $\uparrow$ & 0.9620 & \textcolor{gray!70}{0.9918} & \underline{0.9794} & \textbf{0.9851} & 0.9736 & 0.9243 & 0.9346 & 0.9114 & 0.4632 \\
        & RMSD $\downarrow$ & 1.0847 & \textcolor{gray!70}{0.4377} & \underline{0.7401} & \textbf{0.5867} & 1.0156 & 1.9410 & 1.4069 & 2.1726 & 12.6203 \\
        \addlinespace[0.45em]
        \multirow{2}{*}{Zero-shot} & TM-score $\uparrow$ & 0.9159 & \textcolor{gray!70}{0.9673} & \textbf{0.9466} & \underline{0.9202} & 0.8466 & 0.7137 & 0.8256 & 0.7662 & 0.4155 \\
        & RMSD $\downarrow$ & \underline{1.5131} & \textcolor{gray!70}{0.8193} & \textbf{1.1152} & 2.0105 & 6.5075 & 12.2182 & 4.5465 & 10.4405 & 19.2612 \\
        \midrule
        \multirow{2}{*}{Average} & TM-score $\uparrow$ & 0.9456 & \textcolor{gray!70}{0.9844} & \textbf{0.9687} & \underline{0.9470} & 0.9291 & 0.8311 & 0.8788 & 0.8394 & 0.4000 \\
        & RMSD $\downarrow$ & \underline{1.4722} & \textcolor{gray!70}{0.6172} & \textbf{0.9364} & 1.8189 & 2.8905 & 7.3650 & 3.0106 & 6.0002 & 18.2178 \\
        \bottomrule
    \end{tabular}}
\end{table}

\subsubsection{SLL Unsupervised PST Full-Run}
\label{app:sll_unsupervised_pst_full_run}
\Cref{tab:app_sll_unsupervised_pst_full_run} compares GCP-VQVAE Lite 2 with the original GCP-VQVAE tokenizers on unsupervised PST tasks.
These scores are reported as secondary representation diagnostics for SLL.

\begin{table}[h]
    \centering
    \scriptsize
    \caption{Unsupervised PST full-run comparison following GCP-VQVAE~\cite{gcp_vqvae}. GCP-VQVAE Lite 2 denotes SLL. The grayed GCP-VQVAE Large row is not used for bold and underline ranking.}
    \label{tab:app_sll_unsupervised_pst_full_run}
    \begin{tabular}{lcccc}
        \toprule
        \multirow{2}{*}{Model} &
        \multicolumn{2}{c}{ApoHolo} &
        \multicolumn{2}{c}{FoldSwitching} \\
        \cmidrule(lr){2-3}\cmidrule(lr){4-5}
        & PCC (\%) $\uparrow$ & Spearman (\%) $\uparrow$ & PCC (\%) $\uparrow$ & Spearman (\%) $\uparrow$ \\
        \midrule
        ProteinMPNN~\cite{proteinmpnn} & 35.80 & 45.22 & 51.46 & 55.91 \\
        MIF~\cite{mif} & 35.82 & 43.55 & 54.48 & 59.27 \\
        FoldSeek~\cite{foldseek} & 34.79 & 43.03 & 51.88 & 56.54 \\
        ProTokens~\cite{protokens} & 43.32 & 54.05 & 61.20 & 65.90 \\
        ESM3~\cite{esm3} & 39.76 & 50.97 & 57.12 & 62.23 \\
        VanillaVQ~\cite{structtokenbench} & 39.62 & 50.61 & 56.51 & 61.51 \\
        AIDO~\cite{aido_structure_tokenizer} & 43.02 & 54.25 & \underline{61.59} & 66.11 \\
        AminoAseed~\cite{structtokenbench} & 42.47 & 54.88 & \textbf{65.61} & \textbf{75.89} \\
        \addlinespace[0.35em]
        \textcolor{gray!70}{GCP-VQVAE Large~\graycite{gcp_vqvae}} & \textcolor{gray!70}{48.16} & \textcolor{gray!70}{59.84} & \textcolor{gray!70}{59.63} & \textcolor{gray!70}{72.33} \\
        GCP-VQVAE Lite~\cite{gcp_vqvae} & \textbf{48.56} & \textbf{59.33} & 61.11 & \underline{73.94} \\
        GCP-VQVAE Lite 2 & \underline{47.10} & \underline{57.23} & 60.69 & 67.95 \\
        \bottomrule
    \end{tabular}
\end{table}

\subsubsection{SLL Sequence-to-Structure Compatibility}
\label{app:sll_s2s_tokenizer_ablation}
This experiment tests whether SLL is easier to predict from sequence under a next-token objective.
We use a matched Prot2Token-style sequence-to-structure setup and change only the target structure tokenizer.
Both runs use 200,000 AFDB representative training samples, 10,000 validation samples, and the same eight-epoch training budget.
The sequence encoder, autoregressive decoder, data split, and optimizer setup are fixed across the two runs.

\Cref{fig:sll_s2s_tokenizer_ablation} reports train loss and validation perplexity across epochs.
The speed annotations summarize when GCP-VQVAE Lite 2 reaches comparable train-loss and validation-perplexity regimes relative to the original Lite tokenizer.
Both runs reach their best validation perplexity at epoch four and then overfit, but GCP-VQVAE Lite 2 remains lower at every epoch.
Its final validation perplexity, 22.16, is still below the best validation perplexity of the original Lite tokenizer, 22.68.

\subsection{Autoregressive Protein Language Models}
\label{app:ar_plm_additional_results}

\subsubsection{PLLM Unconditional Generation}
\label{app:pll_unconditional_generation}
\begin{figure}[h]
    \centering
    \includegraphics[width=0.78\linewidth]{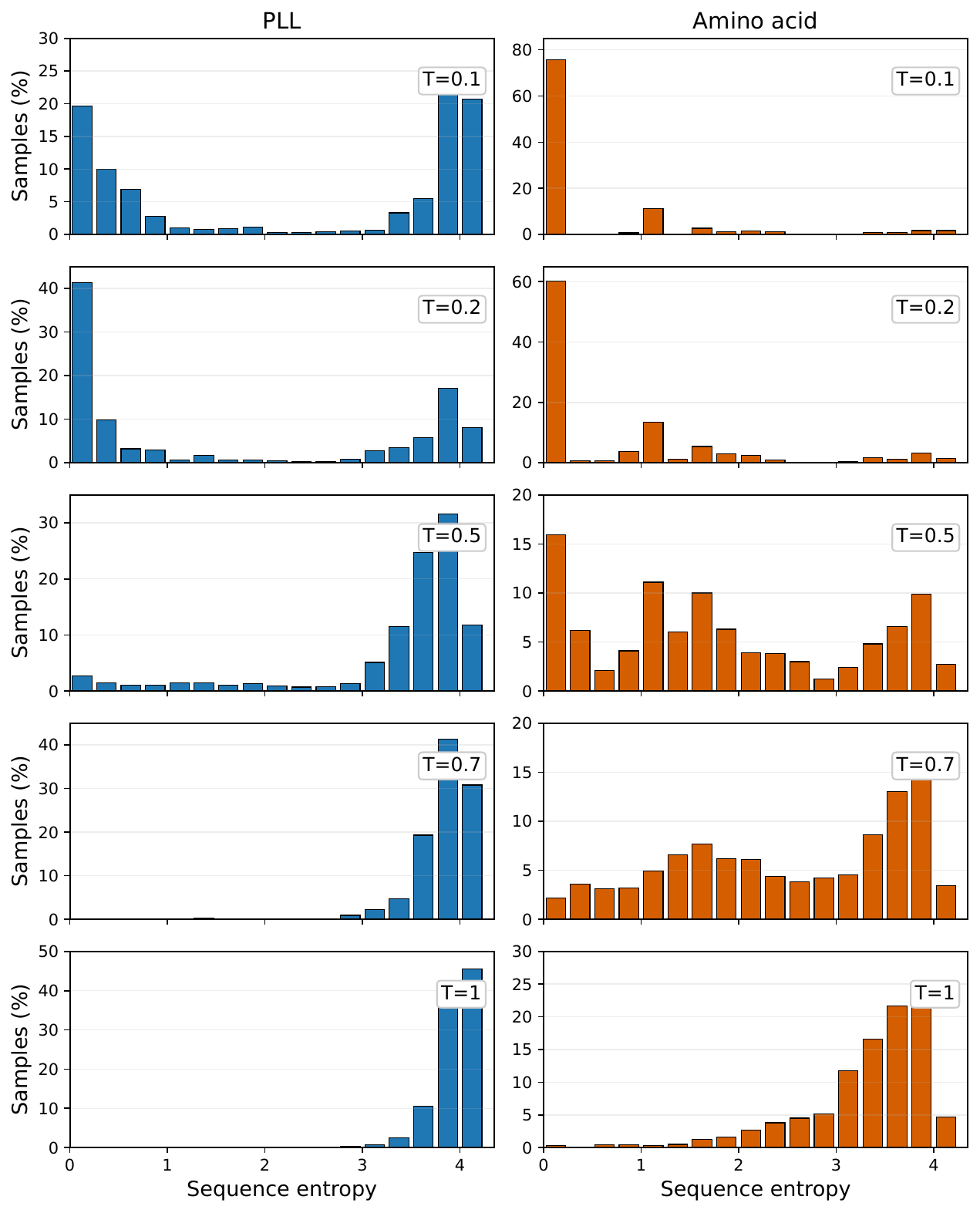}
    \caption{Unconditional sequence-entropy distributions for PLLM and amino acid generation across sampling temperatures. PLLM approaches the UniRef50-like high-entropy regime around $T=0.6$ to $T=1.0$.}
    \label{fig:app_pll_entropy_temperature_panel}
\end{figure}

\Cref{fig:app_pll_entropy_temperature_panel} shows the full entropy distribution sweep across sampling temperatures.
The UniRef50 source distribution in \cref{fig:uniref50_statistics} is concentrated near four bits, and PLLM approaches this regime by $T=0.6$ to $T=1.0$.
Direct amino acid generation remains lower-entropy across most of the sweep and only approaches the reference entropy range at much hotter sampling.
At $T=1.0$, PLLM reaches mean sequence entropy 3.927, close to the UniRef50 mean of 3.952, while the amino acid model remains lower at 3.331.
Across the \num{12} unconditional sampling temperatures, \num{5154}/\num{12000} amino acid samples (43.0\%) and \num{2383}/\num{12000} PLLM samples (19.9\%) have $H_{\mathrm{seq}}(x)<1.5$ bits. This denominator includes the unconditional generations before folding or entropy filtering and is separate from the conditional evaluation populations.

\begin{figure}[h]
    \centering
    \includegraphics[width=0.78\linewidth]{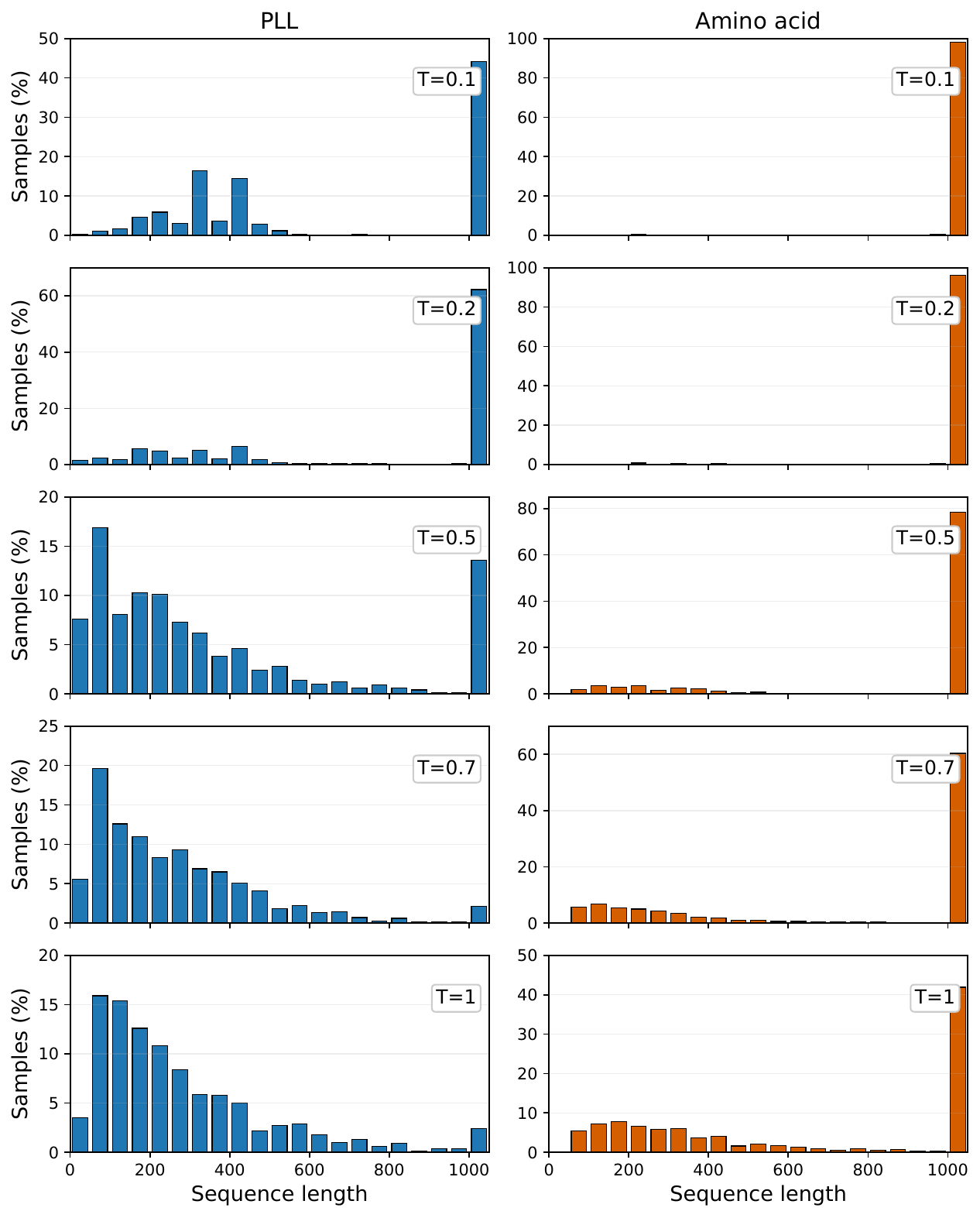}
    \caption{Unconditional length distributions for PLLM and amino acid generation across sampling temperatures. Direct amino acid generation overproduces capped sequences, while PLLM is closest to the UniRef50 length regime around $T=0.6$ to $T=1.0$.}
    \label{fig:app_pll_length_temperature_panel}
\end{figure}

\Cref{fig:app_pll_length_temperature_panel} shows the corresponding length distributions.
PLLM moves toward the right-tailed UniRef50-like length regime by moderate temperatures, while direct amino acid generation keeps a large mass near the context-length cap even at $T=1.0$.
This supports treating the amino acid failure as a stopping-prior problem, not only a missing-\texttt{<EOS>} artifact.
At $T=1.0$, PLLM has mean length 277.1 residues, close to the UniRef50 mean of 273.6 residues.
The amino acid model is much longer at the same temperature, with mean length 609.7 residues and 41.7\% of samples near the context-length cap.
Sampling hotter does not fix this tradeoff for amino acid generation.
At $T=2.0$, amino acid entropy improves, but the median length returns to the cap and 58.5\% of samples are capped.

\subsubsection{PLLM Conditional Generation}
\label{app:pll_conditional_generation}
For the fixed-length sequence-side comparison, we use conditional generation for both the amino acid model and PLLM.
The raw conditional diagnostics cover target lengths 100, 200, 300, and 500 and \num{10} temperatures from $T=0.1$ to $T=1.0$, with \num{200} generations per condition and \num{8000} samples per model.
The length token specifies the requested content length, but the model must still emit the \texttt{<EOS>} token to complete the sample.
This gives a like-for-like comparison at the same intended lengths without truncating uncontrolled generations.
We avoid post hoc length cutting because cutting a sequence before \texttt{<EOS>} can create incomplete samples and hide failures of the model's stopping policy.
Length conditioning nearly eliminates the target-length failure for both modalities.

\begin{figure}[h]
    \centering
    \includegraphics[width=0.92\linewidth]{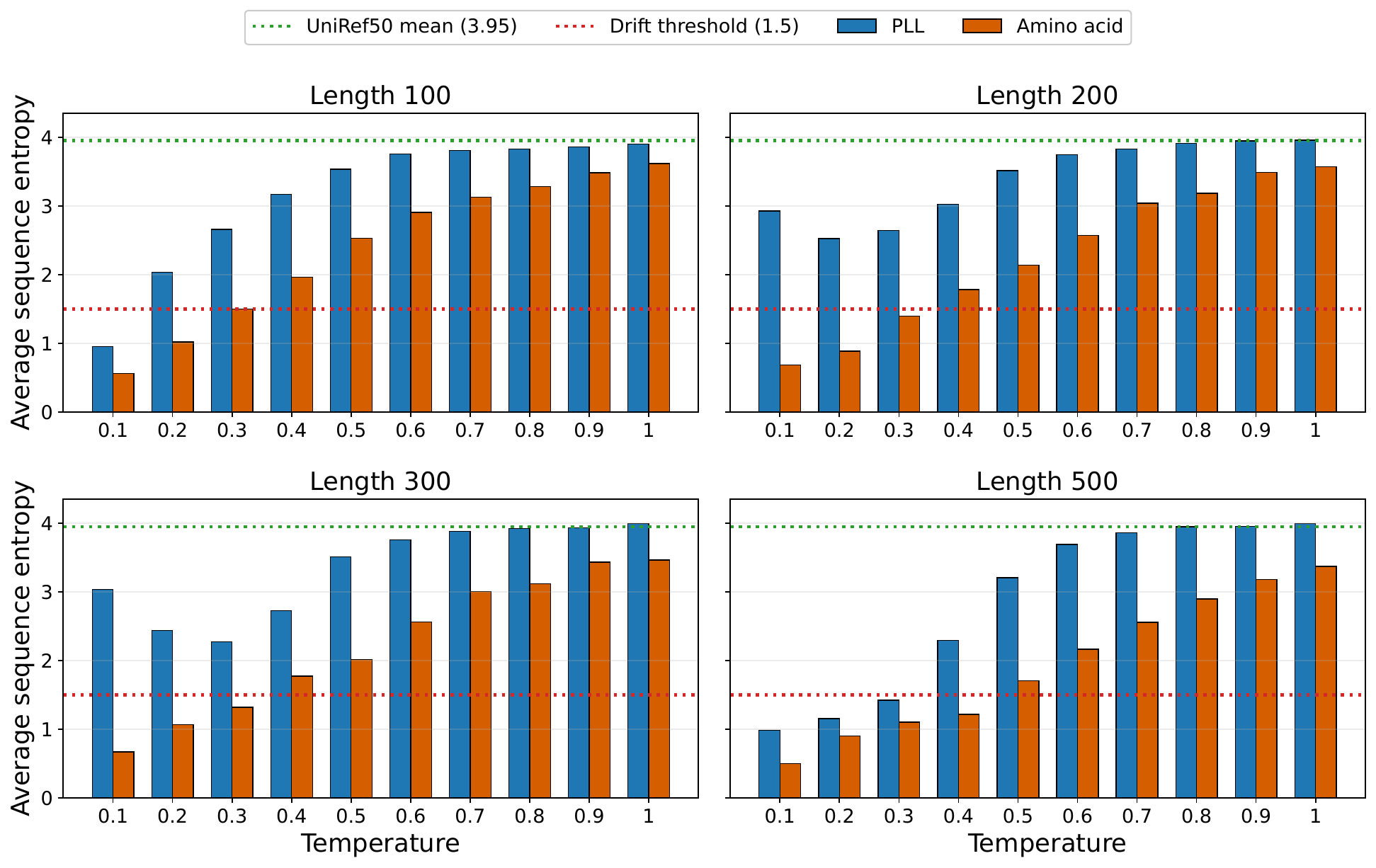}
    \caption{Length-conditioned mean sequence entropy across temperatures and target lengths. The green line marks the UniRef50 mean entropy, and the red line marks the drift threshold.}
    \label{fig:app_pll_conditional_entropy_means}
\end{figure}

\begin{figure}[p]
    \centering
    \includegraphics[width=\linewidth,height=0.78\textheight,keepaspectratio]{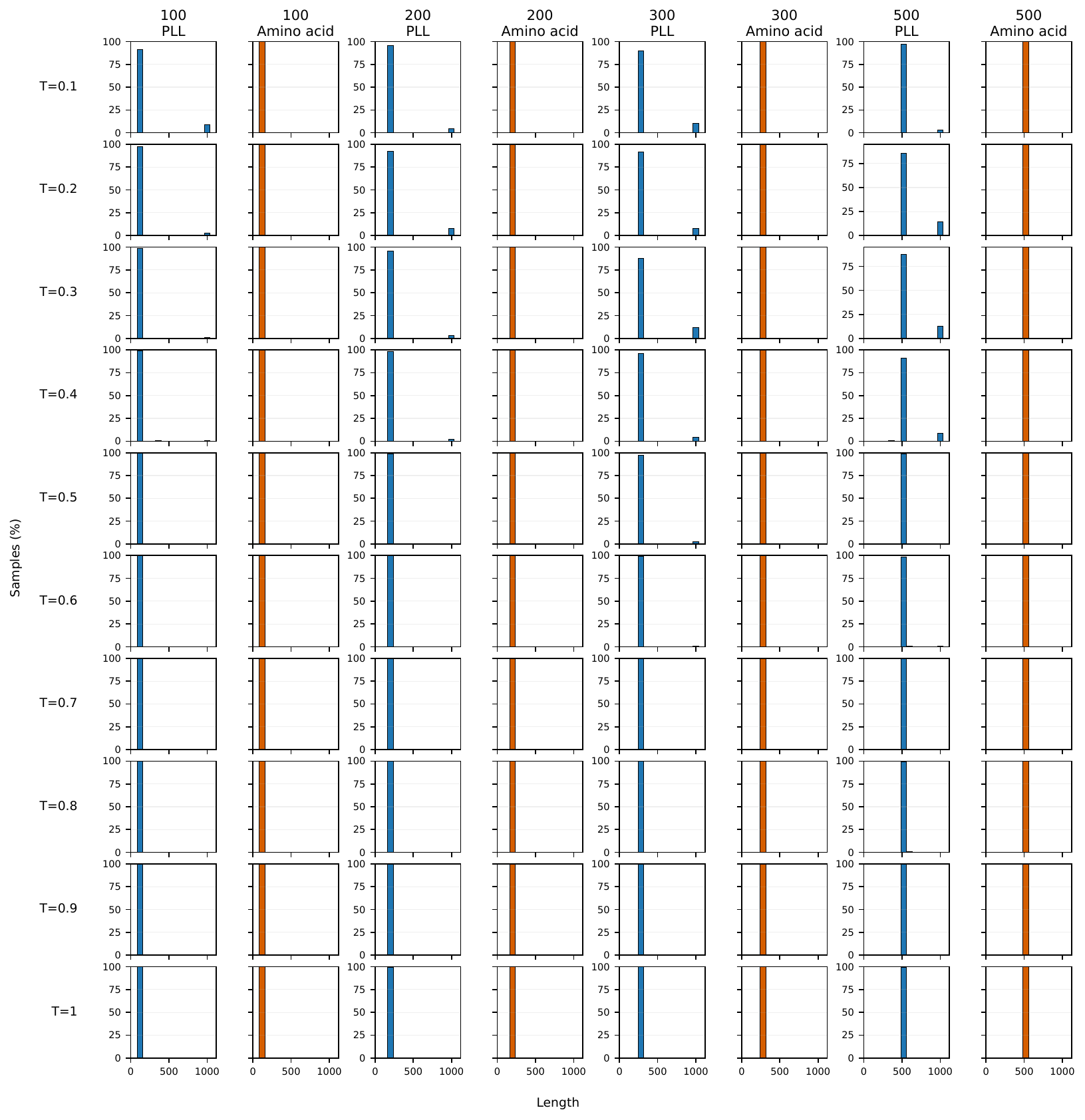}
    \caption{Length-conditioned sequence-length distributions across target lengths and temperatures. Both modalities concentrate near the requested length, showing that the length prefix largely fixes the stopping and target-size issue.}
    \label{fig:app_pll_conditional_length_distribution}
\end{figure}

\begin{figure}[p]
    \centering
    \includegraphics[width=\linewidth,height=0.78\textheight,keepaspectratio]{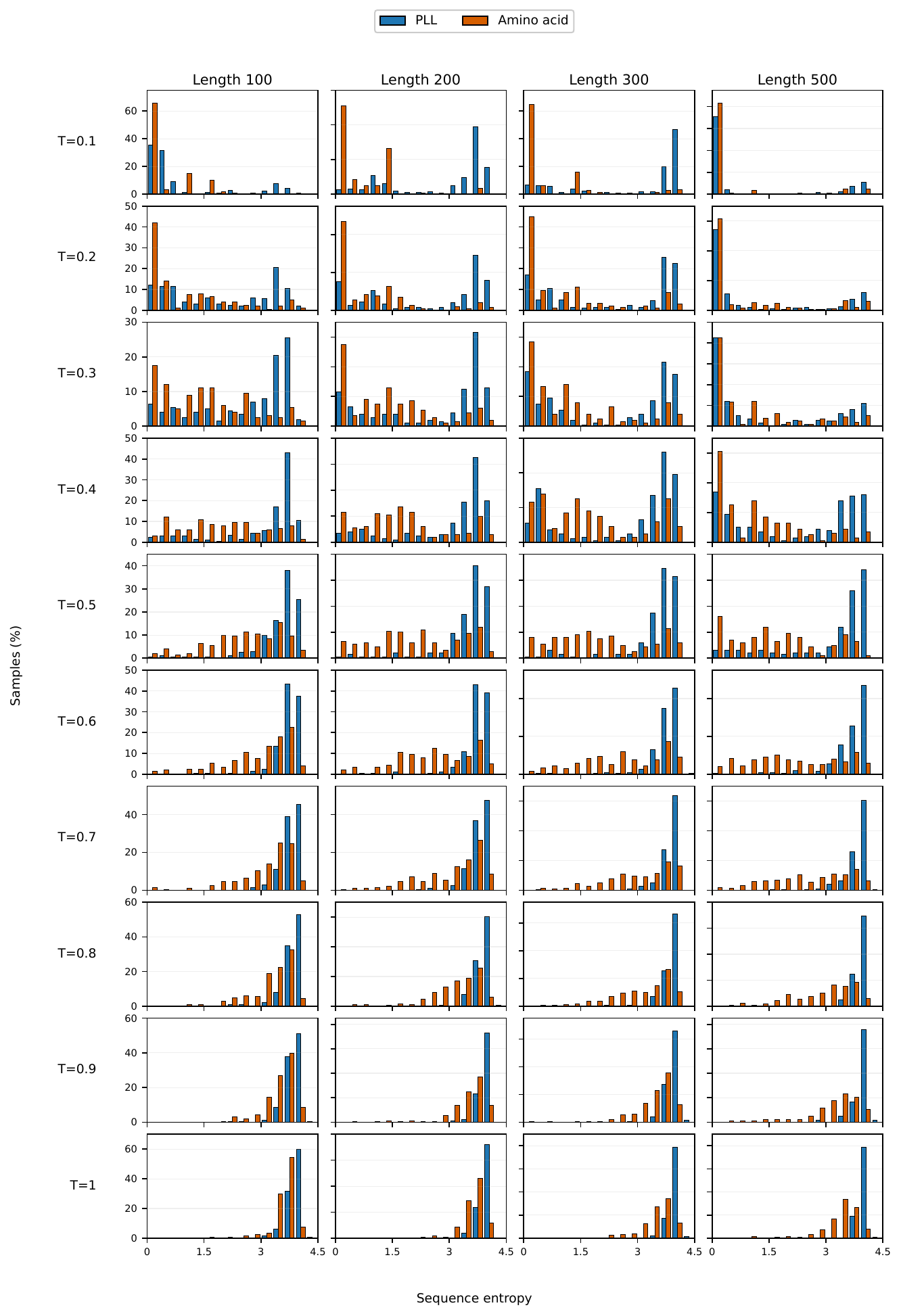}
    \caption{Length-conditioned sequence-entropy distributions across target lengths and temperatures. PLL shifts into the high-entropy natural-protein regime more consistently than amino acid generation.}
    \label{fig:app_pll_conditional_entropy_distribution}
\end{figure}

The amino acid model reaches the requested length in 99.99\% of samples, while PLLM reaches the exact requested length in 95.71\% of samples and is within five residues in 96.98\%.
However, the entropy distribution remains substantially lower for amino acid generation, showing that its low-complexity drift is not only a length-control or \texttt{<EOS>} problem.
\Cref{fig:app_pll_conditional_entropy_means,fig:app_pll_conditional_length_distribution,fig:app_pll_conditional_entropy_distribution} provide the supporting conditional-generation diagnostics.
The mean-entropy panel shows that PLLM approaches the UniRef50 entropy reference by moderate to high temperatures across target lengths, whereas amino acid generation remains lower.
The length panel confirms that the length token largely controls the requested output size for both models.
The full entropy distributions show that PLL concentrates near the natural high-entropy regime more consistently than amino acid generation.

\begin{figure}[h]
    \centering
    \includegraphics[width=0.78\linewidth]{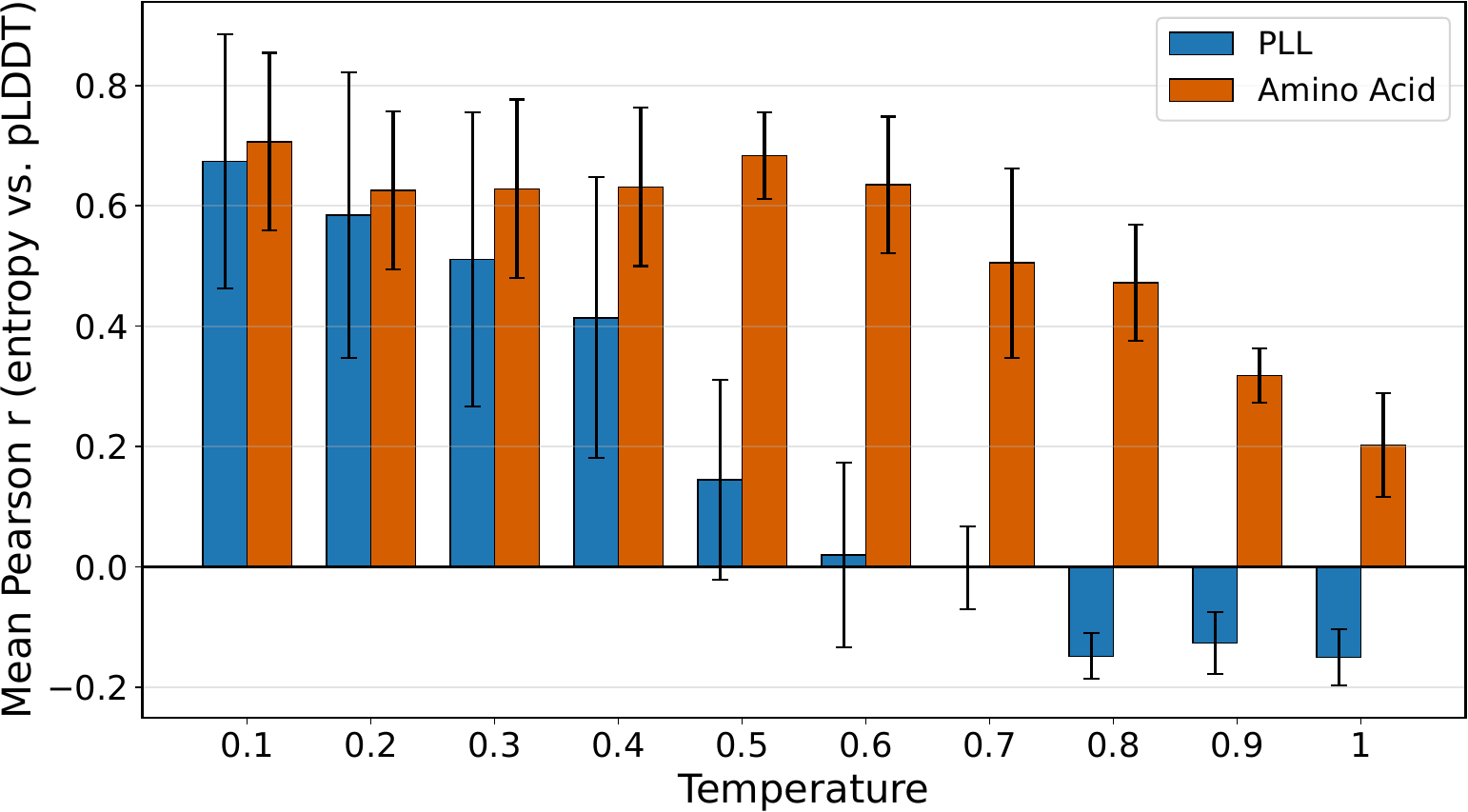}
    \caption{Length-conditioned entropy-pLDDT correlation without removing drifted sequences.}
    \label{fig:app_pll_conditional_entropy_plddt_unfiltered}
\end{figure}

\Cref{fig:app_pll_conditional_entropy_plddt_unfiltered} shows the corresponding entropy-pLDDT correlation without removing drifted sequences.
In this unfiltered view, the amino acid model keeps a stronger positive correlation across the temperature sweep, while PLL weakens toward zero and becomes negative at higher temperatures.

For the entropy-pLDDT correlation diagnostic in the right panel of \cref{fig:pll_unconditional_de_novo}, we use target lengths 100, 200, 300, and 500 across temperatures $T=0.1$ to $T=1.0$.
Each modality-length-temperature condition contributes 50 entropy-filtered ESMFold-folded sequences, giving 40 conditions and 2000 folded sequences per modality.
For both modalities, we first retain non-drifted sequences with $H_{\mathrm{seq}}(x)>1.5$ bits, fold them with ESMFold~\cite{esm2}, and parse mean pLDDT from the predicted structures.
We then compare matched retained sets across amino acid and PLLM generation.
When the direct amino acid model has fewer retained sequences than PLL, sampling is continued until the retained counts match.
This prevents a trivial comparison in which low-complexity amino acid samples dominate the correlation estimate.

After this control, amino acid generation keeps a stronger positive entropy-pLDDT correlation, while PLL is near zero or negative around $T=0.8$ to $T=1.0$.
We interpret this as pLDDT being more entangled with escape from low-complexity residue patterns in the amino acid model.

\subsubsection{SLLM Length-Conditioned Continuation}
\label{app:sll_length_conditioned_continuation}
After fully unconditional SLLM pretraining, we continue the largest SLLM for two epochs with a partial length-conditioning curriculum.
The length token is prepended to 50\% of training samples, while the remaining samples keep the unconditional format.
Validation remains unconditional, so the loss change measures whether the curriculum also improves the base next-token model rather than only a matched conditional validation format.

\begin{figure}[h]
    \centering
    \includegraphics[width=0.78\linewidth]{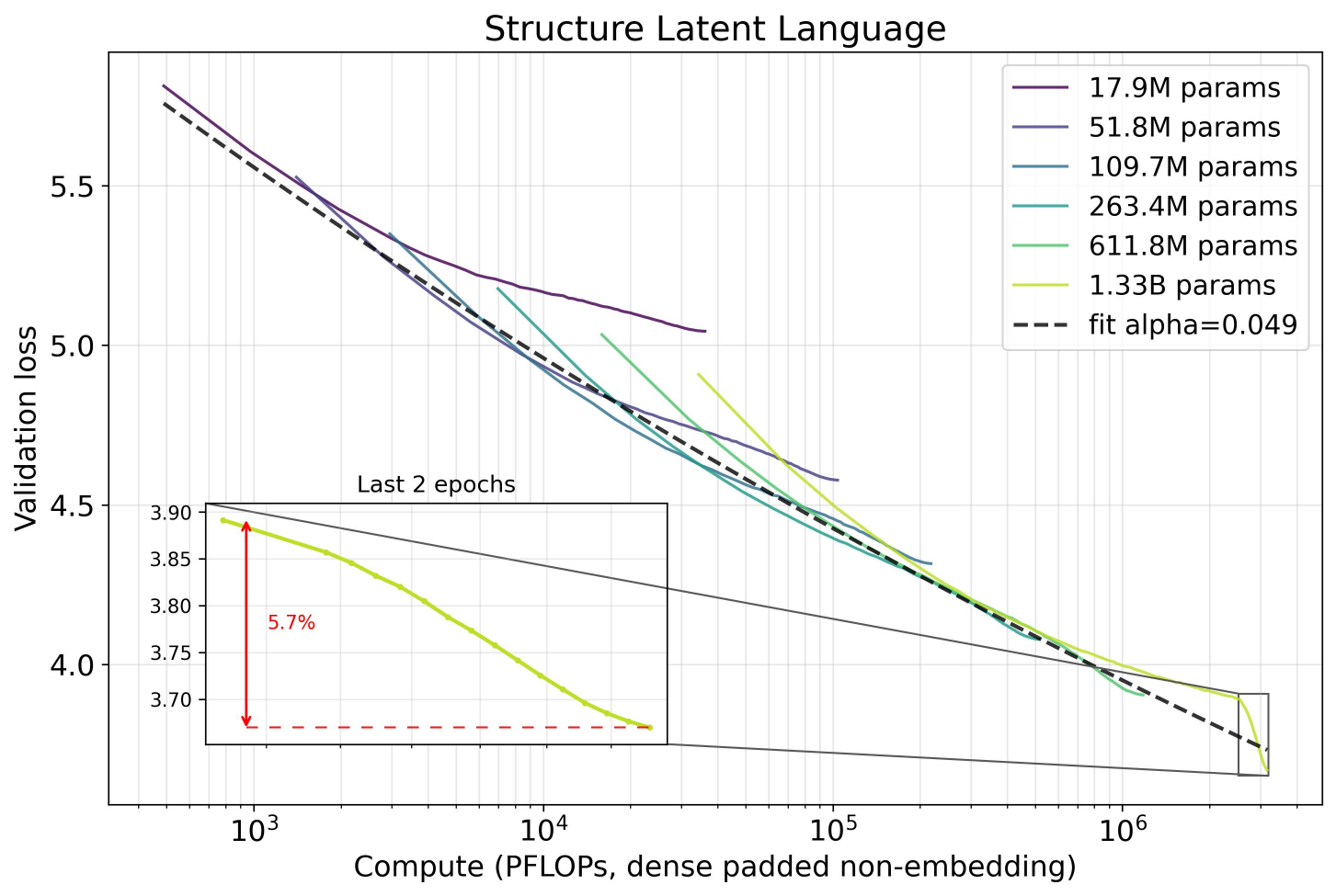}
    \caption{SLL length-conditioned continuation after unconditional pretraining. A two-epoch curriculum with length tokens on 50\% of samples lowers unconditional validation loss and adds a length-control interface for generation.}
    \label{fig:app_sll_length_conditioned_continuation_loss}
\end{figure}

\Cref{fig:app_sll_length_conditioned_continuation_loss} shows that validation loss decreases from 3.891 at the epoch-eight boundary to 3.670 after the two continuation epochs, a 5.70\% reduction.
The same checkpoint is used for length-conditioned SLLM generation.
After continuation, generated samples show nearly exact target-length control and are evaluated directly in the requested length buckets of 50, 100, 200, 300, 400, and 500 residues.
This result suggests that length conditioning is more than a sampling convenience.
It points to curriculum learning during pretraining as a useful way to improve the loss-compute frontier while enabling length-wise generation.

\subsubsection{SLLM Sequence-to-Structure Model Selection}
\label{app:s2s_model_selection}
The sequence-to-structure checkpoint is initialized from the length-conditioned SLLM in \cref{app:sll_length_conditioned_continuation}.
It is fine-tuned with E1 sequence context and reaches the evaluation checkpoint after \num{64000} optimizer steps and \num{4315655500} unmasked next-token targets in the structure-token region.
The model then generates SLL token streams with the \texttt{<sequence\_to\_structure>} condition and decodes selected streams with the fixed SLL decoder.

For confidence-based selection, we tune the DeepConf-style selector on the combined CASP14, CASP15, and CASP16 set.
This tuning set contains 118 targets and uses 64 generated candidates per target.
CAMEO-2024 is reserved for verification after the selector is fixed, with 574 targets and 256 generated candidates per target.
Both runs use temperature 1.0 and top-$p$ sampling with threshold 0.90.

The selected configuration scores each candidate by the maximum mean raw full-vocabulary entropy over sliding windows of 256 generated SLL tokens, using stride one.
Lower score means higher internal confidence.
We retain the most confident 50\% of candidates, require at least two retained samples, and select the token-similarity medoid among the retained set.
This is the offline DeepConf-style rule described in \cref{app:confidence_selection_details}.

\begin{figure}[p]
    \centering
    \begin{minipage}{0.48\linewidth}
        \centering
        \includegraphics[width=\linewidth]{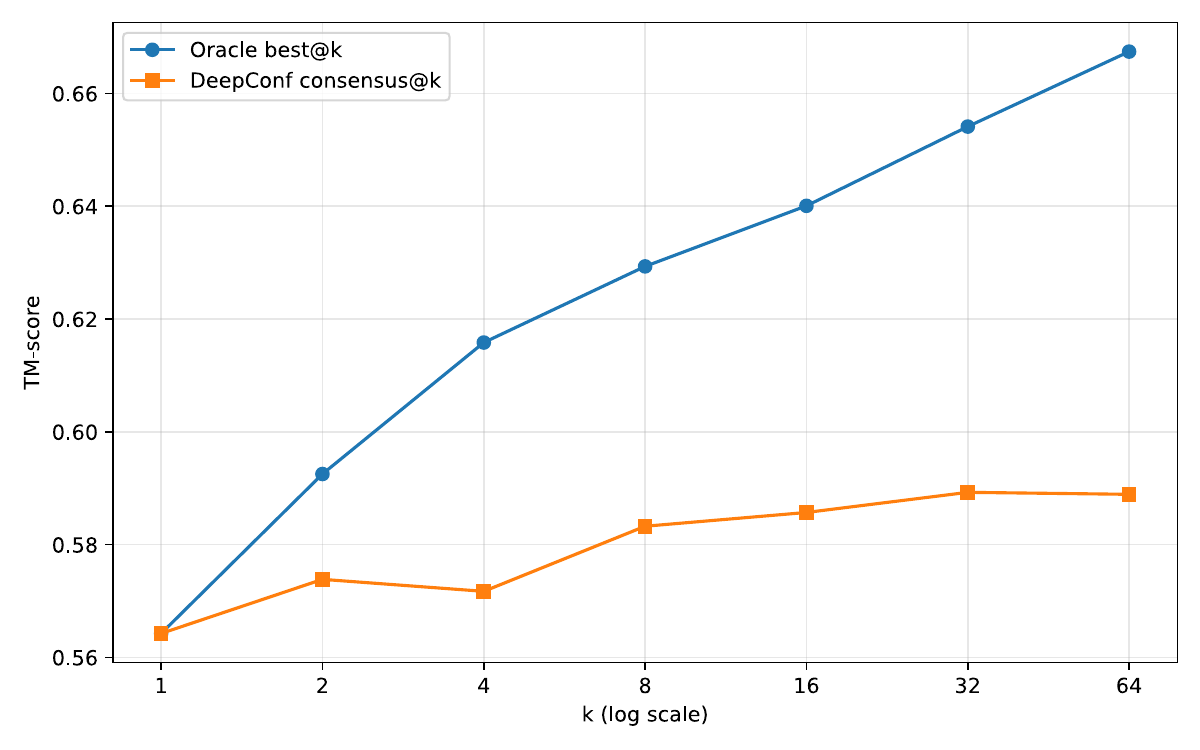}
        \footnotesize (a) TM-score
    \end{minipage}
    \hfill
    \begin{minipage}{0.48\linewidth}
        \centering
        \includegraphics[width=\linewidth]{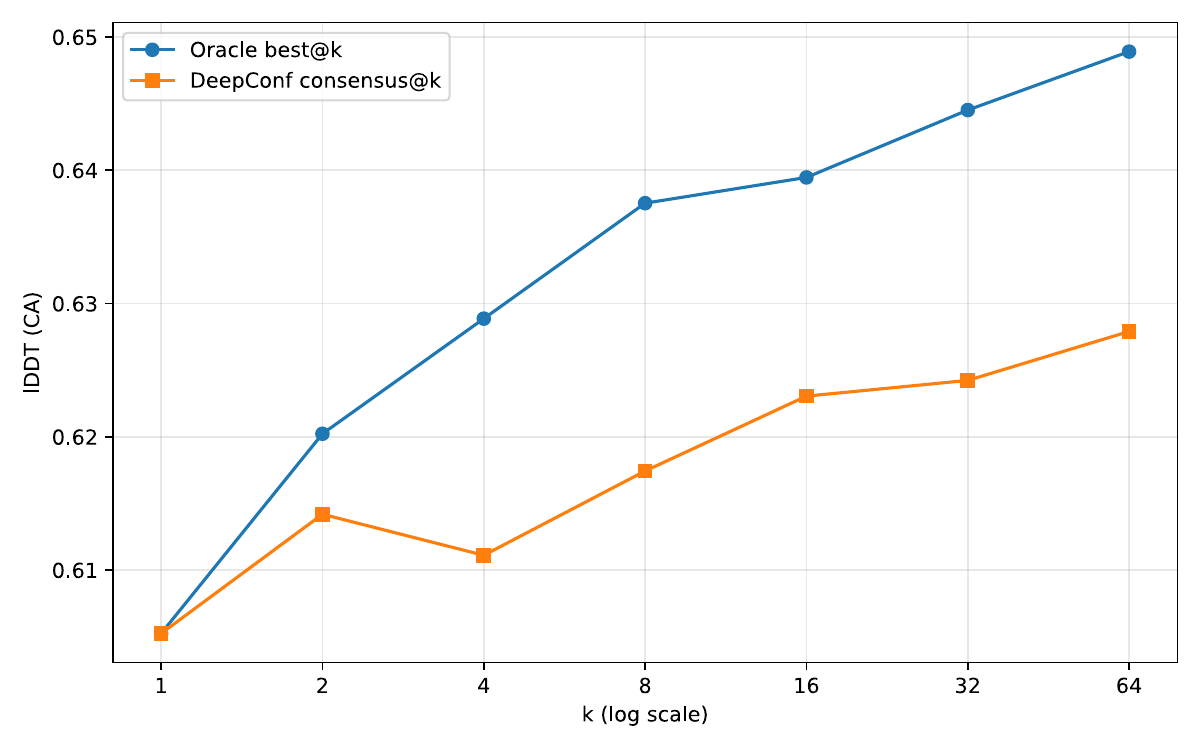}
        \footnotesize (b) lDDT-CA
    \end{minipage}
    \vspace{0.6em}
    \begin{minipage}{0.48\linewidth}
        \centering
        \includegraphics[width=\linewidth]{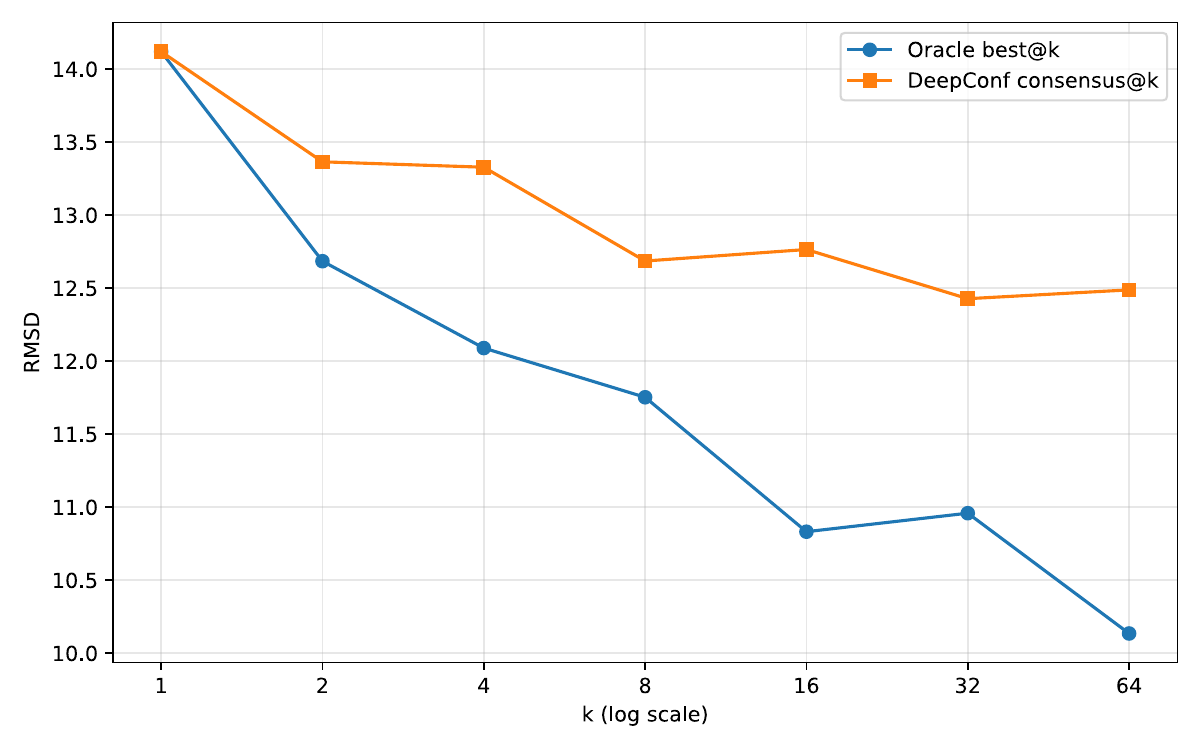}
        \footnotesize (c) RMSD
    \end{minipage}
    \hfill
    \begin{minipage}{0.48\linewidth}
        \centering
        \includegraphics[width=\linewidth]{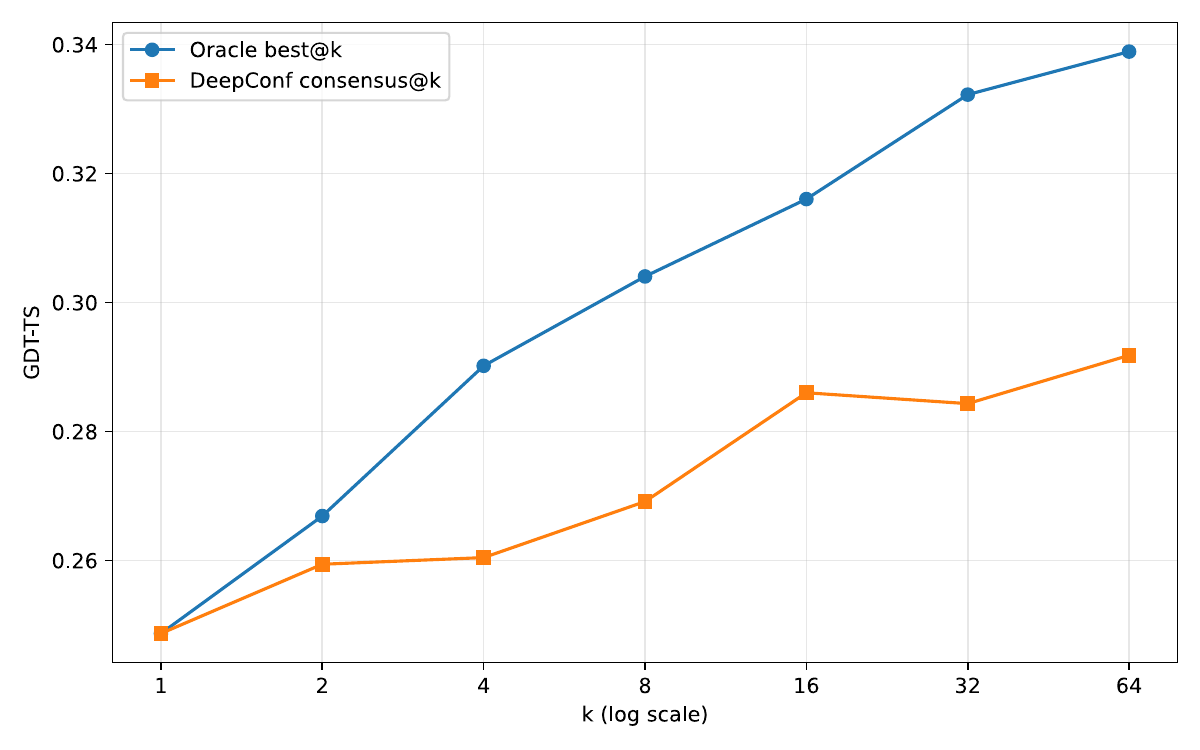}
        \footnotesize (d) GDT-TS
    \end{minipage}
    \caption{Sequence-to-structure inference-time scaling on the combined CASP14, CASP15, and CASP16 tuning set. DeepConf uses internal confidence without ground truth, while oracle best@k is a ceiling on the sampled candidate pool.}
    \label{fig:app_s2s_casp_deepconf}
\end{figure}

\begin{figure}[p]
    \centering
    \begin{minipage}{0.48\linewidth}
        \centering
        \includegraphics[width=\linewidth]{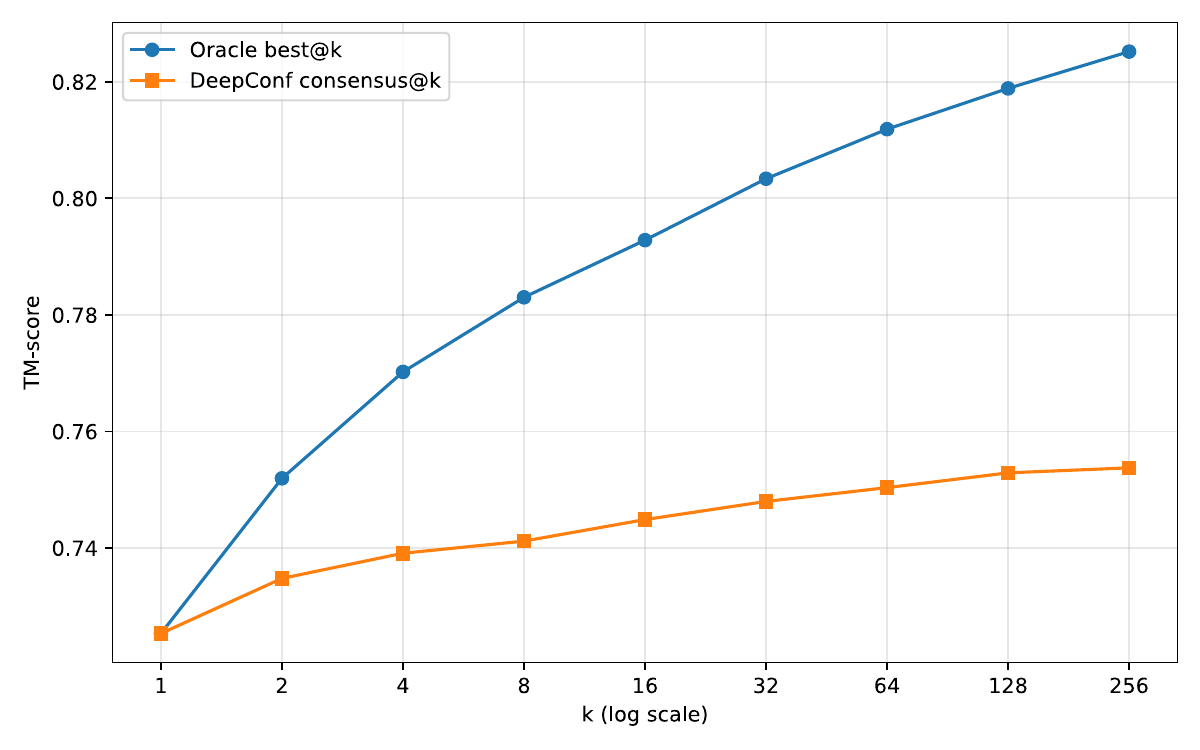}
        \footnotesize (a) TM-score
    \end{minipage}
    \hfill
    \begin{minipage}{0.48\linewidth}
        \centering
        \includegraphics[width=\linewidth]{s2s_cameo_pass_k_lddt_ca_log.pdf}
        \footnotesize (b) lDDT-CA
    \end{minipage}
    \vspace{0.6em}
    \begin{minipage}{0.48\linewidth}
        \centering
        \includegraphics[width=\linewidth]{s2s_cameo_pass_k_rmsd_log.pdf}
        \footnotesize (c) RMSD
    \end{minipage}
    \hfill
    \begin{minipage}{0.48\linewidth}
        \centering
        \includegraphics[width=\linewidth]{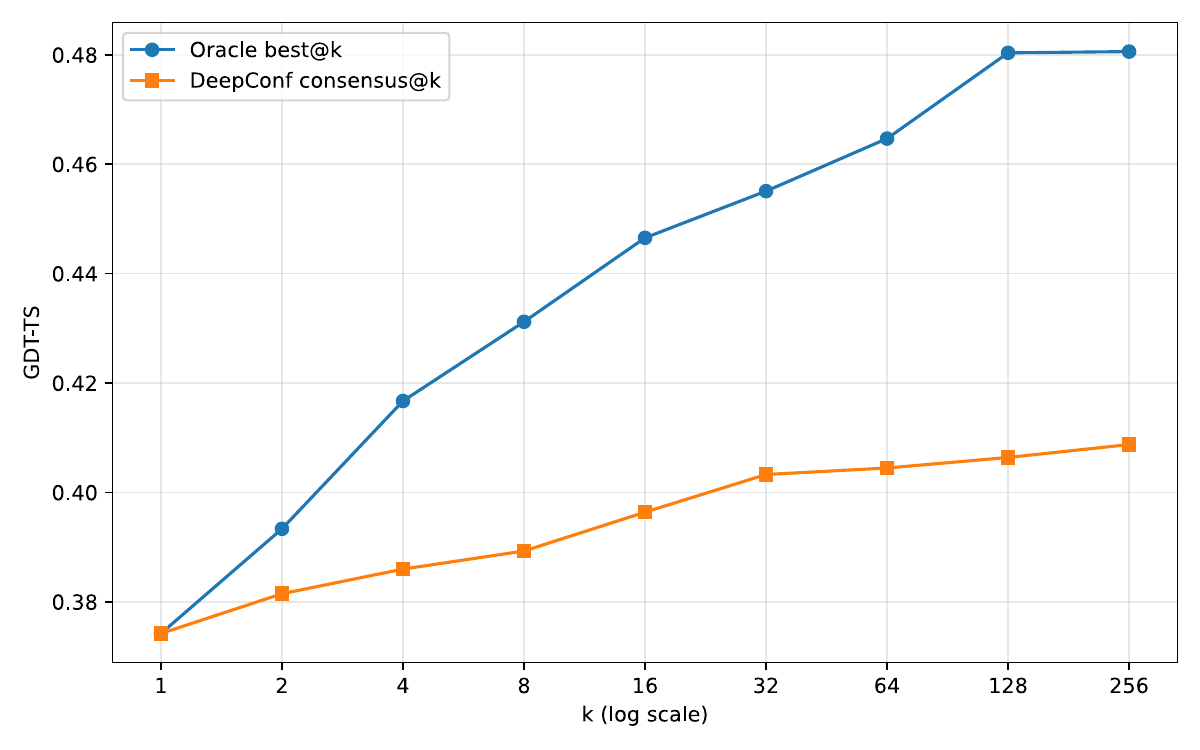}
        \footnotesize (d) GDT-TS
    \end{minipage}
    \caption{Sequence-to-structure inference-time scaling on CAMEO-2024 after fixing the selector on CASP14/15/16.}
    \label{fig:app_s2s_cameo_deepconf}
\end{figure}

\Cref{fig:app_s2s_casp_deepconf,fig:app_s2s_cameo_deepconf} show the full inference-time scaling curves.
Generation is fast enough to make this selection regime practical.
With KV caching, FA2, and model compilation, the sequence-to-structure model reaches about \num{1400} generated tokens per second on two NVIDIA RTX 6000 Ada GPUs, or about three completed samples per second on this workload.
Longer targets mainly add more generated tokens, so cached autoregressive decoding scales close to linearly with sequence length.
The fixed SLL decoder adds negligible overhead after token generation.
In our measurements, candidate-token generation for long proteins is approximately \num{1000} times faster than MSA-based AlphaFold2~\cite{alphafold} running on an NVIDIA A100 GPU.

The oracle best@k curves improve approximately log-linearly with the number of sampled candidates.
This is not an inference method because it uses the native structure, but it shows that the sampled SLL candidate pool contains better structures as $k$ increases.
On the CASP14/15/16 tuning set, DeepConf selection at $k=64$ improves average TM-score from 0.564 for a single sample to 0.589, while oracle best@64 reaches 0.667.
On CAMEO-2024, the fixed selector at $k=256$ improves average TM-score from 0.725 to 0.754, while oracle best@256 reaches 0.825.
The gap between DeepConf and oracle best@k is therefore useful headroom, not a failure of inference-time scaling.
It indicates that fast generation can partially compensate for the lower single-sample quality of the model, and that internal confidence from latent-token prediction can recover part of this gain without an external structure-quality metric.

\end{document}